\documentclass[10pt,journal,compsoc]{IEEEtran}
\usepackage{epsfig}
\usepackage{graphicx}
\usepackage{amsmath}
\usepackage{amssymb}
\usepackage{tabularx}
\usepackage{subcaption}
\usepackage[pdftex,dvipsnames]{xcolor}
\usepackage{threeparttable}
\usepackage{algorithm}
\usepackage{subfloat}
\usepackage{algorithmic}
\usepackage{enumitem}

\usepackage{multirow}
\usepackage{url}
\usepackage{wrapfig}
\usepackage{bm}
\usepackage{colortbl}
\usepackage{listings}
\usepackage{mathtools}
\usepackage{multirow}
\usepackage{booktabs}
\usepackage{lipsum}
\usepackage{color, colortbl}
\usepackage{bm}
\usepackage{textcomp}
\usepackage{makecell}

\definecolor{dark-gray}{gray}{0.20}
\definecolor{mygreen}{HTML}{39b54a}
\newcommand{\reshl}[2]{
	\textbf{#1} \fontsize{7.5pt}{1em}\selectfont\color{mygreen}{$\uparrow$ \textbf{#2}}
}
\newcommand{\pub}[1]{{\color{dark-gray}{\tiny{[{#1}]}}}}
\usepackage[pagebackref=false,breaklinks=true,colorlinks,bookmarks=false]{hyperref}

\usepackage{xargs}
\ifCLASSOPTIONcompsoc
\usepackage[nocompress]{cite}
\else
\usepackage{cite}
\fi
\ifCLASSINFOpdf
\else
\fi

\newcolumntype{x}[1]{>{\centering\arraybackslash}p{#1pt}}
\usepackage{xspace}
\makeatletter
\DeclareRobustCommand\onedot{\futurelet\@let@token\@onedot}
\def\@onedot{\ifx\@let@token.\else.\null\fi\xspace}
\newcommand{\tablestyle}[2]{\setlength{\tabcolsep}{#1}\renewcommand{\arraystretch}{#2}\centering\footnotesize}
\def\eg{\emph{e.g}\onedot} 
\def\ie{\emph{i.e}\onedot} 
 
\def\etc{\emph{etc}\onedot} \def\vs{\emph{vs}\onedot}

\def\etal{\emph{et al}\onedot}
\newcolumntype{P}[1]{>{\centering\arraybackslash}p{#1}}

\begin{document}
	
\title{Scaling Up Your Kernels: Large Kernel Design in ConvNets towards Universal Representations }
	
\author{Yiyuan~Zhang, Xiaohan~Ding, Xiangyu~Yue

\IEEEcompsocitemizethanks{
%%%%%%%%%%%%%%%%%%%%%%%%%%%
\IEEEcompsocthanksitem Y. Zhang is with the Department of Information Engineering, The Chinese University of Hong Kong, Hong Kong, China.
(Email: yiyuanzhang.ai@gmail.com)

\IEEEcompsocthanksitem X. Ding is with Tencent AI Lab, Shenzhen, China.

\IEEEcompsocthanksitem X. Yue is with the Department of Information Engineering, The Chinese University of Hong Kong, Hong Kong, China.
(Email:xyyue@ie.cuhk.edu.hk)

\IEEEcompsocthanksitem Preliminary versions of this work have appeared in CVPR 2022~\cite{ding2022scaling} and CVPR 2024~\cite{ding2024unireplknet}.
%%%%%%%%%%%%%%%%%%%%%%%%%%%
}% <-this % stops an unwanted space
}

\markboth{Journal of \LaTeX\ Class Files,~Vol.~14, No.~8, August~2015}%
	{Shell \MakeLowercase{\textit{et al.}}: Bare Demo of IEEEtran.cls for Computer Society Journals}

\IEEEtitleabstractindextext{%
\begin{abstract}
This paper proposes the paradigm of large convolutional kernels in designing modern Convolutional Neural Networks (ConvNets). We establish that employing a few large kernels, instead of stacking multiple smaller ones, can be a superior design strategy. Our work introduces a set of architecture design guidelines for large-kernel ConvNets that optimize their efficiency and performance. We propose the UniRepLKNet architecture, which offers systematical architecture design principles specifically crafted for large-kernel ConvNets, emphasizing their unique ability to capture extensive spatial information without deep layer stacking. This results in a model that not only surpasses its predecessors with an ImageNet accuracy of 88.0\%, an ADE20K mIoU of 55.6\%, and a COCO box AP of 56.4\% but also demonstrates impressive scalability and performance on various modalities such as time-series forecasting, audio, point cloud, and video recognition. These results indicate the universal modeling abilities of large-kernel ConvNets with faster inference speed compared with vision transformers. Our findings reveal that large-kernel ConvNets possess larger effective receptive fields and a higher shape bias, moving away from the texture bias typical of smaller-kernel CNNs. All codes and models are publicly available at \url{https://github.com/AILab-CVC/UniRepLKNet}, promoting further research and development in the community.
\end{abstract}
\begin{IEEEkeywords}
Convolutional Neural Network, Large-kernel ConvNets, Multimodal Learning, Neural Network Architecture Design
\end{IEEEkeywords}}

\maketitle
\IEEEdisplaynontitleabstractindextext
\IEEEpeerreviewmaketitle

\section{Introduction}\label{sec:intro}

\begin{figure}[t]
    \centering
    \includegraphics[width=0.48\linewidth]{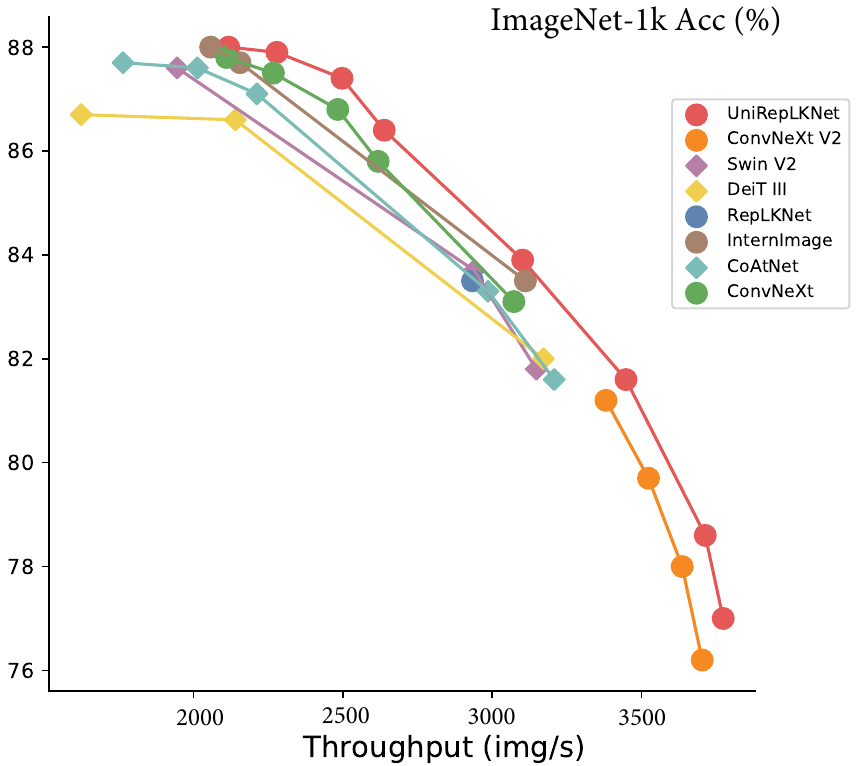}
    \includegraphics[width=0.48\linewidth]{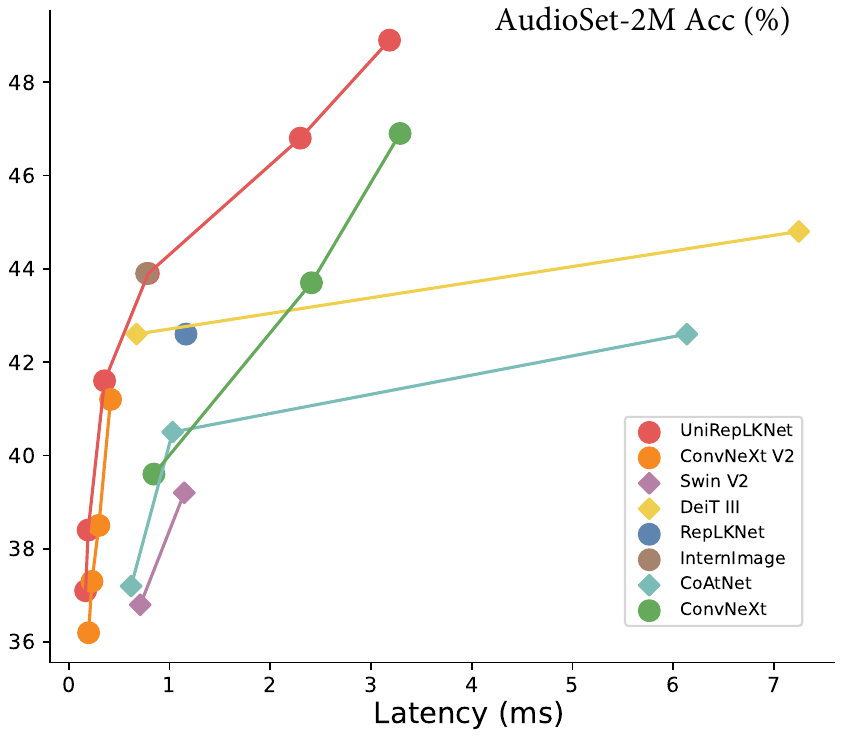}

    \includegraphics[width=0.485\linewidth]{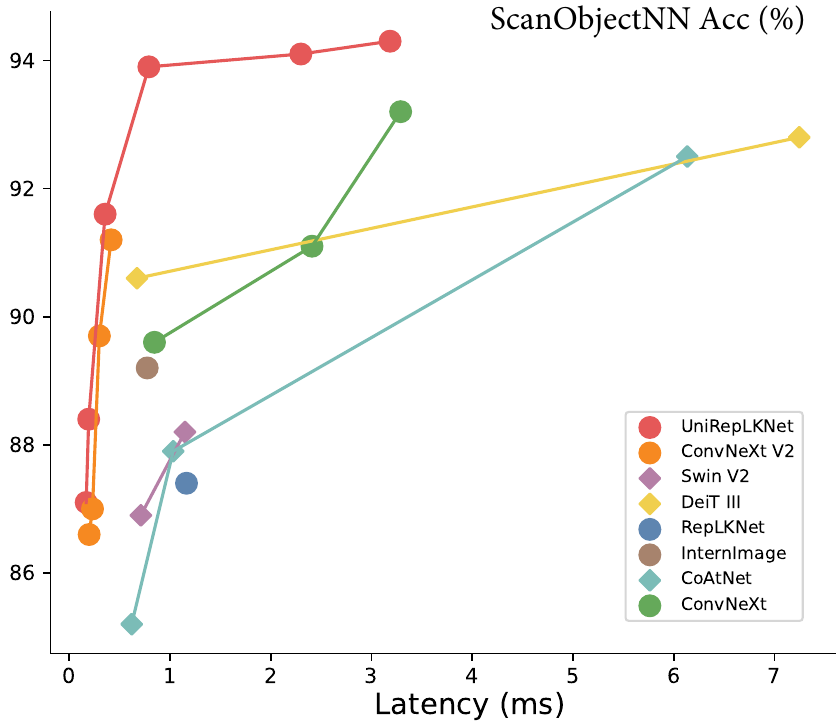}
    \includegraphics[width=0.47\linewidth]{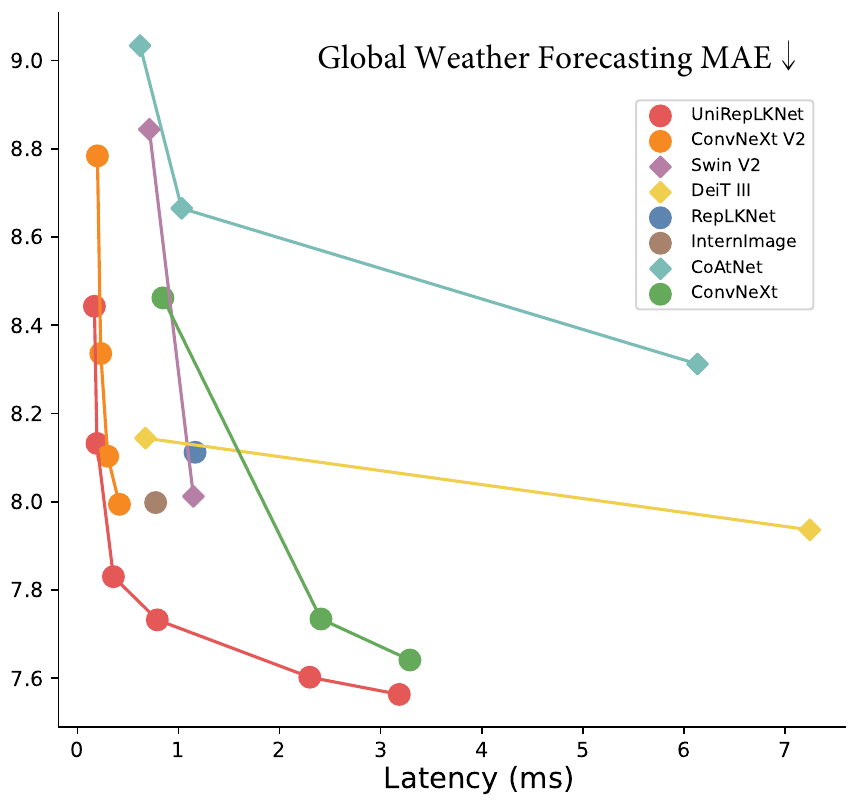}
    \caption{ \textbf{UniRepLKNet models learn universal representation across multiple modalities}. Regarding precision and efficiency across image, audio, point Cloud, and time-series modalities, UniRepLKNet delivers better scaling abilities between performance and computation burdens. The latency is tested with an A100 GPU, batch size of 128, and full precision (fp32).}
    \label{fig:cmp}
\end{figure}
\IEEEPARstart{C}{onvolutional} neural networks (ConvNets) are widely adopted in the computer vision community~\cite{krizhevsky2012imagenet,szegedy2015going,szegedy2016rethinking,szegedy2017inception,he2016deep,huang2017densely,lecun1995convolutional,caffe-lenet,zhang2018shufflenet,ding2021repvgg,chen2017deeplab,chollet2017xception,dai2017deformable,liu2022convnet,mbv1,simonyan2014very}. Recently, the dominance of ConvNets has been significantly challenged by Vision Transformers (ViTs)~\cite{vit,swin,deit,pvt,ge2023advancing} which utilize global attention~\cite{vit,pvt,bot} and window-based attention~\cite{swin,halonet,sasa}. In addition to image recognition, ViTs are also widely applied across various modalities~\cite{girdhar2023imagebind,zhang2023meta,han2024onellm}, including audio~\cite{gong2021ast}, point cloud~\cite{zhao2021point}, video~\cite{li2022mvitv2}, etc., demonstrating their potent capability of universal modeling for perception tasks.
However, the quadratic complexity, high memory costs, and slow inference speed hinder broader applications of ViTs, such as the perception of high-resolution images and long-form videos. Therefore, we ask the following question:

\vspace{0.1in}
\noindent\textit{Can we build a \textbf{ConvNet} that offers similar \textbf{universal modeling} capabilities as ViT, but with reduced complexity and significantly faster inference speed}?
\vspace{0.1in}

\begin{table*}[t]
\caption{Inference speed of a stack of 24-layer depth-wise convolutions with various kernel sizes and resolutions on a single GTX 2080Ti GPU. The input shape is (64, 384, $R$, $R$). Baselines are evaluated with Pytorch 1.9.0 + cuDNN 7.6.5, in FP32 precision.}
\label{table-speed-kernelsize}
\centering
\resizebox{0.98\linewidth}{!}{
\begin{tabular}{llccccccccccccc}
\hline
\multirow{2}{*}{Resolution $R$} & \multirow{2}{*}{Impl.} & \multicolumn{10}{c}{Latency (ms) @ Kernel size} \\
&                           
& 3     & 5     & 7     & 9     & 13   & 17     & 21    & 27    & 29    & 31        \\ \hline
\multirow{2}{*}{$16\times 16$}         & Pytorch        
& 5.6   & 11.0  & 14.4  & 17.6  & 36.0 & 57.2   & 83.4  & 133.5 & 150.7 & 171.4       \\
& Ours                      
& 5.6   & 6.5   & 6.4   & 6.9   & 7.5  & 8.4    & 8.4   & 8.4   & 8.3   & 8.4       \\ \hline
\multirow{2}{*}{$32\times 32$}         & Pytorch        
& 21.9  & 34.1  & 54.8  & 76.1  & 141.2 & 230.5 & 342.3 & 557.8 & 638.6 & 734.8       \\
& Ours                      
& 21.9  & 28.7  & 34.6  & 40.6  & 52.5  & 64.5  & 73.9  & 87.9  & 92.7  & 96.7       \\ \hline
\multirow{2}{*}{$64\times 64$}         & Pytorch       
& 69.6  & 141.2 & 228.6 & 319.8 & 600.0 & 977.7 & 1454.4 & 2371.1 & 2698.4 & 3090.4      \\
& Ours  
& 69.6  & 112.6 & 130.7 & 152.6 & 199.7 & 251.5 & 301.0 & 378.2 & 406.0 & 431.7       \\ \hline
\end{tabular}
}
\end{table*}

\begin{figure*}[t]
\begin{center}
    \includegraphics[width=0.98\linewidth]{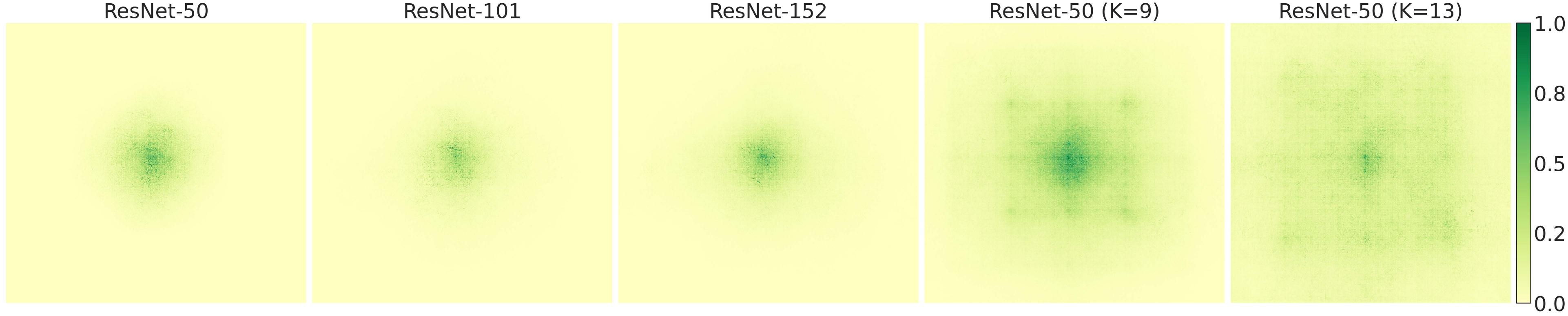}
    \vspace{-3mm}
    \caption{The \emph{Effective Receptive Field (ERF)} of ResNet-50/101/152 and the large kernel (K) variants of ResNets, respectively. A more widely distributed dark area indicates a larger ERF. More layers (\eg, from ResNet-101 to ResNet-152) help little in enlarging ERFs. Instead, the large-kernel ConvNets effectively obtain large ERFs.}
    \label{fig:erf}
\end{center}
\end{figure*}	

Diving into the advantages of ViTs, the global attention mechanism brings out long-range dependencies and contextual relationships~\cite{hinton2021represent,zhu2019empirical,han2021demystifying,wu2019pay,cordonnier2019relationship}. This prompts us to consider: \textit{how to enhance long-range dependencies and contextual relationship in ConvNets?} Large convolutional kernels appear to be the solution for ConvNets after a decade's exploration~\cite{xu2014deep,peng2017large,liu2022convnet,ding2022scaling,liu2022more,ding2024unireplknet}. In 2014, Xu~\etal~\cite{xu2014deep} proposed the inverse kernel and deconvolution to add larger spatial support for image denoising. Following this, large kernels were introduced to segmentation tasks in 2017 for larger ERFs~\cite{peng2017large}. Additionally, in 2022, Liu~\etal~\cite{liu2022convnet} scaled kernels up to \(7\times 7\) within the macro architecture of Swin Transformer~\cite{liu2021swin}. Then SLak~\cite{liu2022more} utilized sparse large kernels of size \(51\times 51\), demonstrating the efficiency and superiority of large-kernel ConvNets. Despite these advancements, a significant question becomes more clear: \textit{How can we design a large-kernel ConvNet with universal modeling abilities, high efficiency, and promising scalability for both data and parameters?}

In this paper, we explore the design of an efficient and universal architecture, specifically large-kernel ConvNets, by rethinking the traditional design of using a deep stack of small kernels. When we add a 3$\times$3 convolution to a small-kernel ConvNet, we expect it to have three simultaneous effects - \textbf{1)} expanding the receptive field, \textbf{2)} increasing the abstraction hierarchy of spatial patterns (\eg, from angles and textures to object shapes), and \textbf{3)} improving the model's general representation capability by increasing its depth, thus introducing more learnable parameters and non-linearities. In contrast, we argue that such three effects in a large-kernel architecture should be decoupled, as the model should leverage the substantial strength of large kernels - \emph{the ability to see wide without going deep}. Since increasing the kernel size is more effective than stacking layers for enlarging the ERF~\cite{erf}~\footnote{Referring to this paper, the growth order of ERF is \(\mathcal{O}(k\sqrt{n})\), where \(k\) is the kernel size and \(n\) is the depth of the convolutional layer.}, a sufficient ERF can be established with only a few large-kernel layers. This allows the compute budget to be allocated to other efficient structures that more effectively increase the abstract hierarchy of spatial patterns or the overall depth. For example, when the objective is to extract higher-level local spatial patterns from lower-level ones, a 3$\times$3 convolution might be a more suitable option than a large-kernel convolution layer. The reason is that the latter demands more computations and may result in patterns that are no longer confined to smaller local regions, which could be undesirable in specific scenarios. 

Concretely, we propose a roadmap (\S~\ref{sec:method}) to \textit{Uniervsal ConvNets} on both macro and micro designs of a large-kernel ConvNet architecture:
\begin{itemize}
    \item \textbf{Step 1}: making large-kernels practical (\S~\ref{sec:step1}), which should be both efficient~(\S~\ref{sec:step1:sub1}) and effective~(\S~\ref{sec:step1:sub2}).
    \item \textbf{Step 2}: designing a modern large-kernel ConvNet architecture, including deep blocks design~(\S~\ref{sec:step2:sub1}), micro design with structural re-paramterization~(\S~\ref{sec:method:micro}), kernel size principle~(\S~\ref{sec:step2:kernel}), and scaling rules~(\S~\ref{sec:step2:scaling}) of large kernel ConvNets, respectively.
    \item \textbf{Step 3}: generalizing large-kernel ConvNets to multiple modalities including time-series, audio, point cloud, and video~(\S~\ref{sec:step3}).
    \item \textbf{Step 4}: fusing multimodal features with large kernel convolution operators, an alternative to the cross-attention mechanism~(\S~\ref{sec:step4}).
\end{itemize} 

A ConvNet constructed following such guidelines (Fig.~\ref{fig-arch}) achieves the three aforementioned effects separately. It utilizes a modest number of large kernels to guarantee a large ERF, as shown in Fig.~\ref{fig:erf}, employs small kernels to extract complicated spatial patterns more efficiently, and incorporates multiple lightweight blocks to further increase depth and enhance representational capacity.

As shown in Fig.~\ref{fig:cmp}, our architecture achieves leading performance on universal understanding tasks including ImageNet classification~\cite{deng2009imagenet}, AudioSet-2M~\cite{gemmeke2017audio}, ScanObjectNN~\cite{uy-scanobjectnn-iccv19}, and Global Weather Forecasting tasks~\cite{wu2023interpretable}. In image recognition, UniRepLKNet outperforms existing large-kernel ConvNets such as RepLKNet~\cite{ding2022scaling}, SLaK~\cite{liu2022more}, and recent powerful architectures including ConvNeXt V2~\cite{woo2023convnext}, FastViT~\cite{vasu2023fastvit}, Swin V2~\cite{liu2022swin} and DeiT III~\cite{touvron2022deit}, in terms of both accuracy and efficiency. Moreover, our architecture exhibits a significantly higher shape bias~\cite{tuli2021convolutional,modelvshuman} compared to existing ConvNets and ViTs. Specifically, it makes predictions based more on the overall shapes of objects than on textures, which aligns with the human visual system and results in better generalization. This may explain its superiority in downstream tasks. In addition, as we scale our model to 1.4B with training data of 10B image-text pairs from LAION-5B dataset~\cite{schuhmann2022laion} for CLIP~\cite{openclip} pretraining, it demonstrates impressive zero-shot abilities across 26 datasets (Table~\ref{tab:clip}) on the widely adopted CLIP benchmark\footnote{\url{https://github.com/LAION-AI/CLIP_benchmark}}. Moreover, UniRepLKNet also shows outstanding performance on the large vision-language model benchmarks (Table~\ref{tab:vlm_eval}).

RepLKNet~\cite{ding2022scaling} was proposed partly ``in defense of ConvNets'' as ViTs began to dominate multiple image recognition tasks previously led by ConvNets. Moreover, given that transformers have demonstrated universal perception capability across multiple modalities~\cite{zhang2024multimodal,zhang2023meta}, this work aims not only to reclaim the leading position in image recognition tasks by surpassing the performance of ViTs but also to contribute to areas where ConvNets were not traditionally dominant. Specifically, we achieve impressive performance on \emph{audio, video, point cloud, and time-series} tasks, with remarkably universal and simple solutions. We use modality-specific preprocessing approaches to transform all data into 3D embedding maps, similar to how images are processed, and use the same architecture as the backbone to process these embedding maps. Our model demonstrates \textbf{\emph{uni}versal perception ability across multiple modalities with a \emph{uni}fied architecture}, hence the name \textbf{UniRepLKNet}. Impressively, UniRepLKNet achieves remarkable results even on modalities that were not considered the stronghold of ConvNet, \eg, audio and temporal data. On a large-scale time-series forecasting task predicting the global temperature and wind speed, UniRepLKNet even outperforms the latest state-of-the-art transformer customized for the task. These results not only signify a \emph{``\textbf{comeback}''} for ConvNet in its original domain but also highlight the potential of large-kernel ConvNet to \emph{``\textbf{conquer}''} new territories, expanding its applicability and versatility across various tasks.

This work builds upon our preliminary conference papers in CVPR 2022~\cite{ding2022scaling} and CVPR 2024~\cite{ding2024unireplknet}, and we present a substantial extension of it in various aspects. \textit{First}, we further develop the large-kernel convolution operators as a higher-efficiency alternative of attention mechanism on both learning universal representations (\S~\ref{sec:step1} \& \S~\ref{sec:step2} \& \S~\ref{sec:step3}) and fusing diverse features across modalities (\S~\ref{sec:step4}). 
\textit{Second}, we continue to explore the potential of large-kernel ConvNets on additional large-scale multimodal comprehension abilities (Table~\ref{tab:audio} \& Table~\ref{tab:video} \& Table~\ref{tab:pcd}) including AudioSet-2M for audio and Objaverse for point clouds, \etc. 
\textit{Third}, we scale the proposed architectures to 1.4B parameters and validate the transferable abilities of UniRepLKNet in learning 10 billion image-text pairs with CLIP~\cite{clip} for zero-shot recognition tasks, further illustrating their efficiency and advancements in architectural and data scalability (Table~\ref{tab:clip}).
\textit{Fourth}, to thoroughly investigate the efficiency advantages of ConvNets, we use UniRepLKNet for training large vision-language models (Table~\ref{tab:vlm_eval}), which shows promising performance on comprehensive zero-shot visual question-answering benchmarks. 
\textit{Last but not least}, we summarize the architectural design as a roadmap to universal ConvNets, hoping to foster research efforts in designing more efficient architectures.

\section{Related works}\label{sec:related}
\noindent\textbf{Large kernels in early ConvNets}. Early ConvNets, such as AlexNet~\cite{krizhevsky2012imagenet} and Inception~\cite{szegedy2015going,szegedy2016rethinking,szegedy2017inception}, initially used large kernels (\(7\times 7\) or \(11\times 11\)) to capture spatial features. However, the trend shifted with VGG-Net, which favored smaller, more frequent layers~\cite{simonyan2014very}. Innovatively, the Global Convolution Network (GCN) \cite{peng2017large} utilized very large kernels (\(1\times K\) followed by \(K\times 1\)) to improve semantic segmentation. Local Relation Networks (LR-Net) \cite{hu2019local} explored dynamic kernel sizes and found that performance peaked with \(7\times 7\) kernels but declined with larger sizes, illustrating the challenges of balancing kernel size with network efficiency.

\noindent\textbf{Explorations with large kernels}. Expanding the traditional definition of kernels in convolutional networks, Swin Transformer~\cite{swin} innovatively employed shifted attention mechanisms with window sizes ranging from 7 to 12, effectively functioning as dynamic kernels. Research by Han \textit{et al.}\cite{han2021demystifying} demonstrated that replacing the attention layers in Swin Transform with either static or dynamic \(7\times7\) convolution layers yielded results comparable to the original model. Additionally, the MetaFormer\cite{yu2021metaformer} proposed that a large-kernel pooling layer could serve as a viable alternative to self-attention mechanisms. Further extending the concept, the Global Filter Network (GFNet)~\cite{rao2021global} refined spatial connection weights via the Fourier domain, achieving a global convolution effect similar to circular convolutions in the spatial domain, underscoring the versatile applications of large-scale kernels across different network architectures.

\noindent\textbf{Modern ConvNets with very large kernels}. 
The introduction of RepLKNet~\cite{ding2022scaling} marked a significant shift in ConvNet design by demonstrating that enlarging kernel sizes can improve performance, particularly in downstream applications. This approach introduced several key design strategies, such as integrating shortcuts with large kernels for better microstructural efficiency. While RepLKNet was inspired by the straightforward architecture of the Swin Transformer, subsequent research has expanded on this idea. Liu~\textit{et al.}~\cite{liu2022more} and others pushed the boundaries further by scaling up kernel sizes, applying these concepts to 3D vision tasks~\cite{chen2023largekernel3d}, image dehazing~\cite{luo2023lkd} and super-resolution~\cite{xie2023large}. Despite these advances, the architectural nuances of ConvNets with large kernels remain relatively unexplored, indicating a promising area for future research.

The growing interest in large-kernel ConvNets is driven by their effectiveness in capturing fine-grained and global spatial features. However, existing models often integrate large kernels with additional mechanisms, limiting the understanding of their standalone potential. Research shows that scaling kernel sizes improves performance, yet a universal large-kernel ConvNet architecture remains undeveloped. This work proposes a simplified, universal design that retains the spatial extraction benefits of large kernels, bridging the flexibility of Transformer models with the efficiency of traditional ConvNets, and extending applicability across diverse tasks.

\section{A Roadmap to Universal ConvNets}~\label{sec:method}
\begin{figure}[t]
    \begin{center}
        \includegraphics[width=0.93\linewidth]{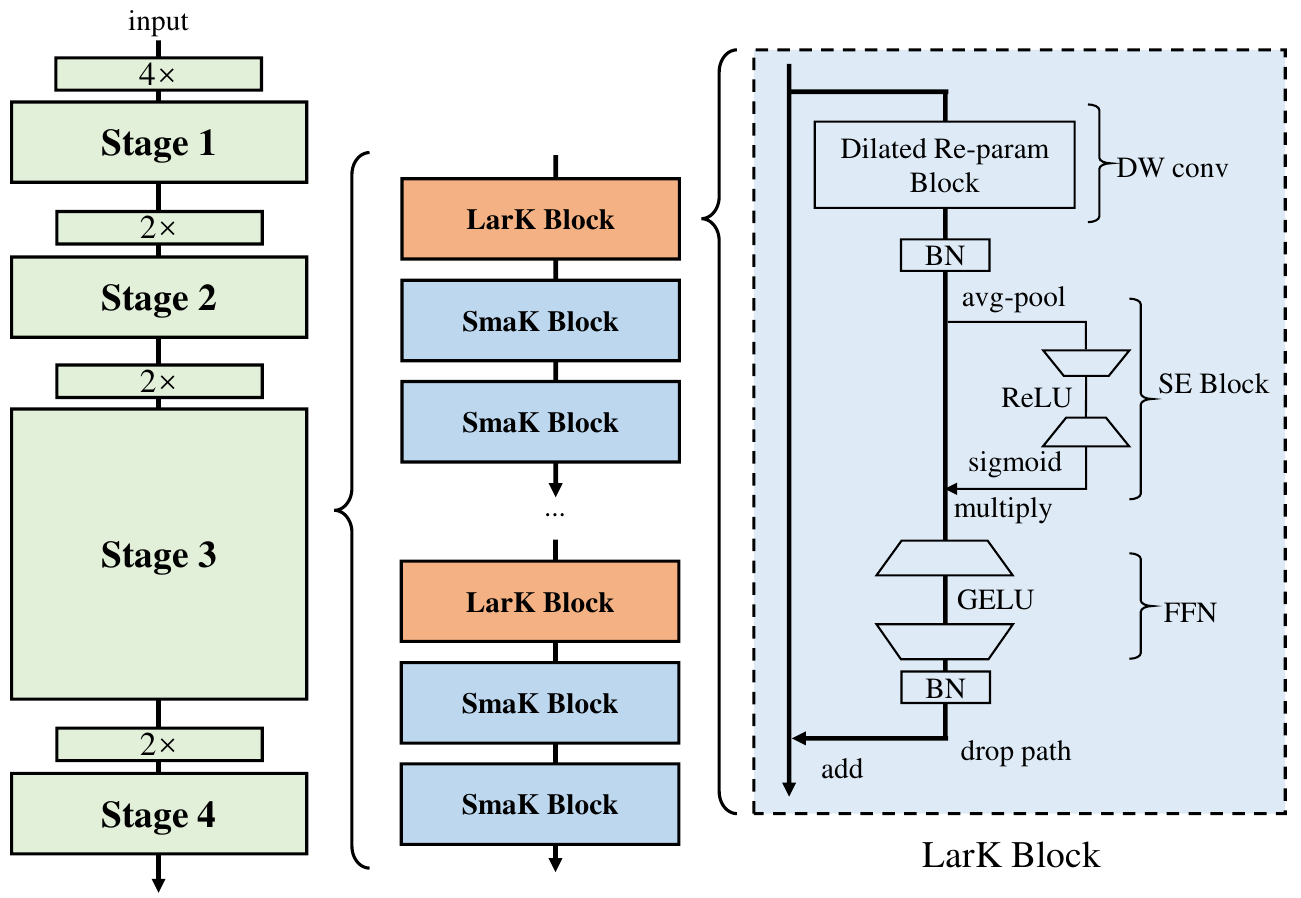}
        \vspace{-2mm}
        \caption{Architectural design of UniRepLKNet. A LarK Block comprises a Dilated Reparam Block proposed in this paper, an SE Block~\cite{hu2018squeeze}, an FFN, and Batch Normalization (BN)~\cite{ioffe2015batch} layers. The only difference between a SmaK Block and a LarK Block is that the former uses a depth-wise 3$\times$3 conv layer in replacement of the Dilated Reparam Block in the latter. Stages are connected by down-sampling blocks implemented by stride-2 dense 3$\times$3 conv layers. We may flexibly arrange the blocks in different stages and the details of our provided instances are shown in Table~\ref{table-instances}.}
        \label{fig-arch}
    \end{center}
\end{figure}

\begin{table*}[ht]
\vspace{-.2em}
\centering
\caption{\textbf{Architectural design choices of Step 1: Making Large Kernels Practical}. We report the Top-1 Accuracy (\%) on the ImageNet-1k classification and mIoU (\%) for Cityscapes, and ADE-20K segmentation tasks.}
\label{tab:ablations_step1} \vspace{-.5em}
\begin{subtable}[t]{0.365\textwidth} 
    \caption{Kernel sizes \& shortcut of MobileNet V2.}
    \label{table-mob2-shortcut}
    \resizebox{0.92\linewidth}{!}{
    \begin{tabular}{lcccc}
        \hline
        \scriptsize Shortcut 		& \scriptsize Kernel size		&  \scriptsize IN-1k (\%)	\\
        \hline
        &  \scriptsize	3$\times3$		&  \scriptsize	\textbf{68.67}		\\
         \scriptsize \checkmark		&	 \scriptsize 3$\times$3		&	 \scriptsize \reshl{71.76}{3.09}		\\
        \hline
        &	 \scriptsize 13$\times$13	&	 \scriptsize 53.98		\\
         \scriptsize \checkmark		&	 \scriptsize 13$\times$13	&	 \scriptsize \reshl{72.53}{18.55}		\\	
        \hline
    \end{tabular}
    }
\end{subtable}
\begin{subtable}[t]{.56\textwidth} 
\centering 
\caption{Kernel sizes \& re-parameterization on MobileNet V2.}
\label{table-mob2-reparam}
\resizebox{0.90\linewidth}{!}{
\begin{tabular}{lcccc}
    \hline
    \scriptsize Kernel size 					&	\scriptsize 3$\times$3 re-param		& \scriptsize IN-1k (\%)	&	\scriptsize Cityscapes (\%)	\\
    % \hline
    % 3$\times$3				&  N/A						&	71.76				&	72.31	\\
    %3$\times$3				&	\checkmark			&	todo				&	todo	\\
    \hline
    \scriptsize 9$\times$9			&						&	\scriptsize 72.67				&	\scriptsize 76.11	\\
    \scriptsize 9$\times$9			&	\scriptsize \checkmark			&	\scriptsize \reshl{73.09}{0.42}				&	\scriptsize \reshl{76.30}{0.19}	\\
    \hline
    \scriptsize 13$\times$13		&						&	\scriptsize 72.53				&	\scriptsize 75.67	\\
    \scriptsize 13$\times$13		&	\scriptsize \checkmark			&	\scriptsize \reshl{73.24}{0.71}				&	\scriptsize \reshl{76.60}{0.93}	\\			
    \hline
\end{tabular}
}
\end{subtable}
\vspace{1.5mm}

\begin{subtable}[t]{.46\textwidth} 
    \centering 
    \caption{Kernel sizes in the \emph{last stage} of MobileNet V2.}
    \label{table-mob2-smallfeature}
    \resizebox*{1.05\linewidth}{!}{
    \begin{tabular}{lcccccc}
    	\hline
    	Kernel size		& IN-1k (\%)	&	Cityscapes (\%)	& \#Params & FLOPs \\
    	\hline
    	3$\times$3		&	71.76				&	72.31	& 2.64M & 214.5M \\
    	7$\times$7		&	{72.00}				&	74.30	& 2.67M & 215.9M \\
    	9$\times$9		&	{71.83}				&	74.15	& 2.69M & 217.1M \\
    	13$\times$13	&	\reshl{71.97}{0.21}				&	\reshl{74.62}{2.31}	& 2.75M~\textcolor{lightgray}{(+4.2\%)} & 220.2M\textcolor{lightgray}{(+2.7\%)} \\		
    	\hline
    	\end{tabular}
    }
\end{subtable}
\begin{subtable}[t]{.52\textwidth} 
\centering 
\caption{Kernel sizes of different stages applying (a) (b) (c) in base model.}
\label{table-replknet-224}
\resizebox*{0.9\linewidth}{!}{
    \begin{tabular}{l|ccc|ccc}
    \hline
    \multirow{2}{*}{S1-S2-S3-S4}& \multicolumn{3}{c|}{IN-1k Classification} & \multicolumn{3}{c}{ADE20K Segmentation} \\
            &	Top-1 (\%)		& \makecell{Params}		&\makecell{FLOPs}	&	mIoU (\%)	& Params	& FLOPs	\\
    \hline
    3-3-3-3         &   82.11           &   71.8M       &   12.9G   &   46.05   &   104.1M  &   1119G\\
    7-7-7-7         &   82.73           &   72.2M       &   13.1G   &   48.05   &   104.6M  &   1123G\\
    13-13-13-13		&	83.02			&	73.7M		&	13.4G	&   48.35   &   106.0M  &   1130G\\ 
        \hline
    \end{tabular}
}
\end{subtable}
%#################################################
% \vspace{-5mm}
\end{table*}
Our roadmap to universal large-kernel ConvNets (UniRepLKNet, Fig.~\ref{fig-arch}) comprises four steps: \textbf{1}) We first explore why large kernel convolutions are not commonly used in modern ConvNets and propose 5 guidelines to make them more practical and evaluate their effectiveness~(\S~\ref{sec:step1}). \textbf{2}) We propose 4 guidelines for building a powerful and competitive large-kernel ConvNet architecture~(\S~\ref{sec:step2}). \textbf{3}) We propose to generalize the large-kernel ConvNets to multimodal understanding tasks~(\S~\ref{sec:step3}). \textbf{4}) Finally, we propose asymmetric large-kernel convolution to efficiently fuse multimodal features in contrast to cross-attention~(\S~\ref{sec:step4}).
\subsection{Step 1: Making Large Kernels Practical}
~\label{sec:step1}
\vspace{-4mm}
\subsubsection{Making Large Kernels Efficient}
~\label{sec:step1:sub1}
\textbf{The first reason why large kernels were rarely used is that they were believed to be computationally expensive} due to the quadratic increase in the number of parameters and FLOPs with kernel size.
However, we argue that this drawback can be significantly mitigated by using depth-wise (DW) convolutions \cite{mbv1,chollet2017xception}. As DW convolutions only consume a minor fraction of the total computational budget of a ConvNet, increasing the kernel sizes does not significantly make the model larger or slower. For example, as shown in Table~\ref{table-mob2-smallfeature}, increasing the kernel sizes of DW convolutions in MobileNet V2~\cite{mbv2} from 3$\times$3 to 13$\times$13 results in only a 2.7\% increase in FLOPs and 4.2\% increase in parameters, which is acceptable given the corresponding +2.31\% mIoU improvement in Cityscapes segmentation. The remaining 1$\times$1 convolutions dominate most of the complexity. 

One may be concerned that DW convolutions could be inefficient on modern parallel computing devices, such as GPUs. It is true for conventional DW 3$\times$3 kernels~\cite{mbv1,mbv2,zhang2018shufflenet}, as DW operations introduce a low ratio of computation \vs memory access cost~\cite{ma2018shufflenet}, which is not friendly to modern computing architectures. However, we find that as the kernel size increases, the computational density also increases. For example, in a DW 11$\times$11 kernel, each value loaded from the feature map can be used in up to 121 multiplications, while in a 3$\times$3 kernel, the number is only 9. Therefore, according to the roofline model~\cite{mbv2}, the actual latency should not increase as much as the FLOPs when the kernel size becomes larger.

The discussions above reveal that large-kernel DW convolutions can run faster with better implementation. In practice, we propose a block-wise (inverse) \emph{implicit GEMM} algorithm to replace the original operator.~\footnote{For PyTorch, we have released the efficient implementation at \url{https://github.com/AILab-CVC/UniRepLKNet} as a plug-and-play module.} Table~\ref{table-speed-kernelsize} shows that our implementation is significantly more efficient compared to the PyTorch baseline. 

Therefore, we propose our first guideline as follows.
\noindent{\textbf{Guideline 1: use depth-wise large-kernel convolution with proper operator-level implementation.}} 

\subsubsection{Making Large kernels Effective}~\label{sec:step1:sub2}
\textbf{The second reason why large kernels were rarely used is that they were believed to harm the model's performance}. However, we argue that large kernels are not harmful; they were simply not used properly. We propose three guidelines to use large kernels correctly in modern ConvNets.

\noindent{\textbf{Guideline 2: identity shortcut is vital, especially for networks with very large kernels.}}
To demonstrate this, we use \emph{MobileNet V2}~\cite{mbv2} for benchmarking, since it heavily employs DW layers and has two published variants (with or without shortcuts). For the large-kernel counterparts, we simply replace all the DW 3$\times$3 kernels with 13$\times$13. All the models are trained on ImageNet with identical training configurations for 100 epochs (see Appendix A for details). Table~\ref{table-mob2-shortcut} shows that large kernels improve the accuracy of MobileNet V2 with shortcuts from 71.76\% to 72.53\%. However, for the model without shortcuts, large kernels reduce the accuracy to only 53.98\%. We explain this phenomenon from a perspective similar to \cite{veit2016residual}: shortcuts make the model an implicit ensemble of numerous models with different receptive fields (RFs), allowing it to benefit from a much larger maximum RF without losing the ability to capture small-scale patterns.

\begin{figure*}[t]
    \centering
    \includegraphics[width=0.98\linewidth]{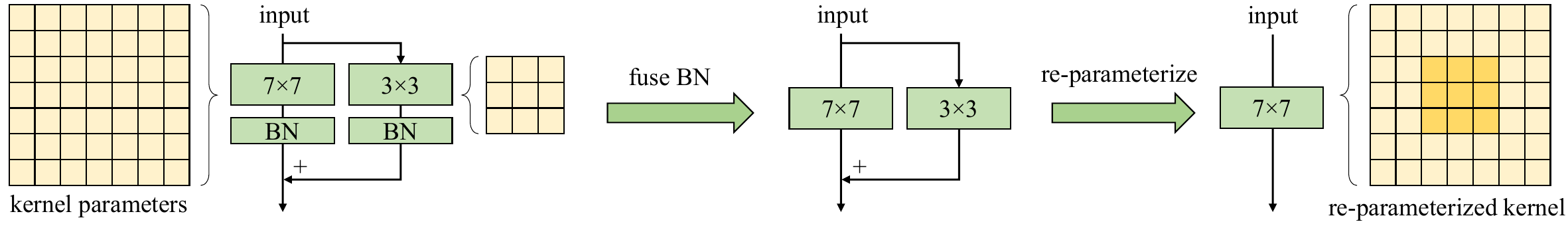}
    \vspace{-2.5mm}
    \caption{An example of re-parameterizing a small kernel (\eg, 3$\times$3) in Table~\ref{table-mob2-reparam} into a large one (\eg, 7$\times$7). We use the structural re-parameterization as previous practices \cite{ding2019acnet,ding2021repvgg}. }
    \label{fig:kernel_reparam}
    \vspace{-3.5mm}
\end{figure*}

\noindent{\textbf{Guideline 3: re-parameterizing large kernels with small kernels improves the performance}.} To better understand the effect of the aforementioned ensemble of different RFs, we explore whether using small kernels to produce a bigger ensemble of more different RFs improves the performance. Specifically, we replace the 3$\times$3 layers of \emph{MobileNet V2} with 9$\times$9 and 13$\times$13, and optionally adopt the \emph{Structural Re-parameterization}~\cite{ding2021repvgg,ding2021repmlpnet,ding2019acnet} methodology to add small kernels without altering the inference structure of the resultant model. Specifically, we construct a 3$\times$3 layer parallel to the large-kernel layer and add their outputs together after the \emph{Batch normalization (BN)}~\cite{ioffe2015batch} layers (Fig.~\ref{fig:kernel_reparam}). After training, we merge the small kernel and BN parameters into the large kernel, so the resultant model will be mathematically equivalent to the training model but no longer has small kernels. Table~\ref{table-mob2-reparam} shows that directly increasing the kernel size from 9 to 13 reduces accuracy, while re-parameterization addresses this issue. 
	
We then transfer the ImageNet-trained models to semantic segmentation with DeepLabv3+~\cite{chen2018encoder} on Cityscapes~\cite{cityscapes}. We only replace the backbone and keep all the default training settings of MMSegmentation~\cite{mmseg2020}. The observation is similar to that on ImageNet: 3$\times$3 re-parameterization improves the mIoU of the 9$\times$9 model by +0.19\% and the 13$\times$13 model by +0.93\%; with re-parameterization, increasing the kernel size from 9 to 13 no longer degrades performance on either ImageNet or Cityscapes.

\noindent{\textbf{Guideline 4: large kernels (\eg, 13\bm{$\times$}13) are effective even on small feature maps (\eg, 7\bm{$\times$}7)}}. To validate it, We enlarge the DW convolutions in the \emph{last stage} of \emph{MobileNet V2} to 7$\times$7 or 13$\times$13, hence the kernel size is on par with or even larger than feature map size (7$\times$7 by default). We apply re-parameterization to the large kernels as suggested by Guideline 3. Table~\ref{table-mob2-smallfeature} shows although convolutions in the last stage already involve very large receptive field, further increasing the kernel sizes still leads to performance improvements, especially on downstream tasks such as \emph{Cityscapes}. 

\begin{figure}[t]
    \centering
    \includegraphics[width=\linewidth]{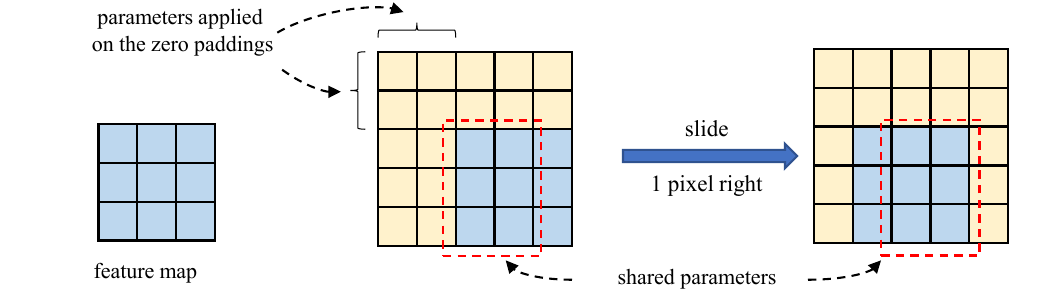}
    \vspace{-0.15in}
    \caption{Illustration to convolution with small feature map and large kernel. Two outputs at adjacent locations only share a part of kernel weights. Translational equivariance does not strictly hold.}
    \label{fig-large-kernel-small-feature}
    \vspace{-5mm}
\end{figure}

\noindent \textbf{Remark.} When kernel size becomes large, notice that the translational equivariance of CNNs does not strictly hold. As illustrated in Fig.~\ref{fig-large-kernel-small-feature}, two outputs at adjacent spatial locations share only a fraction of the kernel weights, \ie, and are transformed by different mappings. The property also agrees with the ``philosophy'' of \emph{ViTs} -- relaxing the symmetric prior to obtaining more capacity. Interestingly, we find 2D \emph{Relative Position Embedding (RPE)}~\cite{shaw2018self,bello2019attention}, which is widely used in the transformer community, can also be viewed as a large depth-wise kernel of size $(2H-1)\times(2W-1)$, where $H$ and $W$ are feature map height and width respectively. Large kernels not only help to learn the relative positions between concepts but also encode the \emph{absolute position} information due to \emph{padding effect}~\cite{kayhan2020translation}. 

\subsubsection{Evaluating Large-kernels ConvNets}~\label{sec:step1:sub3}
\textbf{The third reason to abandon large kernels, even though the large-kernel ConvNet is designed properly, is that its ImageNet accuracy looks no better than a small-kernel ConvNet.} However, Table~\ref{table-mob2-reparam} (after \emph{re-param}) shows increasing the kernel size of MobileNet V2 from 3$\times$3 to 9$\times$9 improves the \emph{ImageNet} accuracy by 1.33\%, but the \emph{Cityscapes} mIoU by 3.99\%. Such a phenomenon indicates that models of similar ImageNet scores could have very different capabilities in downstream tasks. 

\noindent \textbf{Remark.} What causes the phenomenon? First, large kernel design significantly increases the \emph{Effective Receptive Fields (ERFs)} \cite{erf}, as shown in Figure~\ref{fig:erf}. Numerous works have demonstrated ``contextual'' information, which implies large ERFs, is crucial in many downstream tasks like object detection and semantic segmentation \cite{peng2017large,long2015fully,yu2017dilated,wang2020deep,yu2015multi}. Second, We deem another reason might be that large kernel design contributes more shape biases to the network. Briefly speaking, ImageNet pictures can be correctly classified according to either texture or shape, as proposed in \cite{geirhos2018imagenet,brendel2019approximating}. However, humans recognize objects mainly based on shape cue rather than texture, therefore a model with stronger shape bias may transfer better to downstream tasks. A recent study \cite{tuli2021convolutional} points out ViTs are strong in shape bias, which partially explains why ViTs are super powerful in transfer tasks. In contrast, conventional CNNs trained on ImageNet tend to bias towards texture \cite{geirhos2018imagenet,brendel2019approximating}. 
Fortunately, we find that simply enlarging the kernel size in ConvNets can effectively improve the shape bias, which means a large-kernel model makes decisions based more on the shapes of objects than the textures.

Therefore, we propose another guideline regarding the evaluation of large-kernel ConvNets.

\noindent{\textbf{Guideline 5: evaluate large-kernel ConvNets by the performance of downstream tasks.}} 

\begin{figure*}[ht]
\begin{center}
    \includegraphics[width=0.98\linewidth]{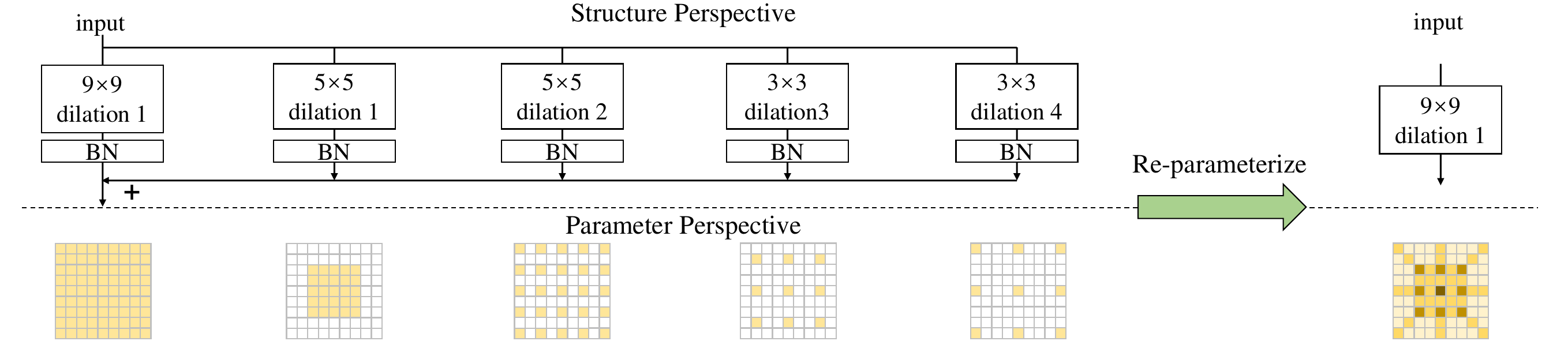}
    \vspace{-1mm}
    \caption{Dilated Reparam Block~(\S~\ref{sec:method:micro}) uses dilated small-kernel conv layers to enhance a non-dilated large-kernel layer. Such dilated layers are equivalent to a non-dilated conv layer with a larger sparse kernel, as shown from the parameter perspective so that the whole block can be equivalently transformed into a single large-kernel conv. This example shows $K$=9, and we may use more dilated layers for larger $K$.}
    \label{fig-reparam}
    \vspace{-2mm}
\end{center}
\end{figure*}
\subsection{Step 2: Designing a Competitive Large-Kernel Architecture}
~\label{sec:step2}
As discussed above, we have been aware of five basic guidelines for making large kernels practical and then seek to explore how to design a powerful and competitive large-kernel architecture. 

\textit{We first construct a vanilla architecture as a baseline to verify which design choices work well with large kernels.}

\noindent\textbf{Vanilla architecture}. As a common practice, the main body of the model is split into four stages connected by downsampling blocks. Specifically, the first downsampling block uses two stride-2 3$\times$3 convolutional layers to transform the raw input into $C$-channel feature maps, where $C$ is an architectural hyper-parameter. The other three downsampling blocks each use one stride-2 3$\times$3 conv layer performing 2$\times$ channel expansion so that the numbers of channels in the four stages are $C$, $2C$, $4C$, and $8C$, respectively. A stage comprises blocks whose vanilla design resembles ConvNeXt, \ie, a \emph{depthwise (DW)} conv layer and a \emph{Feed-Forward Network (FFN)} with GRN unit~\cite{woo2023convnext}. However, we use Batch Normalization (BN) instead of Layer Normalization~\cite{ba2016layer} after the conv layer as BN can be equivalently merged into the conv layer to eliminate its inference costs. We use another BN after the FFN, which can also be equivalently merged into the preceding layer (\ie, the second linear layer in FFN). The numbers of such blocks in the four stages are denoted by $\bm{N}\coloneqq (N_1,N_2,N_3,N_4)$. Following ConvNeXt-T, the vanilla architecture uses $C=96$ and $\bm{N}=(3,3,9,3)$. By default, the last three stages use DW 13$\times$13 as the convolutional layer, and the first stage uses DW 3$\times$3.

\begin{figure}[ht]
\begin{center}
    \includegraphics[width=1.0\linewidth]{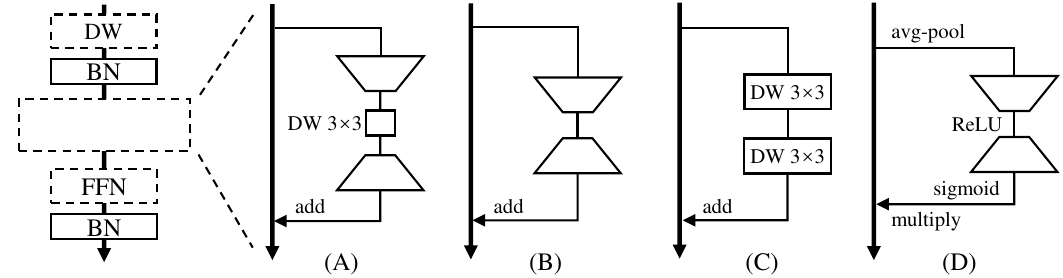}
    \vspace{-0.15in}
    \caption{Options of the extra structures to increase the depth.}
    \label{fig-se}
    \vspace{-0.15in}
\end{center}
\end{figure}

\noindent\textbf{Experimental settings and metrics}. According to Guideline 5, large-kernel ConvNets should be evaluated on downstream tasks, as their full potential may not be accurately reflected by ImageNet accuracy alone. Therefore, in addition to reporting the ImageNet-1K accuracy after 100 epochs of training, we transfer the trained model with UPerNet~\cite{xiao2018unified} to ADE20K to evaluate its performance on semantic segmentation. We report the single-scale mIoU after a 160k-iteration standard finetuning process~\cite{mmseg2020}. Besides the parameters and FLOPs, we test the actual throughput on an A100 GPU with a batch size of 128 and an input resolution of 224$\times$224, measured in images per second (img/s). See the Appendix for detailed configurations.

\textit{We then discuss and verify a series of design choices made in a large-kernel ConvNets. In the following, we summarize our conclusion as a guideline and present the experimental evidence.}

\subsubsection{Block Design for Large-Kernel ConvNets}\label{sec:step2:sub1}
\begin{table*}[ht]
% \vspace{-2mm}
\centering
\caption{\textbf{Comparisons among design choices of Step 2}. We report the Top-1 Accuracy (\%) and mIoU (\%) on the ImageNet-1k and ADE-20K datasets.}
\label{tab:ablations_step2} \vspace{-.5em}
\begin{subtable}[t]{.48\textwidth} 
    \centering 
    \caption{Different efficient extra structures to increase the depth.}
    \label{table-guide1}
    \resizebox{1.0\linewidth}{!}{
    \begin{tabular}{lccccc}
        \hline
    Extra structure	       & Params    &  FLOPs & Img/s  & Acc &	mIoU\\
        \hline
        None               & 31.3M   & 4.92G   &   1954    &   81.2    &   45.1 \\
        (A) Bottleneck          & 32.9M   & 5.18G   &   1716    &   81.5    &   46.3\\
        (B) Two 1$\times$1      & 32.9M   & 5.17G   &   1745    &   81.3    &   46.2\\
        (C) Two DW 3$\times$3   & 31.4M   & 4.96G   &   1659    &   81.3    &   45.4\\
        (D) SE Block            & 32.9M & 4.92G   &   1863      &   \textbf{81.6}    &   \textbf{46.5} \\
        \hline
    \end{tabular}
    }
\end{subtable}
\begin{subtable}[t]{.51\textwidth} 
	\centering 
	\caption{Different forms of Structural Re-parameterization on the 13$\times$13 conv layers based on the Vanilla architecture.}
	\label{table-reparam}
	\resizebox{1.0\linewidth}{!}{
		\begin{tabular}{lccccc}
			\hline
			Re-param	            &   $\bm{k}$      &   $\bm{r}$   &      Acc &	mIoU\\
			\hline
			None                    &   N/A         &   N/A         &   81.44$\pm$0.04       &   45.78$\pm$0.05\\          
			Dilated Reparam  &   5,7,3,3,3 &  1,2,3,4,5            &   81.63$\pm$0.02    &   \textbf{46.37}$\pm$0.10 \\    
			Same kernel size    &   5,7,3,3,3 &  1,1,1,1,1          &   81.55$\pm$0.01    &   46.07$\pm$0.07\\             
			Same eq kernel size        &   5,13,7,9,11 &  1,1,1,1,1 &   81.59$\pm$0.02    &   46.17$\pm$0.04     \\       
			\hline
	\end{tabular}}
\end{subtable}

\vspace*{1.5mm}

\begin{subtable}[t]{.48\textwidth} 
    \centering 
    \caption{Different kernel sizes in the four stages denoted by S1 - S4.}
    \label{table-guide3}
    \resizebox{1.0\linewidth}{!}{
    \begin{tabular}{lcccccccc}
    \hline
  S1  & S2   & S3  & S4  &  Params & FLOPs  & Img/s & Acc &  mIoU\\
    \hline
3& 13 & 13 & 13 & 32.9M & 4.92G   &   1863    &   81.6    &   \textbf{46.5} (42.4)    \\ 
\hline
    3& 11 & 11 & 11 &  32.6M   & 4.86G  &   1876      & 81.6      &   45.5 (41.9)    \\
    3&  3& 13 & 13    & 32.8M   &   4.85G   &   2006    & 81.7  &   46.1\\
      3& 13 &  3& 13    & 32.4M   &   4.81G   &   2015    & 81.6  &   45.9\\
      3& 13 & 13 & 3     & 32.5M   &   4.90G   &   1884    & 81.4  &   45.8\\                             
     \hline
     3& 15 & 15 & 15 &   33.3M  &  4.99G    &   1851    &    81.7   &   45.9 (\textbf{42.7})    \\
     13 & 13 & 13 & 13 &    33.0M   &5.06G  &   1547    &   81.6    & 44.9 (42.4)  \\
     \hline
\end{tabular}
    }
\end{subtable}
    \centering
\begin{subtable}[t]{.5\textwidth} 
    \centering 
    \caption{Different numbers of Large-Kernel and Small-Kernel Blocks in Stage 3.}
    \label{table-guide4}
    \resizebox{1.0\linewidth}{1.1cm}{
    \begin{tabular}{lcccccccc}
    \hline
  $N_3$   & LarK & SmaK     &  Params & FLOPs  & Img/s & Acc &  mIoU\\
    \hline
    9   &   9   &   0        & 32.9M & 4.92G   &   1863    &   81.6    &   \textbf{46.5}   \\
    \hline
    27  &   27  &   0         &   56.7M   &9.31G  &   1145   &   82.3   &   49.0\\
    27  &   14   &   13,  3$\times$3  &   55.9M   &9.15G  &   1229  &   82.3   &   48.8\\
    27  &   9   &   18,   3$\times$3  &   55.6M   &9.10G  &   1264   &   82.3   &   48.8\\
    \hline
    27  &   9   &   18, w/o 3$\times$3    &   55.5M   &9.08G  &   1289   &   82.2   &   47.8\\
    \hline
\end{tabular}
    }
\end{subtable}
    \centering
%#################################################
\vspace{-2mm}
\end{table*}
\noindent\textbf{Guideline 6: regarding the block design, use efficient structures that perform both inter-channel communications and spatial aggregations to increase the depth}. We first aim to enhance the model's representational capacity by universally incorporating structures that provide nonlinearity and efficient trainable transformations. To achieve this, we employ a bottleneck consisting of a 1$\times$1 conv that reduces the channels to 1/4, followed by a DW 3$\times$3 conv, and another 1$\times$1 conv to expand the channels back (Fig.~\ref{fig-se}). BN and ReLU are applied after each conv layer as standard practice. As shown in Table~\ref{table-guide1}, this approach improves performance with acceptable overhead (+1.2 mIoU with a 12\% slowdown). Performance degrades when we remove the DW 3$\times$3 conv, leaving only two 1$\times$1 conv layers, or when we replace the bottleneck structure with two DW 3$\times$3 layers. This indicates that effective structures require both spatial aggregation transformations and channel mixing. Motivated by this, considering that SE Block~\cite{hu2018squeeze} elegantly realizes both transformations in a more efficient way (\ie, global average pooling and nonlinear mapping of the pooled vectors), we try it also with 1/4 channel reduction and observe a better performance and higher throughput. 

\textit{We therefore use the SE Block as a substructure of our block design in the following explorations.}

\subsubsection{Micro Design with Structural Re-parameterization for Large-Kernel ConvNets}~\label{sec:method:micro}

\noindent\textbf{Guideline 7: use dilated small kernels to re-parameterize a large kernel}. We then explore the micro (\ie, layer-level) design for large-kernel ConvNet. According to Guideline 3, we should use a parallel small-kernel conv together with a large-kernel layer, as the former helps capture the small-scale patterns during training. Former discussions, however, have primarily focused on simple methods that make large kernels more practical and on explaining the underlying mechanism, rather than offering a competitive solution for building a powerful large-kernel architecture. While we now aim at the latter goal, we recognize that simply using a small kernel to re-parameterize a large kernel may not be optimal, as both capture dense patterns despite their different receptive fields. More than that, we reckon that, except for small-scale patterns, enhancing the large kernel's capability to capture \emph{sparse patterns} (\ie, a pixel on a feature map may be more related to some distant pixels than its neighbors) may yield features of higher quality. The need to capture such patterns exactly matches the mechanism of dilated convolution - from a sliding-window perspective, a dilated conv layer with a dilation rate of $r$ scans the input channel to capture spatial patterns where each pixel of interest is $r-1$ pixels away from its neighbor. Therefore, we use dilated conv layers parallel to the large kernel and add up their outputs. 

To eliminate the inference cost of the extra dilated conv layers, we propose to equivalently transform the whole block into a single non-dilated conv layer for inference. Since \emph{ignoring pixels of the input is equivalent to inserting extra zero entries into the conv kernel}, \emph{a dilated conv layer with a small kernel can be equivalently converted into a non-dilated (\ie, $r=1$) layer with a sparse larger kernel}. Let $k$ be the kernel size of the dilated layer, by inserting zero entries, the kernel size of the corresponding non-dilated layer will be $(k-1)r+1$, which is referred to as the \emph{equivalent kernel size} for brevity. We further note that such transformation from the former kernel $\mathrm{W}\in\mathcal{R}^{k\times k}$ to the latter $\mathrm{W}^\prime\in\mathcal{R}^{((k-1)r+1)\times ((k-1)r+1)}$ can be elegantly realized by a transpose convolution with a stride of $r$ and an identity kernel $\mathrm{I}\in\mathcal{R}^{1\times1}$, which is scalar 1 but viewed as a kernel tensor.~\footnote{We showcase a single-channel conv and it is easy to generalize the transformation to multi-channel cases. See the Appendix for details.} With PyTorch-style pseudo code, that is
\begin{equation}\label{eq-merge}
    \mathrm{W}^\prime = \mathtt{conv\_transpose2d}(\mathrm{W}, \mathrm{I}, \text{stride}=r) \,.
\end{equation}
The equivalency can be easily verified - given an arbitrary $\mathrm{W}\in\mathcal{R}^{k\times k}$ and an arbitrary input channel, a convolution with $\mathrm{W}$ and a dilation rate $r$ always yields identical results to a non-dilated convolution with $\mathrm{W}^\prime$.~\footnote{In common cases where the output and input have the same size, \ie, the padding of the former is $\frac{k-1}{2}$, note the padding of the latter should be $\frac{(k-1)r}{2}$ since the size of the equivalent sparse kernel is $(k-1)r+1$.}

Based on such equivalent transformations, we propose a novel module named \emph{Dilated Reparam Block}, which uses a non-dilated small-kernel and multiple dilated small-kernel layers to enhance a non-dilated large-kernel conv layer. Its hyper-parameters include the size of large kernel $K$, the size of parallel conv layers $\bm{k}$, and the dilation rate $\bm{r}$. The shown case (Fig.~\ref{fig-reparam}) with four parallel layers is denoted by $K$=9, $\bm{r}$=(1, 2, 3, 4), $\bm{k}$=(5, 5, 3, 3). For a larger $K$, we may use more dilated layers with larger kernel sizes or dilation rates. The kernel sizes and dilation rates of the parallel branches are flexible, and the only constraint is $(k-1)r+1 \leq K$. For example, with $K$=13 (the default setting in our experiments), we use five layers with $\bm{k}$=(5, 7, 3, 3, 3), $\bm{r}$=(1, 2, 3, 4, 5), so the equivalent kernel sizes will be (5, 13, 7, 9, 11), respectively. To convert a Dialted Reparam Block into a large-kernel conv layer for inference, we first merge every BN into the preceding conv layer, convert every layer with dilation $r>1$ with function~\ref{eq-merge}, and add up all the resultant kernels with appropriate zero-paddings. For example, the layer in Fig.~\ref{fig-reparam} with $k$=3, $r$=3 is converted into a sparse 7$\times$7 kernel and added to the 9$\times$9 kernel with one-pixel zero paddings on each side. 
For a fair comparison with Dilated Reparam Block, we try two variants with the same numbers of parallel branches composed of non-dilated layers with \textbf{A)} the same kernel sizes or \textbf{B)} the same equivalent kernel sizes. For our default setting of $K$=13, $\bm{r}$=(1, 2, 3, 4, 5), $\bm{k}$=(5, 7, 3, 3, 3), the kernel sizes of the five branches will be $\bm{k}$=(5, 7, 3, 3, 3) or (5, 13, 7, 9, 11) for the two variants, respectively. All the models end up with the same inference structure but the training structures differ. Table~\ref{table-reparam} shows lower performance of variants, suggesting that large kernel benefits from the parallel dilated conv layers' abilities to capture sparse patterns, rather than merely the extra small kernels (variant A) or the combination of different receptive fields (variant B). 

\textit{We therefore use Dilated Reparam Block by default.}~\footnote{While this paper describes the architecture, using a $K$$\times$$K$ ($K$$\geq$9) conv means a $K$$\times$$K$ Dilated Reparam Block, unless otherwise noted.}

\subsubsection{Kernel Size of Large-Kernel ConvNets}~\label{sec:step2:kernel}

\noindent\textbf{Guideline 8: decide kernel size according to the downstream task and usually use large kernels in middle- and high-level layers}. As introduced above, the vanilla architecture uses 3$\times$3 conv in the first stage and 13$\times$13 in the last three stages. Table~\ref{table-guide3} shows that replacing the large kernels in the last three stages with 3$\times$3 or changing $K$ from 13 to 11 degrades the models, especially in the ADE20K mIoU, which highlights the significance of large kernels. Interestingly, using 13$\times$13 in Stage 1 or enlarging $K$ from 13 to 15 makes almost no difference in the ImageNet accuracy but reduces the ADE20K mIoU.

\noindent\textbf{Remark}. We argue that this phenomenon does not mean larger kernels result in lower feature quality. It is due to the structural priors of UPerNet, which takes the features extracted by the low-level layers of the backbone and assumes they should only encode local information so that combining them with the high-level features extracted from the last layers of the backbone results in better segmentation. With larger kernels in lower stages, the low-level features are no longer confined to small local areas, so the UPerNet benefits less from combining them with the high-level features. We verify this explanation by making the UPerNet only use the high-level features (\ie, outputs of Stage 4) to evaluate the quality of the eventual features alone. Under this setting, $K$=15 delivers the best mIoU (42.7), the model with large kernels in Stage 1 performs as well as the baseline (42.4), and $K$=11 performs the worst (41.9). Such observations confirm that large kernels, even when they are used inappropriately, \emph{do not damage the feature quality} of ConvNet but \emph{merely make the low-level features less favorable for certain downstream models that require local low-level features}, suggesting we should decide the kernel size according to the specific downstream tasks and framework. 

\textit{Considering this, we employ 13$\times$13 kernels in the middle- and high-level stages by default.}

\subsubsection{Scaling Rule of Large-Kernel ConvNets}~\label{sec:step2:scaling}

\noindent\textbf{Guideline 9: while scaling up the depth, the added blocks should use small kernels.} The scaling rule of existing large-kernel ConvNets follows the traditional ConvNets, \ie, stacking more large kernels to build up a deeper model, but we argue that a large-kernel ConvNet may not benefit from more large kernels. In this group of experiments (Table~\ref{table-guide4}), we scale up $N_3$ from 9 to 27, following ConvNeXt-S~\cite{liu2022convnet}. Considering that nine 13$\times$13 blocks may have already built up sufficient receptive field, we examine if the added blocks should also use large kernels. Specifically, we refer to the block with a Dilated Reparam Block as the \emph{Large Kernel Block (LarK Block)} and name a block that uses a DW 3$\times$3 conv as a \emph{Small Kernel Block (SmaK Block)} so that there are 3 SmaK Blocks in Stage 1 and 3/9/3 LarK Blocks in Stage 2/3/4 of the shallow model. While scaling up the depth of Stage 3, we tried the following options. \textbf{A)} All of the 27 blocks are LarK Blocks. \textbf{B)} We interleave SmaK with LarK Blocks so that Stage 3 has 14 LarK Blocks and 13 SmaK Blocks. \textbf{C)} We place two SmaK Blocks after a LarK Block so that the resultant model will have the same 9 LarK Blocks as before but 18 extra SmaK Blocks. \textbf{D)} We remove the DW 3$\times$3 layers in SmaK Blocks. Table~\ref{table-guide4} shows that scaling up the depth brings significant improvements, which is expected, and 9 LarK Blocks are sufficient. Though 27 LarK Blocks perform slightly better in the ADE20K mIoU, the inference speed is observably slowed down. Besides, the model without 3$\times$3 conv in SmaK Blocks shows significantly lower mIoU with only minor improvements in the throughput, suggesting such small kernels in SmaK Blocks are useful while scaling up the depth of large-kernel ConvNet as they increase the abstract hierarchy of spatial patterns, though they may not effectively enlarge the ERF~\cite{ding2022scaling,erf}. This observation supports our motivation to decouple the effects of conv layers in enlarging the ERF and extracting more complicated spatial patterns, as discussed in Sec.~\ref{sec:intro}.

\subsubsection{Architectural Specifications}\label{sec-arch-spec}

Following our proposed guidelines, we instantiate a series of models (Table~\ref{table-instances}). For a fair comparison with ConvNeXt V2~\cite{woo2023convnext}, UniRepLKNet-A/F/P/N follows its configurations. We scale up the depth to build UniRepLKNet-T/S and scale up the width to construct UniRepLKNet-S/B/L/XL/H. 

\begin{table}[ht]
\caption{Architectural hyper-parameters of UniRepLKNet instances, including the number of blocks in the four stages $N_1, N_2, N_3, N_4$ and channels $C$ of the first stage. Stage 1 uses SmaK Blocks, and Stages 2 and 4 use LarK Blocks only. For Stage 3, \eg, ``9 + 18'' means 9 LarK Blocks and 18 SmaK Blocks.}
\label{table-instances}
\begin{center}
\resizebox{1.0\linewidth}{!}{
    \tiny
    \begin{tabular}{lcccccccc}
        \hline
       &$N_1$ & $N_2$ & $N_3$ & $N_4$ & $C$ &  Params    \\
        \hline
        UniRepLKNet-A & 2   & 2     & 6 + 0     & 2     &   40   &  4.4M\\
        UniRepLKNet-F & 2   & 2     & 6 + 0     & 2     &   48   &  6.2M\\
        UniRepLKNet-P & 2   & 2     & 6 + 0     & 2     &   64   &  10.7M\\
        UniRepLKNet-N & 2   & 2     & 8 + 0     & 2     &   80   &  18.3M\\
        UniRepLKNet-T & 3   & 3     & 9 + 9     & 3     &   80   &  31.0M\\
        UniRepLKNet-S & 3   & 3     & 9 + 18    & 3     &   96   &  55.6M\\
        UniRepLKNet-B & 3   & 3     & 9 + 18    & 3     &   128   & 97.9M\\
        UniRepLKNet-L & 3   & 3     & 9 + 18    & 3     &   192   & 218.3M\\
        UniRepLKNet-XL & 3   & 3     & 9 + 18    & 3     &   256  & 386.4M\\
        UniRepLKNet-H & 3   & 3     & 9 + 18    & 3     &   480  & 1.4B\\
        \hline
    \end{tabular}
}
\end{center}
\vspace{-3mm}
\end{table}
\subsection{Step 3: Generalizing Large-Kernel ConvNets to Multiple Modalities}
~\label{sec:step3}
To utilize the universal perception ability of UniRepLKNet, we preprocess the data of different modalities into $B\times C^\prime\times H\times W$ embedding maps, where $B$ is the batch size and $C^\prime$ is determined by the modality, and configure the input channel of the first layer of UniRepLKNet to $C^\prime$. For simplicity, the other parts of the models are the same as the UniRepLKNet initially designed for the image without any modality-specific customization. 

\noindent\textbf{Time-series}. Let $L$ and $D$ be the length and dimensions of a time-series sequence $\boldsymbol{x}_T\in \mathbb{R}^{B\times L \times D}$, we adopt the embedding layer in Corrformer~\cite{wu2023interpretable} to split it into $n$ nodes then project it into a latent space $\mathbb{R}^{Bn \times L \times D^\prime}$ ($D^\prime$ and $n$ are configurable hyper-parameters of the embedding layer). Then we simply reshape it into a single-channel embedding map:
\begin{equation}\label{eq:time:token}
    \begin{aligned}
         \boldsymbol{x}_T \in\mathbb{R}^{B\times L \times D } & \rightarrow \mathbb{R}^{Bn \times L \times \frac{D}{n}} \rightarrow \mathbb{R}^{Bn \times L \times D^\prime} \\ &\rightarrow \mathbb{R}^{Bn\times 1 \times H \times W }\, \text{s.t.} \,\,HW=LD^\prime.
    \end{aligned}
\end{equation}
% --------------------------------------------------------------------------
\textbf{Audio}. Let $T$ and $F$ be the numbers of time frames and frequency bins, we use $\boldsymbol{x}_A \in \mathbb{R}^{B\times T\times F}$ to represent audio data. A sample is seen as a $1\times T\times F$ embedding map that resembles a single-channel image so $C^\prime$=1, $H$=$T$, $W$=$F$.
\begin{equation}
    \boldsymbol{x}_A \in \mathbb{R}^{B\times T \times F} \rightarrow \mathbb{R}^{B\times 1 \times T \times F}.
    \label{eq:audio:token}
\end{equation}
% --------------------------------------------------------------------
\textbf{Point cloud}. Assume a sample comprises \(P\) points each represented by the X/Y/Z coordinates, we use a series of conv layers to generate three-view projections~\cite{zhang2023meta}. We configure the resolution of the generated projections to be 224 so that $H$=$W$=224, $C^\prime$=3.
\begin{equation}
    \boldsymbol{x}_P \in \mathbb{R}^{B\times P \times 3} \rightarrow  \mathbb{R}^{B\times 3 \times 224 \times 224 }\,.
    \label{eq:pcd:token}
\end{equation}
% --------------------------------------------------------------------
\textbf{Video}. We represent a video as $N_F$ frames and each frame is a $3\times h \times w$ image. We reshape it by merging the frame dimension into the height and width dimensions so that we obtain a representation that can be viewed as a single image created by laying out (\ie, concatenating) the $N_F$ frames. For example, in our experiments, we have $N_F$=16 and $h$=$w$=224 so that $H$=$W$=896. Generally,
\begin{equation}
    \label{eq:video:token}
    \boldsymbol{x}_V\in\mathbb{R}^{B\times N_F\times3\times h \times w}\rightarrow \mathbb{R}^{B\times 3\times H\times W}\text{s.t.}\frac{HW}{hw}=N_F\,.
\end{equation}

\subsection{Step 4: Fusing Multimodal Features with Large Kernel Convolution}~\label{sec:step4}
In addition to extracting features, we further explore large-kernel convolution to fuse multimodal features as cross-attention~\cite{vaswani2017attention}. Inspired by the flexibility of asymmetric convolution in fusing features of diverse shapes~\cite{han2024asymmetric}, we propose the asymmetric large-kernel convolution to broadly fuse features across diverse shapes and modalities. As cross-attention mechanism to fuse two features of \(\bm{X}\) and \(\bm{Y}\), where  \(\bm{X} \in \mathbb{R}^{L_1\times D}, \bm{Y} \in \mathbb{R}^{L_2\times D}\), \(L_1\) and \(L_2\) denote the length of a sequence of tokens, \(D\) denotes the feature dimension (Note that the feature map \(\bm{X} \in \mathbb{R}^{L_1\times D}\) can be easily reshaped as \(\bm{X} \in \mathbb{R}^{H\times W \times C}\)).

The asymmetric large-kernel convolution uses one feature map as the convolutional kernel to convolve another feature map, allowing for dynamic and context-aware fusion of multimodal features. Specifically, the convolution operation is performed by treating \(\bm{Y}\) as the convolutional kernel that is applied to \(\bm{X}\). In this setup, each element of \(\bm{Y}\) serves as a dynamic filter that modulates \(\bm{X}\) according to its contextual information. The output feature map \(\bm{Z}\) can be expressed as:
\[
\bm{Z}_{i,j} = \sum_{k=1}^{L_2} \bm{X}_{i+k-1} \cdot \bm{Y}_{k,j},
\]
where \(\bm{Z}_{i,j}\) represents the correlation between \(\bm{X}\) starting at position \(i\) with the filter defined by \(\bm{Y}_{j}\). This approach allows \(\bm{X}\) to be dynamically influenced by the patterns in \(\bm{Y}\), facilitating an adaptive and effective fusion of the two feature maps. It efficiently captures the intrinsic correlation between the features, making it a computationally efficient alternative for multimodal feature fusion tasks.

\section{Experiments}~\label{sec:exp}
\vspace{-5mm}
\subsection{Experiments for Visual Recognition}
\begin{table}[ht!]
    \centering
    \renewcommand\arraystretch{0.89}
    \setlength{\tabcolsep}{0.8mm}
    \footnotesize
    \caption{\textbf{ImageNet classification}. Throughput is tested with an A100 GPU and batch size of 128. ``T/C'' denote transformer/ConvNet. ``$^\ddagger$" indicates ImageNet-22K~\cite{deng2009imagenet} pretraining.}
    \vspace{-0.1in}
    \begin{tabular}{l|c|c|c|c|c|c}
\hline
    \multirow{2}{*}{Method} & \multirow{2}{*}{Type} & Input & Params & FLOPs & Throughput & Acc \\
    & & size &(M)&(G)&(img/s)&(\%) \\ 
    \hline
    \rowcolor{gray!20}
    \textbf{UniRepLKNet-A}   &   C   &   $224^2$     &   4.4    &   0.6    &   \textbf{5942}    &   \textbf{77.0}    \\
    \rowcolor{gray!20}
    \textbf{UniRepLKNet-F}   &   C   &   $224^2$     &   6.2    &   0.9    &   \textbf{5173}    &    \textbf{78.6}       \\
    ConvNeXt V2-A~\cite{woo2023convnext}   &   C   &   $224^2$     &   3.7    &   0.5    &   5054    &   76.2        \\
    FastViT-T8~\cite{vasu2023fastvit}      &   T   &   $256^2$     &   3.6    &   0.7    &   5025    &   75.6    \\
    ConvNeXt V2-F~\cite{woo2023convnext}   &   C   &   $224^2$     &   5.2    &   0.8    &   4329    &   78.0        \\
    \hline
    \rowcolor{gray!20}
    \textbf{UniRepLKNet-P}   &   C   &   $224^2$   & 10.7 & 1.6    &   \textbf{3949}    &    \textbf{80.2}    \\
    FastViT-T12~\cite{vasu2023fastvit}      &   T   &   $256^2$    &   6.8     &   1.4    &   3407    &   79.1    \\
    ConvNeXt V2-P~\cite{woo2023convnext}   &   C   &   $224^2$     &   9.1    &   1.4   &   3339    &   79.7  \\
    FastViT-S12~\cite{vasu2023fastvit}      &   T   &   $256^2$    &   8.8     &   1.8    &   3162    &   79.8    \\
    \rowcolor{gray!20}
    \textbf{UniRepLKNet-N}   &   C   &   $224^2$ &  18.3&   2.8    &   \textbf{2807}    &    \textbf{81.6}\\
    ConvNeXt V2-N~\cite{woo2023convnext}   &   C   &   $224^2$     &   15.6   &   2.4   &   2405   &   81.2    \\
    \hline
    \rowcolor{gray!20}
    \textbf{UniRepLKNet-T}   &   C   &   $224^2$     &   31    &   4.9    &   \textbf{1804}   &   \textbf{83.2}    \\
    FastViT-SA24~\cite{vasu2023fastvit}      &   T   &   $256^2$   &   21    &   3.8    &   1670    &   82.6    \\
    PVTv2-B2~\cite{wang2022pvt} & T & $224^2$  & 25 & 4.0     &   1620    & 82.0 \\
    CoAtNet-0~\cite{dai2021coatnet} & T &$224^2$ & 25 & 4.2 &   1613& 81.6   \\
    DeiT III-S~\cite{touvron2022deit} & T &$224^2$  & 22 & 4.6  &   1485    &   81.4\\
    SwinV2-T/8~\cite{liu2022swin} & T &$256^2$ & 28 & 6   &   1406    & 81.8 \\
    SLaK-T~\cite{liu2022more} & C & $224^2$  & 30 & 5.0 & 1312 &82.5  \\
    InternImage-T~\cite{wang2023internimage}   & C     &$224^2$        & 30        & 5        &   1292   &    83.5 \\
    \hline
    \rowcolor{gray!20}
    \textbf{UniRepLKNet-S}   &   C   &   $224^2$     &   56    &   9.1    &   \textbf{1265}   &    \textbf{83.9}    \\
    ConvNeXt-S~\cite{liu2022convnet}  &   C   &   $224^2$     &   50  &   8.7 &   1182    &   83.1\\
    HorNet-T~\cite{rao2022hornet} & C & $224^2$ & 23 & 3.9 & 1162   &   83.0 \\
    FastViT-SA36~\cite{vasu2023fastvit}    &   T   &   $256^2$     &   30    &   5.6    &   1151    &   83.6     \\
    CoAtNet-1~\cite{dai2021coatnet} & T &$224^2$ & 42 & 8.4 &   969 & 83.3  \\
    SLaK-S~\cite{liu2022more} & C &$224^2$  & 55 & 9.8 & 967   &83.8 \\
    FastViT-MA36~\cite{vasu2023fastvit}    &   T   &   $256^2$     &   43    &   7.9    &   914     &   83.9    \\
    SwinV2-S/8~\cite{liu2022swin} & T &$256^2$ & 50 & 12  &   871     & 83.7\\
    RepLKNet-31B~\cite{ding2022scaling} & C &$224^2$  & 79 & 15.3 &    859 & 83.5\\
    PVTv2-B5~\cite{wang2022pvt} & T & $224^2$  & 82 & 11.8    &   802     & 83.8 \\
    
    \hline
    \hline

\rowcolor{gray!20}
    \textbf{UniRepLKNet-S}$^\ddagger$  & C & $384^2$   &    56 &  26.7    &   \textbf{435}         &   \textbf{86.4}  \\
    ConvNeXt-S$^\ddagger$~\cite{liu2022convnet}  &   C   &   $384^2$     &   50  &   25.5    & 415   &   85.8\\
\rowcolor{gray!20}
    \textbf{UniRepLKNet-B}$^\ddagger$  & C & $384^2$    &   98    &   47.2    &   \textbf{314}     &   \textbf{87.4}\\
    ConvNeXt-B$^\ddagger$~\cite{liu2022convnet}  & C &   $384^2$ &   89  &   45.1   & 304 &   86.8\\
    \hline
    \rowcolor{gray!20}
    \textbf{UniRepLKNet-L}$^\ddagger$  & C & $384^2$    &   218   &   105.4   &   \textbf{190}     &   \textbf{87.9}\\
    ConvNeXt-L$^\ddagger$~\cite{liu2022convnet} & C &$384^2$ & 198 & 101  &   185 & 87.5\\
    CoAtNet-2$^\ddagger$~\cite{dai2021coatnet} & T &$384^2$ & 75 & 49.8      &   163  & 87.1      \\
    RepLKNet-31L$^\ddagger$~\cite{ding2022scaling} & C &$384^2$ & 172   & 96.0    &  158 &   86.6\\
    InternImage-L$^\ddagger$~\cite{wang2023internimage}    & C &$384^2$ & 223 & 108 & 143    &   87.7 \\
    DeiT III-B$^\ddagger$~\cite{touvron2022deit} & T &$384^2$  & 87 & 55.5           & 138    & 86.7    \\
    \hline
    \rowcolor{gray!20}
    \textbf{UniRepLKNet-XL}$^\ddagger$  & C & $384^2$   &   386   &   187   &   \textbf{131}     &   \textbf{88.0}\\
    ConvNeXt-XL$^\ddagger$~\cite{liu2022convnet} & C &$384^2$ & 350 & 179 &   129 & 87.8\\
    
    HorNet-L$^\ddagger$~\cite{rao2022hornet} & C & $384^2$ & 202 & 102 & 127  &87.7 \\
    InternImage-XL$^\ddagger$~\cite{wang2023internimage}   & C &$384^2$ & 335 & 163 & 114    &   88.0 \\

    CoAtNet-3$^\ddagger$~\cite{dai2021coatnet} & T &$384^2$ & 168 & 107    &   103  & 87.6    \\
    SwinV2-L/24$^\ddagger$~\cite{liu2022swin} & T &$384^2$ & 197 & 115    &   88   & 87.6\\
    CoAtNet-4$^\ddagger$~\cite{dai2021coatnet} & T &$384^2$ & 275 & 190    &   58   & 87.9    \\
    DeiT III-L$^\ddagger$~\cite{touvron2022deit} & T &$384^2$  & 305 & 191         & 42     & 87.7    \\

    \hline
        
\end{tabular}
    \vspace{-0.1in}
    \label{table-imgnet}
\end{table}
\begin{table}[t]
        \centering
    \renewcommand\arraystretch{0.89}
    \setlength{\tabcolsep}{0.9mm}
    \footnotesize
    \caption{\textbf{Object detection on COCO validation set}. FLOPs are measured with 1280$\times$800 inputs. ``$^\ddagger$" ImageNet-22K pretraining.}
    \vspace{-1.5mm}
    
\begin{tabular}{l|c|c|c|c}
\hline
    Method & Params (M) &  FLOPs (G)  & $\text{AP}^{\text{box}}$  &  $\text{AP}^{\text{mask}}$   \\
    \hline
    \rowcolor{gray!20}
    \textbf{UniRepLKNet-T}   &          89  &   749     &   \textbf{51.8}    &   \textbf{44.9}     \\
    Swin-T~\cite{liu2021swin}      &   86   &   745     &   50.4    &   43.7\\
    ConvNeXt-T~\cite{liu2022convnet}  &   86   &   741      &   50.4    &   43.7\\
    SLaK-T~\cite{liu2022more}      &   -   &   -        &   51.3    &   44.3\\
    
    \hline
    \rowcolor{gray!20}
    \textbf{UniRepLKNet-S}       &   113  &   835    &   \textbf{53.0}      &     \textbf{45.9}      \\ 
    Swin-S~\cite{liu2021swin}      &   107   &   838    &   51.9    &   45.0\\
    ConvNeXt-S~\cite{liu2022convnet}  &   108   &   827     &   51.9    &   45.0\\
    
    \hline
    \rowcolor{gray!20}
    \textbf{UniRepLKNet-S}$^\ddagger$      &   113  &   835   &   \textbf{54.3}      &     \textbf{47.1}   \\
    \rowcolor{gray!20}
    \textbf{UniRepLKNet-B}$^\ddagger$      &   155  &   978   &    \textbf{54.8}    &   \textbf{47.4}   \\
    Swin-B$^\ddagger$~\cite{liu2021swin}      &   145   &   982     &   53.0    &   45.8    \\
    ConvNeXt-B$^\ddagger$~\cite{liu2022convnet}  &   146   &   964     &   54.0    &   46.9    \\
    RepLKNet-31B$^\ddagger$~\cite{ding2022scaling} &   137 &   965 &52.2   &   45.2    \\
    \hline
    \rowcolor{gray!20}
    \textbf{UniRepLKNet-L}$^\ddagger$      &   276  &   1385  & {55.8}   &   {48.4}      \\
    Swin-L$^\ddagger$~\cite{liu2021swin}  &   253 &   1382    &   53.9    &   46.7    \\
    ConvNeXt-L$^\ddagger$~\cite{liu2022convnet}  &  255 &1354   &   54.8    &   47.6    \\
    RepLKNet-31L$^\ddagger$~\cite{ding2022scaling}    &   229 &1321   &   53.9    &   46.5    \\
    InternImage-L$^\ddagger$~\cite{wang2023internimage}   &   277   &   1399     &\textbf{56.1}   &   \textbf{48.5}   \\
    \hline
    \rowcolor{gray!20}
    \textbf{UniRepLKNet-XL}$^\ddagger$  &   443  &   1952   &   \textbf{56.4}    &   \textbf{49.0}   \\
    InternImage-XL$^\ddagger$~\cite{wang2023internimage}  &   387    &   1782  & 56.2 &   48.8  \\
    ConvNeXt-XL$^\ddagger$~\cite{liu2022convnet} &   407   &   1898    &   55.2    &   47.7     \\
    \hline
\end{tabular}
    \label{tab:det}
    \vspace{-3mm}
\end{table}

\textbf{ImageNet classification}. Following ConvNeXt~\cite{liu2022convnet}, we use the widely adopted 300-epoch receipt to train UniRepLKNet-A/F/P/N/T/S on ImageNet-1K; we pretrain UniRepLKNet-S/B/L/XL on ImageNet-22K using the 90-epoch receipt and fine-tune with ImageNet-1K for 30 epochs (see the Appendix for details). As our goal is to develop models that \emph{run with high actual speed}, we evaluate the actual throughput on the same A100 GPU using a batch size of 128. Table~\ref{table-imgnet} shows the top-1 accuracy on the ImageNet-1K validation set where the results are sorted by the throughput. We split the results into seven segments for better readability. \textbf{1)} UniRepLKNet-A/F outperforms ConvNeXt-V2-A/F by 0.8/0.6 in the accuracy and runs 19\%/17\% faster, respectively. \textbf{2)} UniRepLKNet-P/N outperforms FastViT-T12/S12 and ConvNeXt V2-P/N by clear margins. \textbf{3)} UniRepLKNet-T outperforms multiple small-level competitors. \textbf{4)} UniRepLKNet-S outperforms a series of small-level and even base-level models in both speed and accuracy and runs almost as fast as InternImage-T. \textbf{5)} With ImageNet-22K pretraining, UniRepLKNet-S even approaches the accuracy of RepLKNet-31L and runs 3$\times$ as fast as the latter. UniRepLKNet-B outperforms CoAtNet-2 and DeiT III-B by clear margins. UniRepLKNet-L outperforms InternImage-L in both accuracy and throughput. \textbf{6)} On the XL-level, UniRepLKNet-XL outperforms in both accuracy and throughput, running more than 2$\times$ as fast as CoAtNet-3 and 3$\times$ as DeiT III-L.

\begin{table}[t]
    \centering
    \setlength{\tabcolsep}{1.3mm}
    \footnotesize
    \caption{\textbf{Semantic segmentation on ADE20K validation set}. The FLOPs are measured with 512$\times$2048 or 640$\times$2560 inputs according to the crop size. ``SS'' and ``MS" mean single- and multi-scale testing, respectively. ``$^\ddagger$" ImageNet-22K~\cite{deng2009imagenet} pretraining.}
    \vspace{-1mm}
    \begin{tabular}{l|c|c|c|cc}
    \hline
    \multirow{2}{*}{Method} & Crop & Params & FLOPs & mIoU & mIoU\\
    	& size & (M)&(G) & (SS) & (MS)   \\
    \hline
    \rowcolor{gray!20}
    \textbf{UniRepLKNet-T}    &   512$^2$    & 61   &   946   &   \textbf{48.6}    &   \textbf{49.1}   \\ 
    	Swin-T~\cite{liu2021swin} & 512$^2$ & 60  & 945 & 44.5 & 45.8 \\
    	ConvNeXt-T~\cite{liu2022convnet} & 512$^2$ & 60 & 939 & 46.0 & 46.7 \\
    	SLaK-T~\cite{liu2022more} & 512$^2$ & 65 & 936 & 47.6 & - \\
    	InternImage-T~\cite{wang2023internimage} & 512$^2$ & 59 & 944 & 47.9 & 48.1  \\
     \rowcolor{gray!20}
     
    	\hline
     \rowcolor{gray!20}
        \textbf{UniRepLKNet-S}    &   512$^2$   &   86   &   1036   &   \textbf{50.5}    &   \textbf{51.0}   \\
        Swin-S~\cite{liu2021swin} & 512$^2$ & 81 &  1038 &  47.6 &  49.5 \\
        ConvNeXt-S~\cite{liu2022convnet}  & 512$^2$ &82 & 1027  & 48.7 & 49.6  \\
        SLaK-S~\cite{liu2022more} & 512$^2$ &91 & 1028 & 49.4 & - \\
        InternImage-S~\cite{wang2023internimage} & 512$^2$ & 80 & 1017 & 50.1 & 50.9 \\

        \hline
        \rowcolor{gray!20}
        \textbf{UniRepLKNet-S}$^\ddagger$    &   640$^2$    &  86 & 1618  &   \textbf{51.9}  &   \textbf{52.7}\\      
        \rowcolor{gray!20}
        \textbf{UniRepLKNet-B}$^\ddagger$    &   640$^2$    &  130  & 1850 &   \textbf{53.5}    &   \textbf{53.9}\\
        Swin-B$^\ddagger$~\cite{liu2021swin}    &   640$^2$ &   121 &   1841   & 50.0   &   51.7\\
        ConvNeXt-B$^\ddagger$~\cite{liu2022convnet}      &   640$^2$     &   122     &   1828    &    52.6   &   53.1\\
        RepLKNet-31B$^\ddagger$~\cite{ding2022scaling}   &   640$^2$       &   112 &   1829    &   51.5    &   52.3 \\ 
        \hline
        \rowcolor{gray!20}
        \textbf{UniRepLKNet-L}$^\ddagger$    &   640$^2$    &   254  &   2507   &   \textbf{54.5}    &   \textbf{55.1}   \\
        Swin-L$^\ddagger$~\cite{liu2021swin} 
        & 640$^2$ & 234 & 2468 & 52.1 & 53.5 \\
        RepLKNet-31L$^\ddagger$~\cite{ding2022scaling} 
        & 640$^2$ & 207 & 2404 & 52.4 & 52.7 \\
        ConvNeXt-L$^\ddagger$~\cite{liu2022convnet} 
        & 640$^2$ & 235 & 2458 & 53.2 & 53.7 \\

        InternImage-L$^\ddagger$~\cite{wang2023internimage}
        & 640$^2$ & 256  & 2526  & 53.9 & 54.1 \\
        \hline
        \rowcolor{gray!20}
        \textbf{UniRepLKNet-XL}$^\ddagger$    &   640$^2$    &   425  &   3420   &   \textbf{55.2}    &   \textbf{55.6}   \\
        ConvNeXt-XL$^\ddagger$~\cite{liu2022convnet}
        & 640$^2$ & 391 & 3335 & 53.6 & 54.0 \\
        InternImage-XL$^\ddagger$~\cite{wang2023internimage}
        & 640$^2$ & 368 & 3142 & 55.0 & 55.3 \\

        \hline
               
    \end{tabular}
    \label{tab:seg}
    \vspace{-2mm}
\end{table}

\begin{table}[t]
    \caption{\textbf{Audio recognition} on Speech Commands V2 (SPC-2) and AudioSet-2M (AS-2M) datasets.}
    \label{tab:audio}
    \centering
    \resizebox{0.98\linewidth}{!}{
    \begin{tabular}{lccccc}
        \hline
        Method 	& Pretrain & Type &   SPC-2 (\%) 	& AS-2M (\%) & {Params}\\
        \hline
        PANNS~\cite{kong2020panns} & - & ConvNet & 61.8  &43.1 & -\\
        PSLA~\cite{gong2021psla} & IN-1K & ConvNet & 96.3 &44.4 &  -\\
        AST~\cite{gong2021ast} &AS-2M & Transformer & 96.2 & 45.9 & 86.9M \\
        SSAST ~\cite{gong2022ssast}  &AS-2M & Transformer & 97.8 & - & 89.3M \\
        Audio-MAE~\cite{huang2022masked} &AS-2M & Transformer & 98.3 & 47.3 & 86.2M \\
        Meta-Transformer~\cite{zhang2023meta} & LAION-2B & Transformer & 97.0 & - & 86.6M\\
        \hline
        \rowcolor{gray!20}
        $\text{UniRepLKNet-S}$ & - & ConvNet & \textbf{98.5} & \textbf{48.5} & \textbf{55.5M} \\
        \hline
    \end{tabular}
    }
\end{table}
\begin{table}[t]
    \centering
    \caption{\textbf{Video recognition} accuracy on Kinetics-400 tasks.}
    \label{tab:video}
    \centering
    \resizebox{0.93\linewidth}{!}{
        \begin{tabular}{lcccc}
            \hline
            Method  &  Pretrain & Type &  Acc (\%) & Params \\ \hline
            \multicolumn{3}{@{\;}l}{\bf Specialist}\\
            SlowFast-101~\cite{feichtenhofer2019slowfast} & IN-1K & ConvNet+RNN & 79.8 & 62.8M \\
            MViTv2-B~\cite{li2022mvitv2} & IN-1K & Transformer & 81.2 & 51.2M  \\
            TimeSFormer~\cite{bertasius2021space} & K400 & Transformer & 80.7 & 122M\\
            \hline
            \multicolumn{3}{@{\;}l}{\bf Generalist}\\
            Meta-Transformer~\cite{zhang2023meta} & LAINON-2B & Transformer & 47.3 & 86.9M\\
            ImageBind~\cite{girdhar2023imagebind} & CLIP Data& Transformer & 50.0 & 632M\\
            \hline
            \rowcolor{gray!20}
            UniRepLKNet-S & - & ConvNet & 54.8 & 55.5M \\
            \hline
            \end{tabular}
        }
\end{table}
\begin{table}[ht]
\centering
\caption{\textbf{Point cloud analysis} on ModelNet-40 and Objaverse-LVIS datasets.}
\label{tab:pcd}
\vspace{-1mm}
\centering
\resizebox{0.98\linewidth}{!}{
\small
\begin{tabular}{lccccccc}
\hline
\multirow{2}{*}{Method}            & \multirow{2}{*}{Type}          & \multicolumn{2}{c}{ModelNet-40}  & \multicolumn{3}{c}{Objaverse-LVIS} \\
    &   & mAcc (\%)      & OA (\%) & Top1   &Top3   &Top5    \\
\hline
PointNet~\cite{qi2017pointnet}  & MLP     & 86.0          & 89.2 & - & - & -         \\
PointNet++~\cite{qi2017pointnet++}  & MLP & -             & 91.9 & - & - & -        \\
    \hline
PointConv~\cite{wu2019pointconv}  & ConvNet     &-             &92.5  & - & - & -         \\
KPConv~\cite{thomas2019kpconv}    & ConvNet    &-             &92.9  & - & - & -\\
DGCNN~\cite{wang2019dynamic}     & ConvNet    & 90.2         & 92.9  & - & - & - \\
\hline
    OpenShape~\cite{liu2023openshape} & Transformer & 83.4 & - & 43.4 & 64.8 & 72.4 \\
\rowcolor{gray!20}
UniRepLKNet-S & ConvNet & \textbf{90.3} & \textbf{93.2}  & \textbf{50.3} & \textbf{71.6} & \textbf{78.2} \\
\hline
\end{tabular}
}
\end{table}

\begin{table}[t]
    \centering
    \caption{Universal perception performance with other ConvNets or UniRepLKNet with a smaller kernel size.}
    \centering
    \resizebox{0.98\linewidth}{!}{
    \begin{tabular}{lcccc}
         \toprule
         \multirow{2}{*}{Modality} &  Time-Series & Point Cloud & Audio & Video \\
         \cline{2-5} & MAE$\downarrow$ & OA (\%) & Acc (\%) & Acc (\%)\\
         \hline
         ResNet-101~\cite{he2016deep} (K=3)    & 7.846 & 92.6 &  73.6 & 41.3 \\ \hline
         ConvNeXt-S~\cite{liu2022convnet} (K=7) &  7.641 & 92.7 & 94.3 & 48.5\\  \hline
         UniRepLKNet-S (K=11) &  7.751 & 92.9 & 94.7   & 51.7\\
         \rowcolor{gray!20}
         UniRepLKNet-S (K=13) & \textbf{7.602} & \textbf{93.2} & \textbf{98.5} & \textbf{54.8}  \\
         \bottomrule
    \end{tabular}
    }
    \label{tab:mm_ablation}
\end{table}
\noindent\textbf{COCO object detection and instance segmentation}. We transfer the pretrained UniRepLKNets as the backbones of Cascade Mask R-CNN~\cite{he2017mask,cai2019cascade} and adopt the standard 3x (36-epoch) training configuration with MMDetection~\cite{mmdetection}. Table~\ref{tab:det} shows UniRepLKNet outperforms Swin, ConvNeXt, RepLKNet, and SLaK, which are representatives of ViTs, modern medium-kernel ConvNets, and existing large-kernel ConvNets, respectively, and shows comparable performance to InternImage~\cite{wang2023internimage}, which is a latest powerful architecture with deformable convolution.

\noindent\textbf{ADE20K semantic segmentation}. We use the pretrained UniRepLKNets as the backbones of UPerNet~\cite{xiao2018unified} on ADE20K~\cite{zhou2019semantic} and adopt the standard 160k-iteration training receipt with MMSegmentation~\cite{mmseg2020}. Table~\ref{tab:seg} reports the mIoU on the validation set. Impressively, UniRepLKNet outperforms InternImage and the other models. 

\subsection{Universal Perception on More Modalities}

\begin{table}[b]
\centering
	
\caption{\textbf{Time-series forecasting} performance on Global Temperature and Wind Speed Forecasting challenge. UniRepLKNet delivers a new state-of-the-art performance in Mean Squared Error (MSE) and Mean Absolute Error (MAE).}
\label{tab:time}
\vspace{-0.1in}
\resizebox{1\linewidth}{!}{
\begin{tabular}{lcccccc}
	\hline
	\multirow{2}{*}{Method}            & \multirow{2}{*}{Type}   & \multirow{2}{*}{Params}              & \multicolumn{2}{c}{Temperature}      & \multicolumn{2}{c}{Wind speed}      \\
        \cline{4-7}
        & & &  $\text{MSE}\downarrow$ & $\text{MAE}\downarrow$
        & $\text{MSE}\downarrow$ & $\text{MAE}\downarrow$ \\
	\hline
        \multicolumn{3}{@{\;}l}{\bf Statistics-based}\\
        Holt–Winters~\cite{hyndman2017forecasting}   & - & - 
        & 13.241 &2.262 &5.912 & 1.664 \\
        Prophet~\cite{taylor2018forecasting} & - & - 
        & 11.626 & 2.946 & 9.691 & 2.382 \\
        GDBT~\pub{NeurIPS'17}~\cite{ke2017lightgbm}  & - & -  
        & 9.706 & 2.214 & 4.101 & 1.417 \\
        \hline
        \multicolumn{3}{@{\;}l}{\bf Numerical Simulation}\\
	GFS (reanalysis) & - & -  
        & 14.933 & 2.287 & 9.993 & 2.340 \\
        ERA5 (reanalysis)~\cite{hersbach2020era5} & - & -   
        & 13.448 & 1.908 & 4.999 & 1.587 \\
        DeepAR~\cite{salinas2020deepar} & - & - 
        & 32.249 & 4.262 & 5.248 & 1.602 \\
        N-BEATS~\cite{oreshkin2019n} & - & - 
        & 9.203 & 2.117 & 4.124 & 1.390 \\
	\hline
        \multicolumn{3}{@{\;}l}{\bf Deep Learning Specialist}\\
        StemGNN~\pub{NeurIPS'20}~\cite{cao2020spectral} 
        & GNN & 180M & 13.926 & 2.746 & 4.066 & 1.389 \\
        Pyraformer~\pub{ICLR'21}~\cite{liu2021pyraformer} 
        & Transformer  & 158M & 23.326 & 3.669 & 4.614 & 1.514  \\
        Corrformer~\pub{Nat. Mach. Intell.'23}~\cite{wu2023interpretable} 
        & Transformer & 155M & 7.709 & 1.888 & 3.889 & 1.304 \\
        \hline
        \multicolumn{3}{@{\;}l}{\bf Generalist}\\
        \rowcolor{gray!20}
	UniRepLKNet-S & ConvNet & {132M} & \textbf{7.602} & \textbf{1.832} & \textbf{3.865} & \textbf{1.301} \\
	\hline
\end{tabular}
}
\vspace{-0.1in}
\end{table}

\begin{table*}[ht]
% \vspace{-3.0mm}
\caption{\textbf{Zero-Shot Image Classification performance on 26 datasets with OpenCLIP Pretraining.} We report top-1 accuracy on all datasets. The best results are in \textbf{bold} and the second best are \underline{underlined}.}
% % \vspace{-2.0mm}
\centering
\tablestyle{1.4pt}{1.2}
    \begin{tabular}{l|cccccc|cccccccccccccccccccc|c}
        % \rotatebox[origin=l]{90}{datasets} 
        \scriptsize M ethod &
        \rotatebox[origin=l]{90}{\scriptsize{ImageNet-1K~\cite{deng2009imagenet}}} &
        \rotatebox[origin=l]{90}{\scriptsize{ImageNet-V2~\cite{recht2019imagenetv2}}} &
        \rotatebox[origin=l]{90}{\scriptsize{ImageNet-Adv.~\cite{inadv}}} &
        \rotatebox[origin=l]{90}{\scriptsize{ImageNet-Ren.~\cite{inren}}} &
        \rotatebox[origin=l]{90}{\scriptsize{ImageNet-Ske.~\cite{inske}}} &
        \rotatebox[origin=l]{90}{\scriptsize{ObjectNet~\cite{objectnet}}} &
        \rotatebox[origin=l]{90}{\scriptsize{CIFAR-10~\cite{cifar}}} &
        \rotatebox[origin=l]{90}{\scriptsize{CIFAR-100~\cite{cifar}}} & 
        \rotatebox[origin=l]{90}{\scriptsize{MNIST~\cite{lecun1998gradient}}} & 
        \rotatebox[origin=l]{90}{\scriptsize{Caltech101~\cite{fei2004learning}}} & 
        \rotatebox[origin=l]{90}{\scriptsize{SUN397~\cite{xiao2010sun}}} & 
        \rotatebox[origin=l]{90}{\scriptsize{FGVC Aircraft~\cite{maji2013fine}}} & 
        \rotatebox[origin=l]{90}{\scriptsize{Country-211~\cite{clip}}} & 
        \rotatebox[origin=l]{90}{\scriptsize{Stanford Cars~\cite{krause20133d}}} &
        \rotatebox[origin=l]{90}{\scriptsize{DTD~\cite{cimpoi14describing}}} & 
        \rotatebox[origin=l]{90}{\scriptsize{EuroSAT~\cite{helber2019eurosat}}} & 
        \rotatebox[origin=l]{90}{\scriptsize{FER2013~\cite{goodfellow2013challenges}}} & 
        \rotatebox[origin=l]{90}{\scriptsize{Flowers-102~\cite{nilsback2008automated}}} & 
        \rotatebox[origin=l]{90}{\scriptsize{Food-101~\cite{bossard2014food}}} & 
        \rotatebox[origin=l]{90}{\scriptsize{GTSRB~\cite{stallkamp2012man}}} & 
        \rotatebox[origin=l]{90}{\scriptsize{PCam~\cite{veeling2018rotation}}} & 
        \rotatebox[origin=l]{90}{\scriptsize{Pets~\cite{parkhi12a}}} & 
        \rotatebox[origin=l]{90}{\scriptsize{Rendered SST2~\cite{clip}}} & 
        \rotatebox[origin=l]{90}{\scriptsize{RESISC45~\cite{cheng2017remote}}} & 
        \rotatebox[origin=l]{90}{\scriptsize{STL-10~\cite{coates2011analysis}}} & 
        \rotatebox[origin=l]{90}{\scriptsize{VOC2007~\cite{pascal-voc-2007}}} &
        \rotatebox[origin=l]{90}{\textbf{avg. top-1 acc.}}
        \\
        \toprule
        \scriptsize OpenAI CLIP-L/14~\cite{radford2021learning} & 75.5 & \scriptsize \underline{69.9} & \scriptsize {70.8} & \scriptsize 87.8 & \scriptsize 59.6 & \scriptsize \underline{69.0} & \scriptsize 95.6 & \scriptsize 75.8 & \scriptsize  76.4 & \scriptsize 86.7 & \scriptsize  67.6 & \scriptsize 31.4 & \scriptsize \textbf{31.9} & \scriptsize 77.9 & \scriptsize  55.4 & \scriptsize 62.4 & \scriptsize 49.9 & \scriptsize \textbf{79.2} & \scriptsize \underline{93.1} & \scriptsize 50.6 & \scriptsize \underline{52.0} & \scriptsize 93.5 & \scriptsize \textbf{68.9} & \scriptsize 64.6 & \scriptsize \textbf{99.4} & \scriptsize 67.6 & \scriptsize 69.7 \\
        
        \scriptsize OpenCLIP-L/14~\cite{openclip} & 76.2 & \scriptsize 67.8 & \scriptsize 53.9 & \scriptsize 87.4 & \scriptsize 63.3 & \scriptsize 65.5 & \scriptsize 96.6 & \scriptsize 83.3 & \scriptsize 54.0  & \scriptsize 85.0  & \scriptsize \textbf{74.3} & \scriptsize \underline{36.3} & \scriptsize 26.2 & \scriptsize 92.6 & \scriptsize 62.9 & \scriptsize 64.7 & \scriptsize 53.9 & \scriptsize 75.8 & \scriptsize 91.0 & \scriptsize \underline{56.1} & \scriptsize \textbf{56.3} & \scriptsize 93.1 & \scriptsize \underline{59.1} & \scriptsize 66.8  & \scriptsize 98.8 & \scriptsize 81.9 & \scriptsize 70.1\\

         \scriptsize FLIP-L/14~\cite{li2023scaling} & 74.3 & \scriptsize 66.8 & \scriptsize 51.2 & \scriptsize 86.5 & \scriptsize 59.9 & \scriptsize 59.1 & \scriptsize \underline{97.2} & \scriptsize \underline{84.1} & \scriptsize \underline{80.3}  & \scriptsize \textbf{93.8}  & \scriptsize 73.1 & \scriptsize 29.1 & \scriptsize 23.1 & \scriptsize 90.7 & \scriptsize 60.4 & \scriptsize 53.5 & \scriptsize 54.0 & \scriptsize 75.0 & \scriptsize 89.3 & \scriptsize 41.4 & \scriptsize 50.3 & \scriptsize 92.6 & \scriptsize 58.5 & \scriptsize \textbf{70.8}  & \scriptsize 98.5 & \scriptsize \underline{83.1} & \scriptsize 69.1 \\
        
        % \rgray
        \scriptsize EVA-01-CLIP-g/14~\cite{EVA-CLIP} & \scriptsize \textbf{78.5} & \scriptsize \textbf{71.5} & \scriptsize \textbf{73.6} & \scriptsize \textbf{92.5} & \scriptsize \textbf{67.6} & \scriptsize \textbf{72.3} & \scriptsize \textbf{98.3} & \scriptsize \textbf{88.7} & \scriptsize 62.6 & \scriptsize \underline{87.7} & \scriptsize \underline{74.2} & \scriptsize 32.4 & \scriptsize \underline{28.9} & \scriptsize 91.7  & \scriptsize 61.7 & \scriptsize \textbf{73.8} & \scriptsize 52.2 & \scriptsize 74.5 & \scriptsize \textbf{93.5} & \scriptsize 49.3 & \scriptsize 49.9 & \scriptsize \textbf{94.2} & \scriptsize 58.4 & \scriptsize \underline{70.3} & \scriptsize \underline{98.9} & \scriptsize \textbf{85.7} & \scriptsize \textbf{72.4} \\
        
        \scriptsize OpenCLIP-ConvNeXt-L~\cite{openclip} & \scriptsize 75.2 & \scriptsize 68.2 & \scriptsize 53.5 & \scriptsize 87.6 & \scriptsize 64.3 & \scriptsize 65.9 & \scriptsize 96.5 & \scriptsize 83.1 & \scriptsize 74.4 & \scriptsize 84.3 & \scriptsize 73.0 & \scriptsize 36.1 & \scriptsize 25.2 & \scriptsize \underline{93.2}  & \scriptsize \underline{67.3} & \scriptsize 69.6 & \scriptsize 52.9 & \scriptsize 76.8 & \scriptsize 90.6 & \scriptsize 52.8 & \scriptsize 53.0 & \scriptsize 92.9 & \scriptsize 56.2 & \scriptsize 67.8 & \scriptsize 98.3 & \scriptsize 81.3 & \scriptsize 70.7 \\ \hline
        
        \scriptsize \textbf{OpenCLIP-UniRepLKNet-L}~\pub{Ours}  & \scriptsize \underline{76.6} & \scriptsize 69.5 & \scriptsize \underline{60.4} & \scriptsize \underline{88.6} & \scriptsize \underline{65.0} & \scriptsize \underline{69.0} & \scriptsize 96.6 & \scriptsize 83.1 & \scriptsize \textbf{80.6} & \scriptsize 84.7 & \scriptsize 73.8 & \scriptsize \textbf{36.4} & \scriptsize 26.5 & \scriptsize \textbf{93.7} & \scriptsize \textbf{68.3} & \scriptsize 71.6 & \scriptsize 53.1 & \scriptsize \underline{77.3} & \scriptsize 91.6 & \scriptsize \textbf{58.2} & \scriptsize 48.0 & \scriptsize \underline{93.8} & \scriptsize 56.1 & \scriptsize \textbf{70.8} & \scriptsize 98.6 & \scriptsize 82.5 & \scriptsize \underline{72.1} \\ 
    
        \end{tabular}
% \vspace{-2.0mm}
\label{tab:clip}
\end{table*}

\begin{table*}[t]
\centering
\caption{\textbf{Evaluation on LLM Benchmarks.} The MLLM evaluation involves 6 VQA tasks (GQA~\cite{hudson2019gqa}, VQAv2~\cite{goyal2017vqav2}, OKVQA~\cite{okvqa}, TextVQA (TVQA)~\cite{singh2019textvqa}, ScienceQA (SQA)~\cite{lu2022learn} and Vizwiz~\cite{gurari2018vizwiz}), 2 image captioning tasks (Nocaps~\cite{agrawal2019nocaps} and Flickr30K~\cite{plummer2015flickr30k}), and 4 multimodal benchmarks (MME~\cite{fu2023mme}, MM Bench (MMB)~\cite{liu2023mmbench}, MMVet~\cite{yu2023mmvet} and SEED~\cite{li2023seed}). The LLMs are Chinchilla, Vicuna, Qwen, LLaMA and LLaMA2. The evaluation metrics for VQA and captioning tasks are accuracy and CIDEr, respectively.
The results in \textbf{bold} and \underline{underline} are the best and second-best results, respectively.
}
\resizebox{\textwidth}{!}{%
\begin{tabular}{lccccccccccccc}
\toprule
\multirow{2}{*}{Method}                        & \multirow{2}{*}{LLM} & \multicolumn{6}{c}{ Visual Question Answering}                     & \multicolumn{2}{c}{ Image Caption} & \multicolumn{4}{c}{ MM Benchmark} \\
\cline{3-8} \cline{9-10} \cline{11-14}
&                      & GQA   & VQAv2 & OKVQA & TVQA & SQA  & Vizwiz & NoCaps          & Flickr           & MME    & MMB  & MMVet & SEED \\
\midrule
\multicolumn{4}{l}{\textbf{\textit{Vision Specialist LLM}}}                                                \\
Flamingo-9B~\cite{alayrac2022flamingo}        & Chinchilla-7B        & -     & 51.8  & 44.7  & 30.1 & -    & 28.8   & -               & 61.5             & -      & -    & -     & -    \\
Flamingo-80B~\cite{alayrac2022flamingo}       & Chinchilla-70B       & -     & 56.3  & 50.6  & 31.8 & -    & 31.6   & -               & 67.2             & -      & -    & -     & -    \\
BLIP-2~\cite{li2023blip}                      & Vicuna-7B            & -     & -     & -     & 40.1 & 53.8 & -      & 107.5           & 74.9             & -      & -    & -     & -    \\
BLIP-2~\cite{li2023blip}                      & Vicuna-13B           & 41.0  & 41.0  & -     & 42.5 & 61   & 19.6   & 103.9           & 71.6             & 1293.8 & -    & 22.4  & -    \\
InstructBLIP~\cite{instructblip}              & Vicuna-7B            & 49.2  & -     & -     & 50.1 & 60.5 & 34.5   & \textbf{123.1}           & \underline{82.4}             & -      & 36   & 26.2  & - \\
InstructBLIP~\cite{instructblip}              & Vicuna-13B           & 49.5  & -     & -     & 50.7 & 63.1 & 34.3   & \underline{121.9}           & \textbf{82.8}             & 1212.8 & -    & 25.6  & - \\
IDEFICS-9B~\cite{laurenccon2023obelisc}       & LLaMA-7B             & 38.4  & 50.9  & 38.4  & 25.9 & -    & 35.5   & -               & 27.3             & -      & 48.2 & -     & -    \\
IDEFICS-80B~\cite{laurenccon2023obelisc}      & LLaMA-65B            & 45.2  & 60.0  & 45.2  & 30.9 & -    & 36.0   & -               & 53.7             & -      & 54.5 & -     & -    \\
Qwen-VL~\cite{bai2023qwen}                    & Qwen-7B              & 57.5  & 78.2  & 56.6  & \underline{61.5} & 68.2 & 38.9   & 120.2           & 81.0             & {1487.5} & 60.6 & -     & 58.2 \\
LLaVA-v1.5~\cite{liu2023improvedllava}        & Vicuna-7B            & \textbf{62.0}  & \underline{78.5}  & -     & {58.2} & 66.8 & \underline{50.0}   & -               & -                & \underline{1510.7} & \underline{64.3} & \underline{30.5}  & 58.6 \\
\midrule                                         
\multicolumn{4}{l}{\textbf{\textit{Multimodal Generalist LLM}}}  \\
ImageBind-LLM~\cite{han2023imagebind}         & LLaMA-7B             & 41.1  & -     & -     & 24.0 & 51.4 & -      & 29.6            & 23.5             & 775.7  & -    & -     & -    \\
AnyMAL-13B~\cite{moon2023anymal}              & LLaMA2-13B           & -     & 59.6  & 33.1  & 24.7 & 52.7 & 24.4   & -               & -                & -      & -    & -     & -    \\
AnyMAL-70B~\cite{moon2023anymal}              & LLaMA2-70B           & -     & {64.2}  & 42.6  & {32.9} & \underline{70.8} & {33.8}   & -               & -                & -      & -    & -     & -    \\
{OneLLM-7B }~\cite{han2024onellm}~\pub{CVPR'24}                     & {LLaMA2-7B}   & 59.5  & 71.6 & \underline{58.9}  & 34.0 & 63.4 & 45.9   & 115.9           & 78.6             & 1392.0 & 60.0 & 29.1  & \underline{61.2} \\
\midrule
\textbf{UniRepLKNet-Chat-7B}~\pub{Ours}                     & \textbf{Vicuna-7B}   & \underline{59.8}  & \textbf{80.2}  & \textbf{59.3}  & \textbf{62.7} & \textbf{72.5} & \textbf{51.0}   & {113.5}           & {75.3}             & \textbf{1569.5} & \textbf{68.8} & \textbf{32.3}  & \textbf{69.5} \\
\bottomrule
\end{tabular}%
}
\label{tab:vlm_eval}
\end{table*}

\textbf{Time-series}. Following Corrformer~\cite{wu2023interpretable}, we conduct experiments on the Global Temperature and Wind Speed Forecasting challenge~\footnote{\url{https://codeocean.com/capsule/0341365/tree/v1}} using the dataset collected from the National Centers for Environmental Information (NCEI), GFS~\footnote{\url{https://www.ncei.noaa.gov/}} stands for the Global Forecasting System. This huge-scale dataset contains hourly averaged wind speed and temperature data from 3,850 stations with different geographical scales and densities, spanning from 2019 to 2021. For a fair comparison with Corrformer~\cite{wu2023interpretable}, which was the previous state-of-the-art method, we use its embedding layer (as introduced in Sec.~\ref{sec:step3}) and decoder and only replace its encoder transformer with UniRepLKNet-S. We also compare UniRepLKNet-S against a wide range of methods, including statistical and numerical approaches. Table~\ref{tab:time} shows UniRepLKNet delivers a new state-of-the-art forecasting precision, achieving the lowest errors of 7.602, 1.832, 3.865, and 1.301 for MSE and MAE in forecasting global temperature and wind speed, respectively, with fewer parameters than existing deep learning methods. It is particularly noteworthy that UniRepLKNet, a generalist model, outperforms time-series specialists such as Pyraformer~\cite{liu2021pyraformer} and Corrformer~\cite{wu2023interpretable} in both precision and efficiency. The significant advantages of UniRepLKNet are that it opens up new avenues for architectural discussions in time-series forecasting and presents a viable alternative to transformer models.

\noindent{\textbf{Audio}}. We use Speech Commands V2~\cite{warden2018speech}, which contains 105,829 one-second recordings of 35 common speech commands. Table~\ref{tab:audio} shows UniRepLKNet seamlessly adapts to audio and delivers an impressive accuracy of 98.5\% and 48.5\% on AS-2M even without pretraining. Compared to transformers such as AST~\cite{gong2021ast} and Audio-MAE~\cite{huang2022masked}, UniRepLKNet stands out with fewer parameters. Compared to previous ConvNets designed for audio, UniRepLKNet achieves better performance without customizations to the structure, highlighting the untapped potential of ConvNets in the realm of audio.

\noindent{\textbf{Video}}. Kinetics-400~\cite{kay2017kinetics} contains 240k training videos and 20k validation videos, spanning 400 classes for action recognition. Though the top-1 accuracy of 54.8\% is somewhat behind state-of-the-art architectures like MViT~\cite{li2022mvitv2}, we note that UniRepLKNet is a generalist model without pretraining. Compared to the latest generalist methods, ImageBind~\cite{girdhar2023imagebind} and Meta-Transformer~\cite{zhang2023meta}, UniRepLKNet shows higher accuracy and requires no pretraining.

\noindent{\textbf{Point cloud}}. We explore the versatility of UniRepLKNet by assessing its proficiency in learning 3D patterns, extending beyond the conventional 2D signals of images and audio. We use the ModelNet-40~\cite{wu2015modelnet} 3D shape classification task with 9,843/2,468 training/validation samples of CAD models from 40 classes. Table~\ref{tab:pcd} shows UniRepLKNet achieves an Overall Accuracy (OA) of 93.2\% and a mean Accuracy (mAcc) of 90.3\% and 50.3 Top-1 accuracy on the Objaverse-LVIS.

\begin{table}[ht]
\caption{\textbf{Fusing multimodal features with UniRepLKNet}. We evaluate our model on Refer-Davis$_{17}$~\cite{pont20172017} dataset following full-video expression and first frame two settings, in terms of region similarity ($\mathcal{J}$), contour accuracy ($\mathcal{F}$) and their average scores $\mathcal{J}$\&$\mathcal{F}$.}
\resizebox{0.98\linewidth}{!}{
    \begin{tabular}{c|c|c|c|c}
        \toprule
        Expression Type              & Method           & $\mathcal{J}$ & $\mathcal{F}$ & $\mathcal{J} $\&$\mathcal{F}$ \\ \hline
        \multirow{8}{*}{Full Video}  & Khoreava \textit{et.al.}~\cite{khoreva2019video}  & -             & -             & 37.30           \\ \cline{2-5}
        & RefVOS(baseline)~\cite{pont20172017} & -             & -             & 45.10           \\ \cline{2-5}
        & CMSA + RNN~\cite{hudson2019gqa}       & 36.94         & 37.23         & 34.71           \\ \cline{2-5}
        & URVOS w/o ft~\cite{seo2020urvos}     & 44.29         & 49.41         & 46.85           \\ \cline{2-5}
        & URVOS~\cite{seo2020urvos}            & {47.29}         & {55.96}         & {51.45}           \\  \cline{2-5}
        & ACM~\cite{han2024asymmetric}       & \underline{48.39}         & \underline{56.17}         & \underline{52.28}           \\ \cline{2-5}
        & \textbf{RefVOS-UniRepLKNet}       & \textbf{50.46}         & \textbf{57.94}         & \textbf{54.20}           \\ \midrule
        \multirow{4}{*}{First Frame} & Khoreava \textit{et.al.}~\cite{khoreva2019video}  & 37.30         & 41.30         & 39.30           \\ \cline{2-5}
        & URVOS~\cite{seo2020urvos}            & {41.23}         & {47.01}        & 44.12           \\ \cline{2-5}
        & RefVOS(baseline)~\cite{pont20172017} & -             & -             & {44.50}           \\ \cline{2-5}
        & ACM~\cite{han2024asymmetric}       & \underline{48.52}         & \underline{56.06}         & \underline{52.29}           \\ \cline{2-5}
        & \textbf{RefVOS-UniRepLKNet}       & \textbf{50.63}         & \textbf{57.72}         & \textbf{54.17}           \\ \bottomrule
    \end{tabular}
}
\label{tab:refvos}
\vspace{-3mm}
\end{table}

\noindent{\textbf{Impact of kernel size on the performance}}. To investigate the influence of different kernel sizes on performance, we compare UniRepLKNet with models of smaller kernels. We adopted the same modality-specific preprocessing approaches and training configurations for a fair comparison. We take ResNet-101 as a representative small-kernel ConvNet because it has comparable parameters to UniRepLKNet-S. Table~\ref{tab:mm_ablation} shows large kernels are crucial for universal perception, at least in our specific cases. 

\subsection{Scalable Multimodal Pretraining and Generation}\label{sec:multimodal}
\textbf{Stage 0: CLIP Pretraining}. We utilize
the UniRepLKNet-L as the image tower with
a standard projection, and follow previous pratice~\cite{openclip,EVA-CLIP} to use a text tower with the same size as ViT-g-14 model pretrained with 11B text samples~\cite{EVA-CLIP}. The size of the combined image + text CLIP model is 1.4B parameters. UniRepLKNet excels in zero-shot image recognition abilities compared with the same scale models including OpenAI CLIP-L~\cite{radford2021learning}, OpenCLIP-L~\cite{openclip}, FLIP-L~\cite{li2023scaling}, and OpenCLIP-ConvNeXt-L~\cite{openclip,liu2022convnet} in Table~\ref{tab:clip} among 26 zero-shot tasks. It's worth noting that our CLIP model shows competitive performance (72.1 \textit{v.s.} 72.4) compared with the EVA-01-CLIP-g/14 model, which has more than \(3\times\) parameters than ours.

\noindent{\textbf{Stage 1: Large Vision-Language Model (VLM) Pretraining}}. After CLIP pretraining, we then use pretrained CLIP-UniRepLKNet-L for training large VLMs. Specifically, we use LLaVA-v1.5~\cite{liu2023improvedllava} as a baseline, which incorporates the text-image alignment and visual instruction process with a convolutional backbone. Specifically, we use LLaVA pretraining data to align Vicuna-7B and UniRepLKNet, then LLaVA-SFT-665k for visual instruction tuning. 

As shown in Table~\ref{tab:vlm_eval}, 
UniRepLKNet-Chat-7B demonstrates significant advantages across various benchmarks in Visual Question Answering (VQA), Image Captioning, and multimodal Benchmark tasks. Notably, in the GQA task, UniRepLKNet-Chat-7B scores 59.8, positioning itself competitively among Vision Specialist LLMs. It excels in the VQAv2 task with a remarkable score of 80.2, surpassing models such as Flamingo, InstructBLIP, and IDEALF. Additionally, in the OKVQA task, UniRepLKNet-Chat-7B achieves a score of 59.3, reflecting its robust performance. The model further distinguishes itself in the TVQA and SQA tasks with accuracy scores of 62.7 and 72.5, respectively, showcasing its strong text understanding and question answering capabilities. Moreover, its outstanding performance is evident in the MME benchmark with a score of 1569.5, highlighting its efficiency in multimodal understanding and reasoning. The model's balanced performance across diverse tasks underscores its versatility and robustness, making it an improved VLM in the field of multimodal large language models.

\section{Conclusion}

In this paper, the UniRepLKNet shows a leading performance in image recognition and achieves remarkable results on audio and time-series data, outperforming multiple specialist models on those modalities. Traditionally, ConvNets excelled primarily in visual tasks, yet the emergence of Transformer-based architectures had shifted focus away from ConvNets as researchers sought new paradigms for tackling multimodal data.  Such results signify a \emph{``\textbf{comeback}''} for ConvNet in its original domain and showcase large-kernel ConvNet's potential to \emph{``\textbf{conquer}''} new territories. We hope this advancement will inspire further research into large-kernel ConvNets, encouraging new applications and optimizations that extend ConvNets’ utility across a broader range of data modalities.

{\small
	\bibliographystyle{IEEEtran}
	\bibliography{egbib}

\begin{thebibliography}{100}
\providecommand{\url}[1]{#1}
\csname url@samestyle\endcsname
\providecommand{\newblock}{\relax}
\providecommand{\bibinfo}[2]{#2}
\providecommand{\BIBentrySTDinterwordspacing}{\spaceskip=0pt\relax}
\providecommand{\BIBentryALTinterwordstretchfactor}{4}
\providecommand{\BIBentryALTinterwordspacing}{\spaceskip=\fontdimen2\font plus
\BIBentryALTinterwordstretchfactor\fontdimen3\font minus \fontdimen4\font\relax}
\providecommand{\BIBforeignlanguage}[2]{{%
\expandafter\ifx\csname l@#1\endcsname\relax
\typeout{** WARNING: IEEEtran.bst: No hyphenation pattern has been}%
\typeout{** loaded for the language `#1'. Using the pattern for}%
\typeout{** the default language instead.}%
\else
\language=\csname l@#1\endcsname
\fi
#2}}
\providecommand{\BIBdecl}{\relax}
\BIBdecl

\bibitem{ding2022scaling}
X.~Ding, X.~Zhang, J.~Han, and G.~Ding, ``Scaling up your kernels to 31x31: Revisiting large kernel design in cnns,'' in \emph{Proceedings of the IEEE/CVF conference on computer vision and pattern recognition}, 2022, pp. 11\,963--11\,975.

\bibitem{ding2024unireplknet}
X.~Ding, Y.~Zhang, Y.~Ge, S.~Zhao, L.~Song, X.~Yue, and Y.~Shan, ``Unireplknet: A universal perception large-kernel convnet for audio video point cloud time-series and image recognition,'' in \emph{Proceedings of the IEEE/CVF Conference on Computer Vision and Pattern Recognition}, 2024, pp. 5513--5524.

\bibitem{krizhevsky2012imagenet}
A.~Krizhevsky, I.~Sutskever, and G.~E. Hinton, ``Imagenet classification with deep convolutional neural networks,'' in \emph{Advances in neural information processing systems}, 2012, pp. 1097--1105.

\bibitem{szegedy2015going}
C.~Szegedy, W.~Liu, Y.~Jia, P.~Sermanet, S.~Reed, D.~Anguelov, D.~Erhan, V.~Vanhoucke, and A.~Rabinovich, ``Going deeper with convolutions,'' in \emph{Proceedings of the IEEE conference on computer vision and pattern recognition}, 2015, pp. 1--9.

\bibitem{szegedy2016rethinking}
C.~Szegedy, V.~Vanhoucke, S.~Ioffe, J.~Shlens, and Z.~Wojna, ``Rethinking the inception architecture for computer vision,'' in \emph{Proceedings of the IEEE conference on computer vision and pattern recognition}, 2016, pp. 2818--2826.

\bibitem{szegedy2017inception}
C.~Szegedy, S.~Ioffe, V.~Vanhoucke, and A.~A. Alemi, ``Inception-v4, inception-resnet and the impact of residual connections on learning,'' in \emph{Thirty-first AAAI conference on artificial intelligence}, 2017.

\bibitem{he2016deep}
K.~He, X.~Zhang, S.~Ren, and J.~Sun, ``Deep residual learning for image recognition,'' in \emph{Proceedings of the IEEE conference on computer vision and pattern recognition}, 2016, pp. 770--778.

\bibitem{huang2017densely}
\BIBentryALTinterwordspacing
G.~Huang, Z.~Liu, L.~van~der Maaten, and K.~Q. Weinberger, ``Densely connected convolutional networks,'' in \emph{2017 {IEEE} Conference on Computer Vision and Pattern Recognition, {CVPR} 2017, Honolulu, HI, USA, July 21-26, 2017}.\hskip 1em plus 0.5em minus 0.4em\relax {IEEE} Computer Society, 2017, pp. 2261--2269. [Online]. Available: \url{https://doi.org/10.1109/CVPR.2017.243}
\BIBentrySTDinterwordspacing

\bibitem{lecun1995convolutional}
Y.~LeCun, Y.~Bengio \emph{et~al.}, ``Convolutional networks for images, speech, and time series,'' \emph{The handbook of brain theory and neural networks}, vol. 3361, no.~10, p. 1995, 1995.

\bibitem{caffe-lenet}
J.~Yangqing, ``{Caffe LeNet-5},'' \url{https://github.com/BVLC/caffe/tree/master/examples/mnist/}, 2014.

\bibitem{zhang2018shufflenet}
X.~Zhang, X.~Zhou, M.~Lin, and J.~Sun, ``Shufflenet: An extremely efficient convolutional neural network for mobile devices,'' in \emph{Proceedings of the IEEE conference on computer vision and pattern recognition}, 2018, pp. 6848--6856.

\bibitem{ding2021repvgg}
X.~Ding, X.~Zhang, N.~Ma, J.~Han, G.~Ding, and J.~Sun, ``Repvgg: Making vgg-style convnets great again,'' in \emph{Proceedings of the IEEE/CVF Conference on Computer Vision and Pattern Recognition}, 2021, pp. 13\,733--13\,742.

\bibitem{chen2017deeplab}
L.-C. Chen, G.~Papandreou, I.~Kokkinos, K.~Murphy, and A.~L. Yuille, ``Deeplab: Semantic image segmentation with deep convolutional nets, atrous convolution, and fully connected crfs,'' \emph{IEEE transactions on pattern analysis and machine intelligence}, vol.~40, no.~4, pp. 834--848, 2017.

\bibitem{chollet2017xception}
F.~Chollet, ``Xception: Deep learning with depthwise separable convolutions,'' in \emph{Proceedings of the IEEE conference on computer vision and pattern recognition}, 2017, pp. 1251--1258.

\bibitem{dai2017deformable}
J.~Dai, H.~Qi, Y.~Xiong, Y.~Li, G.~Zhang, H.~Hu, and Y.~Wei, ``Deformable convolutional networks,'' in \emph{Proceedings of the IEEE international conference on computer vision}, 2017, pp. 764--773.

\bibitem{liu2022convnet}
Z.~Liu, H.~Mao, C.-Y. Wu, C.~Feichtenhofer, T.~Darrell, and S.~Xie, ``A convnet for the 2020s,'' \emph{arXiv preprint arXiv:2201.03545}, 2022.

\bibitem{mbv1}
A.~G. Howard, M.~Zhu, B.~Chen, D.~Kalenichenko, W.~Wang, T.~Weyand, M.~Andreetto, and H.~Adam, ``Mobilenets: Efficient convolutional neural networks for mobile vision applications,'' \emph{arXiv preprint arXiv:1704.04861}, 2017.

\bibitem{simonyan2014very}
K.~Simonyan and A.~Zisserman, ``Very deep convolutional networks for large-scale image recognition,'' \emph{arXiv preprint arXiv:1409.1556}, 2014.

\bibitem{vit}
\BIBentryALTinterwordspacing
A.~Dosovitskiy, L.~Beyer, A.~Kolesnikov, D.~Weissenborn, X.~Zhai, T.~Unterthiner, M.~Dehghani, M.~Minderer, G.~Heigold, S.~Gelly, J.~Uszkoreit, and N.~Houlsby, ``An image is worth 16x16 words: Transformers for image recognition at scale,'' in \emph{9th International Conference on Learning Representations, {ICLR} 2021, Virtual Event, Austria, May 3-7, 2021}.\hskip 1em plus 0.5em minus 0.4em\relax OpenReview.net, 2021. [Online]. Available: \url{https://openreview.net/forum?id=YicbFdNTTy}
\BIBentrySTDinterwordspacing

\bibitem{swin}
Z.~Liu, Y.~Lin, Y.~Cao, H.~Hu, Y.~Wei, Z.~Zhang, S.~Lin, and B.~Guo, ``Swin transformer: Hierarchical vision transformer using shifted windows,'' in \emph{Proceedings of the IEEE/CVF International Conference on Computer Vision}, 2021, pp. 10\,012--10\,022.

\bibitem{deit}
H.~Touvron, M.~Cord, M.~Douze, F.~Massa, A.~Sablayrolles, and H.~J{\'e}gou, ``Training data-efficient image transformers \& distillation through attention,'' in \emph{International Conference on Machine Learning}.\hskip 1em plus 0.5em minus 0.4em\relax PMLR, 2021, pp. 10\,347--10\,357.

\bibitem{pvt}
W.~Wang, E.~Xie, X.~Li, D.-P. Fan, K.~Song, D.~Liang, T.~Lu, P.~Luo, and L.~Shao, ``Pyramid vision transformer: A versatile backbone for dense prediction without convolutions,'' \emph{arXiv preprint arXiv:2102.12122}, 2021.

\bibitem{ge2023advancing}
C.~Ge, X.~Ding, Z.~Tong, L.~Yuan, J.~Wang, Y.~Song, and P.~Luo, ``Advancing vision transformers with group-mix attention,'' \emph{arXiv preprint arXiv:2311.15157}, 2023.

\bibitem{bot}
A.~Srinivas, T.-Y. Lin, N.~Parmar, J.~Shlens, P.~Abbeel, and A.~Vaswani, ``Bottleneck transformers for visual recognition,'' in \emph{Proceedings of the IEEE/CVF Conference on Computer Vision and Pattern Recognition}, 2021, pp. 16\,519--16\,529.

\bibitem{halonet}
A.~Vaswani, P.~Ramachandran, A.~Srinivas, N.~Parmar, B.~Hechtman, and J.~Shlens, ``Scaling local self-attention for parameter efficient visual backbones,'' in \emph{Proceedings of the IEEE/CVF Conference on Computer Vision and Pattern Recognition}, 2021, pp. 12\,894--12\,904.

\bibitem{sasa}
P.~Ramachandran, N.~Parmar, A.~Vaswani, I.~Bello, A.~Levskaya, and J.~Shlens, ``Stand-alone self-attention in vision models,'' \emph{arXiv preprint arXiv:1906.05909}, 2019.

\bibitem{girdhar2023imagebind}
R.~Girdhar, A.~El-Nouby, Z.~Liu, M.~Singh, K.~V. Alwala, A.~Joulin, and I.~Misra, ``Imagebind: One embedding space to bind them all,'' in \emph{Proceedings of the IEEE/CVF Conference on Computer Vision and Pattern Recognition}, 2023, pp. 15\,180--15\,190.

\bibitem{zhang2023meta}
Y.~Zhang, K.~Gong, K.~Zhang, H.~Li, Y.~Qiao, W.~Ouyang, and X.~Yue, ``Meta-transformer: A unified framework for multimodal learning,'' \emph{arXiv preprint arXiv:2307.10802}, 2023.

\bibitem{han2024onellm}
J.~Han, K.~Gong, Y.~Zhang, J.~Wang, K.~Zhang, D.~Lin, Y.~Qiao, P.~Gao, and X.~Yue, ``Onellm: One framework to align all modalities with language,'' in \emph{Proceedings of the IEEE/CVF Conference on Computer Vision and Pattern Recognition}, 2024, pp. 26\,584--26\,595.

\bibitem{gong2021ast}
Y.~Gong, Y.-A. Chung, and J.~Glass, ``Ast: Audio spectrogram transformer,'' \emph{arXiv preprint arXiv:2104.01778}, 2021.

\bibitem{zhao2021point}
H.~Zhao, L.~Jiang, J.~Jia, P.~Torr, and V.~Koltun, ``Point transformer,'' in \emph{ICCV}, 2021.

\bibitem{li2022mvitv2}
Y.~Li, C.-Y. Wu, H.~Fan, K.~Mangalam, B.~Xiong, J.~Malik, and C.~Feichtenhofer, ``Mvitv2: Improved multiscale vision transformers for classification and detection,'' in \emph{Proceedings of the IEEE/CVF Conference on Computer Vision and Pattern Recognition}, 2022, pp. 4804--4814.

\bibitem{hinton2021represent}
G.~Hinton, ``How to represent part-whole hierarchies in a neural network,'' \emph{arXiv preprint arXiv:2102.12627}, 2021.

\bibitem{zhu2019empirical}
X.~Zhu, D.~Cheng, Z.~Zhang, S.~Lin, and J.~Dai, ``An empirical study of spatial attention mechanisms in deep networks,'' in \emph{Proceedings of the IEEE/CVF International Conference on Computer Vision}, 2019, pp. 6688--6697.

\bibitem{han2021demystifying}
Q.~Han, Z.~Fan, Q.~Dai, L.~Sun, M.-M. Cheng, J.~Liu, and J.~Wang, ``Demystifying local vision transformer: Sparse connectivity, weight sharing, and dynamic weight,'' \emph{arXiv preprint arXiv:2106.04263}, 2021.

\bibitem{wu2019pay}
F.~Wu, A.~Fan, A.~Baevski, Y.~N. Dauphin, and M.~Auli, ``Pay less attention with lightweight and dynamic convolutions,'' \emph{arXiv preprint arXiv:1901.10430}, 2019.

\bibitem{cordonnier2019relationship}
J.-B. Cordonnier, A.~Loukas, and M.~Jaggi, ``On the relationship between self-attention and convolutional layers,'' \emph{arXiv preprint arXiv:1911.03584}, 2019.

\bibitem{xu2014deep}
L.~Xu, J.~S. Ren, C.~Liu, and J.~Jia, ``Deep convolutional neural network for image deconvolution,'' \emph{Advances in neural information processing systems}, vol.~27, 2014.

\bibitem{peng2017large}
C.~Peng, X.~Zhang, G.~Yu, G.~Luo, and J.~Sun, ``Large kernel matters--improve semantic segmentation by global convolutional network,'' in \emph{Proceedings of the IEEE conference on computer vision and pattern recognition}, 2017, pp. 4353--4361.

\bibitem{liu2022more}
S.~Liu, T.~Chen, X.~Chen, X.~Chen, Q.~Xiao, B.~Wu, T.~K{\"a}rkk{\"a}inen, M.~Pechenizkiy, D.~Mocanu, and Z.~Wang, ``More convnets in the 2020s: Scaling up kernels beyond 51x51 using sparsity,'' \emph{arXiv preprint arXiv:2207.03620}, 2022.

\bibitem{liu2021swin}
Z.~Liu, H.~Hu, Y.~Lin, Z.~Yao, Z.~Xie, Y.~Wei, J.~Ning, Y.~Cao, Z.~Zhang, L.~Dong \emph{et~al.}, ``Swin transformer v2: Scaling up capacity and resolution,'' \emph{arXiv preprint arXiv:2111.09883}, 2021.

\bibitem{erf}
\BIBentryALTinterwordspacing
W.~Luo, Y.~Li, R.~Urtasun, and R.~S. Zemel, ``Understanding the effective receptive field in deep convolutional neural networks,'' in \emph{Advances in Neural Information Processing Systems 29: Annual Conference on Neural Information Processing Systems 2016, December 5-10, 2016, Barcelona, Spain}, D.~D. Lee, M.~Sugiyama, U.~von Luxburg, I.~Guyon, and R.~Garnett, Eds., 2016, pp. 4898--4906. [Online]. Available: \url{https://proceedings.neurips.cc/paper/2016/hash/c8067ad1937f728f51288b3eb986afaa-Abstract.html}
\BIBentrySTDinterwordspacing

\bibitem{deng2009imagenet}
J.~Deng, W.~Dong, R.~Socher, L.-J. Li, K.~Li, and L.~Fei-Fei, ``Imagenet: A large-scale hierarchical image database,'' in \emph{Computer Vision and Pattern Recognition, 2009. CVPR 2009. IEEE Conference on}.\hskip 1em plus 0.5em minus 0.4em\relax IEEE, 2009, pp. 248--255.

\bibitem{gemmeke2017audio}
J.~F. Gemmeke, D.~P. Ellis, D.~Freedman, A.~Jansen, W.~Lawrence, R.~C. Moore, M.~Plakal, and M.~Ritter, ``Audio set: An ontology and human-labeled dataset for audio events,'' in \emph{2017 IEEE international conference on acoustics, speech and signal processing (ICASSP)}.\hskip 1em plus 0.5em minus 0.4em\relax IEEE, 2017, pp. 776--780.

\bibitem{uy-scanobjectnn-iccv19}
M.~A. Uy, Q.-H. Pham, B.-S. Hua, D.~T. Nguyen, and S.-K. Yeung, ``Revisiting point cloud classification: A new benchmark dataset and classification model on real-world data,'' in \emph{International Conference on Computer Vision (ICCV)}, 2019.

\bibitem{wu2023interpretable}
H.~Wu, H.~Zhou, M.~Long, and J.~Wang, ``Interpretable weather forecasting for worldwide stations with a unified deep model,'' \emph{Nature Machine Intelligence}, pp. 1--10, 2023.

\bibitem{woo2023convnext}
S.~Woo, S.~Debnath, R.~Hu, X.~Chen, Z.~Liu, I.~S. Kweon, and S.~Xie, ``Convnext v2: Co-designing and scaling convnets with masked autoencoders,'' in \emph{Proceedings of the IEEE/CVF Conference on Computer Vision and Pattern Recognition}, 2023, pp. 16\,133--16\,142.

\bibitem{vasu2023fastvit}
P.~K.~A. Vasu, J.~Gabriel, J.~Zhu, O.~Tuzel, and A.~Ranjan, ``Fastvit: A fast hybrid vision transformer using structural reparameterization,'' \emph{arXiv preprint arXiv:2303.14189}, 2023.

\bibitem{liu2022swin}
Z.~Liu, H.~Hu, Y.~Lin, Z.~Yao, Z.~Xie, Y.~Wei, J.~Ning, Y.~Cao, Z.~Zhang, L.~Dong \emph{et~al.}, ``Swin transformer v2: Scaling up capacity and resolution,'' in \emph{Proceedings of the IEEE/CVF conference on computer vision and pattern recognition}, 2022, pp. 12\,009--12\,019.

\bibitem{touvron2022deit}
H.~Touvron, M.~Cord, and H.~J{\'e}gou, ``Deit iii: Revenge of the vit,'' in \emph{European Conference on Computer Vision}.\hskip 1em plus 0.5em minus 0.4em\relax Springer, 2022, pp. 516--533.

\bibitem{tuli2021convolutional}
S.~Tuli, I.~Dasgupta, E.~Grant, and T.~L. Griffiths, ``Are convolutional neural networks or transformers more like human vision?'' \emph{arXiv preprint arXiv:2105.07197}, 2021.

\bibitem{modelvshuman}
bethgelab, ``Toolbox of model-vs-human,'' \url{https://github.com/bethgelab/model-vs-human}, 2022.

\bibitem{schuhmann2022laion}
C.~Schuhmann, R.~Beaumont, R.~Vencu, C.~Gordon, R.~Wightman, M.~Cherti, T.~Coombes, A.~Katta, C.~Mullis, M.~Wortsman \emph{et~al.}, ``Laion-5b: An open large-scale dataset for training next generation image-text models,'' \emph{Advances in Neural Information Processing Systems}, vol.~35, pp. 25\,278--25\,294, 2022.

\bibitem{openclip}
\BIBentryALTinterwordspacing
G.~Ilharco, M.~Wortsman, R.~Wightman, C.~Gordon, N.~Carlini, R.~Taori, A.~Dave, V.~Shankar, H.~Namkoong, J.~Miller, H.~Hajishirzi, A.~Farhadi, and L.~Schmidt, ``Openclip,'' Jul. 2021, if you use this software, please cite it as below. [Online]. Available: \url{https://doi.org/10.5281/zenodo.5143773}
\BIBentrySTDinterwordspacing

\bibitem{zhang2024multimodal}
Y.~Zhang, X.~Ding, K.~Gong, Y.~Ge, Y.~Shan, and X.~Yue, ``Multimodal pathway: Improve transformers with irrelevant data from other modalities,'' \emph{arXiv preprint arXiv:2401.14405}, 2024.

\bibitem{clip}
C.~Jia, Y.~Yang, Y.~Xia, Y.-T. Chen, Z.~Parekh, H.~Pham, Q.~V. Le, Y.~Sung, Z.~Li, and T.~Duerig, ``Scaling up visual and vision-language representation learning with noisy text supervision,'' \emph{arXiv preprint arXiv:2102.05918}, 2021.

\bibitem{hu2019local}
H.~Hu, Z.~Zhang, Z.~Xie, and S.~Lin, ``Local relation networks for image recognition,'' in \emph{Proceedings of the IEEE/CVF International Conference on Computer Vision}, 2019, pp. 3464--3473.

\bibitem{yu2021metaformer}
W.~Yu, M.~Luo, P.~Zhou, C.~Si, Y.~Zhou, X.~Wang, J.~Feng, and S.~Yan, ``Metaformer is actually what you need for vision,'' \emph{arXiv preprint arXiv:2111.11418}, 2021.

\bibitem{rao2021global}
Y.~Rao, W.~Zhao, Z.~Zhu, J.~Lu, and J.~Zhou, ``Global filter networks for image classification,'' \emph{arXiv preprint arXiv:2107.00645}, 2021.

\bibitem{chen2023largekernel3d}
Y.~Chen, J.~Liu, X.~Zhang, X.~Qi, and J.~Jia, ``Largekernel3d: Scaling up kernels in 3d sparse cnns,'' in \emph{Proceedings of the IEEE/CVF Conference on Computer Vision and Pattern Recognition}, 2023, pp. 13\,488--13\,498.

\bibitem{luo2023lkd}
P.~Luo, G.~Xiao, X.~Gao, and S.~Wu, ``Lkd-net: Large kernel convolution network for single image dehazing,'' in \emph{2023 IEEE International Conference on Multimedia and Expo (ICME)}.\hskip 1em plus 0.5em minus 0.4em\relax IEEE, 2023, pp. 1601--1606.

\bibitem{xie2023large}
C.~Xie, X.~Zhang, L.~Li, H.~Meng, T.~Zhang, T.~Li, and X.~Zhao, ``Large kernel distillation network for efficient single image super-resolution,'' in \emph{Proceedings of the IEEE/CVF Conference on Computer Vision and Pattern Recognition}, 2023, pp. 1283--1292.

\bibitem{hu2018squeeze}
J.~Hu, L.~Shen, and G.~Sun, ``Squeeze-and-excitation networks,'' in \emph{Proceedings of the IEEE conference on computer vision and pattern recognition}, 2018, pp. 7132--7141.

\bibitem{ioffe2015batch}
S.~Ioffe and C.~Szegedy, ``Batch normalization: Accelerating deep network training by reducing internal covariate shift,'' in \emph{International Conference on Machine Learning}, 2015, pp. 448--456.

\bibitem{mbv2}
M.~Sandler, A.~Howard, M.~Zhu, A.~Zhmoginov, and L.-C. Chen, ``Mobilenetv2: Inverted residuals and linear bottlenecks,'' in \emph{Proceedings of the IEEE conference on computer vision and pattern recognition}, 2018, pp. 4510--4520.

\bibitem{ma2018shufflenet}
N.~Ma, X.~Zhang, H.-T. Zheng, and J.~Sun, ``Shufflenet v2: Practical guidelines for efficient cnn architecture design,'' in \emph{Proceedings of the European conference on computer vision (ECCV)}, 2018, pp. 116--131.

\bibitem{veit2016residual}
A.~Veit, M.~J. Wilber, and S.~Belongie, ``Residual networks behave like ensembles of relatively shallow networks,'' in \emph{Advances in neural information processing systems}, 2016, pp. 550--558.

\bibitem{ding2019acnet}
X.~Ding, Y.~Guo, G.~Ding, and J.~Han, ``Acnet: Strengthening the kernel skeletons for powerful cnn via asymmetric convolution blocks,'' in \emph{Proceedings of the IEEE International Conference on Computer Vision}, 2019, pp. 1911--1920.

\bibitem{ding2021repmlpnet}
X.~Ding, H.~Chen, X.~Zhang, J.~Han, and G.~Ding, ``Repmlpnet: Hierarchical vision mlp with re-parameterized locality,'' \emph{arXiv preprint arXiv:2112.11081}, 2021.

\bibitem{chen2018encoder}
L.-C. Chen, Y.~Zhu, G.~Papandreou, F.~Schroff, and H.~Adam, ``Encoder-decoder with atrous separable convolution for semantic image segmentation,'' in \emph{Proceedings of the European conference on computer vision (ECCV)}, 2018, pp. 801--818.

\bibitem{cityscapes}
\BIBentryALTinterwordspacing
M.~Cordts, M.~Omran, S.~Ramos, T.~Rehfeld, M.~Enzweiler, R.~Benenson, U.~Franke, S.~Roth, and B.~Schiele, ``The cityscapes dataset for semantic urban scene understanding,'' in \emph{2016 {IEEE} Conference on Computer Vision and Pattern Recognition, {CVPR} 2016, Las Vegas, NV, USA, June 27-30, 2016}.\hskip 1em plus 0.5em minus 0.4em\relax {IEEE} Computer Society, 2016, pp. 3213--3223. [Online]. Available: \url{https://doi.org/10.1109/CVPR.2016.350}
\BIBentrySTDinterwordspacing

\bibitem{mmseg2020}
M.~Contributors, ``{MMSegmentation}: Openmmlab semantic segmentation toolbox and benchmark,'' \url{https://github.com/open-mmlab/mmsegmentation}, 2020.

\bibitem{shaw2018self}
P.~Shaw, J.~Uszkoreit, and A.~Vaswani, ``Self-attention with relative position representations,'' \emph{arXiv preprint arXiv:1803.02155}, 2018.

\bibitem{bello2019attention}
I.~Bello, B.~Zoph, A.~Vaswani, J.~Shlens, and Q.~V. Le, ``Attention augmented convolutional networks,'' in \emph{Proceedings of the IEEE/CVF international conference on computer vision}, 2019, pp. 3286--3295.

\bibitem{kayhan2020translation}
O.~S. Kayhan and J.~C.~v. Gemert, ``On translation invariance in cnns: Convolutional layers can exploit absolute spatial location,'' in \emph{Proceedings of the IEEE/CVF Conference on Computer Vision and Pattern Recognition}, 2020, pp. 14\,274--14\,285.

\bibitem{long2015fully}
J.~Long, E.~Shelhamer, and T.~Darrell, ``Fully convolutional networks for semantic segmentation,'' in \emph{Proceedings of the IEEE Conference on Computer Vision and Pattern Recognition}, 2015, pp. 3431--3440.

\bibitem{yu2017dilated}
F.~Yu, V.~Koltun, and T.~Funkhouser, ``Dilated residual networks,'' in \emph{Proceedings of the IEEE conference on computer vision and pattern recognition}, 2017, pp. 472--480.

\bibitem{wang2020deep}
J.~Wang, K.~Sun, T.~Cheng, B.~Jiang, C.~Deng, Y.~Zhao, D.~Liu, Y.~Mu, M.~Tan, X.~Wang \emph{et~al.}, ``Deep high-resolution representation learning for visual recognition,'' \emph{IEEE transactions on pattern analysis and machine intelligence}, 2020.

\bibitem{yu2015multi}
F.~Yu and V.~Koltun, ``Multi-scale context aggregation by dilated convolutions,'' \emph{arXiv preprint arXiv:1511.07122}, 2015.

\bibitem{geirhos2018imagenet}
R.~Geirhos, P.~Rubisch, C.~Michaelis, M.~Bethge, F.~A. Wichmann, and W.~Brendel, ``Imagenet-trained cnns are biased towards texture; increasing shape bias improves accuracy and robustness,'' \emph{arXiv preprint arXiv:1811.12231}, 2018.

\bibitem{brendel2019approximating}
W.~Brendel and M.~Bethge, ``Approximating cnns with bag-of-local-features models works surprisingly well on imagenet,'' \emph{arXiv preprint arXiv:1904.00760}, 2019.

\bibitem{ba2016layer}
J.~L. Ba, J.~R. Kiros, and G.~E. Hinton, ``Layer normalization,'' \emph{arXiv preprint arXiv:1607.06450}, 2016.

\bibitem{xiao2018unified}
T.~Xiao, Y.~Liu, B.~Zhou, Y.~Jiang, and J.~Sun, ``Unified perceptual parsing for scene understanding,'' in \emph{Proceedings of the European Conference on Computer Vision (ECCV)}, 2018, pp. 418--434.

\bibitem{vaswani2017attention}
A.~Vaswani, N.~Shazeer, N.~Parmar, J.~Uszkoreit, L.~Jones, A.~N. Gomez, {\L}.~Kaiser, and I.~Polosukhin, ``Attention is all you need,'' in \emph{Advances in neural information processing systems}, 2017, pp. 5998--6008.

\bibitem{han2024asymmetric}
W.~Han, X.~Dong, Y.~Zhang, D.~Crandall, C.-Z. Xu, and J.~Shen, ``Asymmetric convolution: An efficient and generalized method to fuse feature maps in multiple vision tasks,'' \emph{IEEE Transactions on Pattern Analysis and Machine Intelligence}, 2024.

\bibitem{wang2022pvt}
W.~Wang, E.~Xie, X.~Li, D.-P. Fan, K.~Song, D.~Liang, T.~Lu, P.~Luo, and L.~Shao, ``Pvt v2: Improved baselines with pyramid vision transformer,'' \emph{Computational Visual Media}, vol.~8, no.~3, pp. 415--424, 2022.

\bibitem{dai2021coatnet}
Z.~Dai, H.~Liu, Q.~V. Le, and M.~Tan, ``Coatnet: Marrying convolution and attention for all data sizes,'' \emph{Advances in neural information processing systems}, vol.~34, pp. 3965--3977, 2021.

\bibitem{wang2023internimage}
W.~Wang, J.~Dai, Z.~Chen, Z.~Huang, Z.~Li, X.~Zhu, X.~Hu, T.~Lu, L.~Lu, H.~Li \emph{et~al.}, ``Internimage: Exploring large-scale vision foundation models with deformable convolutions,'' in \emph{Proceedings of the IEEE/CVF Conference on Computer Vision and Pattern Recognition}, 2023, pp. 14\,408--14\,419.

\bibitem{rao2022hornet}
Y.~Rao, W.~Zhao, Y.~Tang, J.~Zhou, S.~N. Lim, and J.~Lu, ``Hornet: Efficient high-order spatial interactions with recursive gated convolutions,'' \emph{Advances in Neural Information Processing Systems}, vol.~35, pp. 10\,353--10\,366, 2022.

\bibitem{kong2020panns}
Q.~Kong, Y.~Cao, T.~Iqbal, Y.~Wang, W.~Wang, and M.~D. Plumbley, ``Panns: Large-scale pretrained audio neural networks for audio pattern recognition,'' \emph{IEEE/ACM Transactions on Audio, Speech, and Language Processing}, vol.~28, pp. 2880--2894, 2020.

\bibitem{gong2021psla}
Y.~Gong, Y.-A. Chung, and J.~Glass, ``Psla: Improving audio tagging with pretraining, sampling, labeling, and aggregation,'' \emph{IEEE/ACM Transactions on Audio, Speech, and Language Processing}, vol.~29, pp. 3292--3306, 2021.

\bibitem{gong2022ssast}
Y.~Gong, C.-I. Lai, Y.-A. Chung, and J.~Glass, ``Ssast: Self-supervised audio spectrogram transformer,'' in \emph{Proceedings of the AAAI Conference on Artificial Intelligence}, vol.~36, no.~10, 2022, pp. 10\,699--10\,709.

\bibitem{huang2022masked}
P.-Y. Huang, H.~Xu, J.~Li, A.~Baevski, M.~Auli, W.~Galuba, F.~Metze, and C.~Feichtenhofer, ``Masked autoencoders that listen,'' \emph{Advances in Neural Information Processing Systems}, vol.~35, pp. 28\,708--28\,720, 2022.

\bibitem{feichtenhofer2019slowfast}
C.~Feichtenhofer, H.~Fan, J.~Malik, and K.~He, ``Slowfast networks for video recognition,'' in \emph{Proceedings of the IEEE/CVF international conference on computer vision}, 2019, pp. 6202--6211.

\bibitem{bertasius2021space}
G.~Bertasius, H.~Wang, and L.~Torresani, ``Is space-time attention all you need for video understanding?'' in \emph{ICML}, vol.~2, no.~3, 2021, p.~4.

\bibitem{qi2017pointnet}
C.~R. Qi, H.~Su, K.~Mo, and L.~J. Guibas, ``Pointnet: Deep learning on point sets for 3d classification and segmentation,'' in \emph{CVPR}, 2017.

\bibitem{qi2017pointnet++}
C.~R. Qi, L.~Yi, H.~Su, and L.~J. Guibas, ``Pointnet++: Deep hierarchical feature learning on point sets in a metric space,'' in \emph{NeurIPS}, 2017.

\bibitem{wu2019pointconv}
W.~Wu, Z.~Qi, and L.~Fuxin, ``Pointconv: Deep convolutional networks on 3d point clouds,'' in \emph{CVPR}, 2019.

\bibitem{thomas2019kpconv}
H.~Thomas, C.~R. Qi, J.-E. Deschaud, B.~Marcotegui, F.~Goulette, and L.~J. Guibas, ``Kpconv: Flexible and deformable convolution for point clouds,'' in \emph{ICCV}, 2019.

\bibitem{wang2019dynamic}
Y.~Wang, Y.~Sun, Z.~Liu, S.~E. Sarma, M.~M. Bronstein, and J.~M. Solomon, ``Dynamic graph cnn for learning on point clouds,'' \emph{TOG}, 2019.

\bibitem{liu2023openshape}
M.~Liu, R.~Shi, K.~Kuang, Y.~Zhu, X.~Li, S.~Han, H.~Cai, F.~Porikli, and H.~Su, ``Openshape: Scaling up 3d shape representation towards open-world understanding,'' \emph{arXiv preprint arXiv:2305.10764}, 2023.

\bibitem{he2017mask}
K.~He, G.~Gkioxari, P.~Doll{\'a}r, and R.~Girshick, ``Mask r-cnn,'' in \emph{Proceedings of the IEEE international conference on computer vision}, 2017, pp. 2961--2969.

\bibitem{cai2019cascade}
Z.~Cai and N.~Vasconcelos, ``Cascade r-cnn: High quality object detection and instance segmentation,'' \emph{IEEE Transactions on Pattern Analysis and Machine Intelligence}, 2019.

\bibitem{mmdetection}
K.~Chen, J.~Wang, J.~Pang, Y.~Cao, Y.~Xiong, X.~Li, S.~Sun, W.~Feng, Z.~Liu, J.~Xu, Z.~Zhang, D.~Cheng, C.~Zhu, T.~Cheng, Q.~Zhao, B.~Li, X.~Lu, R.~Zhu, Y.~Wu, J.~Dai, J.~Wang, J.~Shi, W.~Ouyang, C.~C. Loy, and D.~Lin, ``{MMDetection}: Open mmlab detection toolbox and benchmark,'' \emph{arXiv preprint arXiv:1906.07155}, 2019.

\bibitem{zhou2019semantic}
B.~Zhou, H.~Zhao, X.~Puig, T.~Xiao, S.~Fidler, A.~Barriuso, and A.~Torralba, ``Semantic understanding of scenes through the ade20k dataset,'' \emph{International Journal of Computer Vision}, vol. 127, no.~3, pp. 302--321, 2019.

\bibitem{hyndman2017forecasting}
R.~J. Hyndman and G.~Athanasopoulos, ``Forecasting: Principles and practice. otexts; 2014,'' \emph{Online at http://otexts. org/fpp}, 2017.

\bibitem{taylor2018forecasting}
S.~J. Taylor and B.~Letham, ``Forecasting at scale,'' \emph{The American Statistician}, vol.~72, no.~1, pp. 37--45, 2018.

\bibitem{ke2017lightgbm}
G.~Ke, Q.~Meng, T.~Finley, T.~Wang, W.~Chen, W.~Ma, Q.~Ye, and T.-Y. Liu, ``Lightgbm: A highly efficient gradient boosting decision tree,'' \emph{Advances in neural information processing systems}, vol.~30, 2017.

\bibitem{hersbach2020era5}
H.~Hersbach, B.~Bell, P.~Berrisford, S.~Hirahara, A.~Hor{\'a}nyi, J.~Mu{\~n}oz-Sabater, J.~Nicolas, C.~Peubey, R.~Radu, D.~Schepers \emph{et~al.}, ``The era5 global reanalysis, qj roy. meteor. soc., 146, 1999--2049,'' 2020.

\bibitem{salinas2020deepar}
D.~Salinas, V.~Flunkert, J.~Gasthaus, and T.~Januschowski, ``Deepar: Probabilistic forecasting with autoregressive recurrent networks,'' \emph{International Journal of Forecasting}, vol.~36, no.~3, pp. 1181--1191, 2020.

\bibitem{oreshkin2019n}
B.~N. Oreshkin, D.~Carpov, N.~Chapados, and Y.~Bengio, ``N-beats: Neural basis expansion analysis for interpretable time series forecasting,'' in \emph{International Conference on Learning Representations}, 2019.

\bibitem{cao2020spectral}
D.~Cao, Y.~Wang, J.~Duan, C.~Zhang, X.~Zhu, C.~Huang, Y.~Tong, B.~Xu, J.~Bai, J.~Tong \emph{et~al.}, ``Spectral temporal graph neural network for multivariate time-series forecasting,'' \emph{Advances in neural information processing systems}, vol.~33, pp. 17\,766--17\,778, 2020.

\bibitem{liu2021pyraformer}
S.~Liu, H.~Yu, C.~Liao, J.~Li, W.~Lin, A.~X. Liu, and S.~Dustdar, ``Pyraformer: Low-complexity pyramidal attention for long-range time series modeling and forecasting,'' in \emph{International conference on learning representations}, 2021.

\bibitem{recht2019imagenetv2}
B.~Recht, R.~Roelofs, L.~Schmidt, and V.~Shankar, ``Do imagenet classifiers generalize to imagenet?'' 2019.

\bibitem{inadv}
D.~Hendrycks, K.~Zhao, S.~Basart, J.~Steinhardt, and D.~Song, ``Natural adversarial examples,'' in \emph{CVPR}, 2021.

\bibitem{inren}
D.~Hendrycks, S.~Basart, N.~Mu, S.~Kadavath, F.~Wang, E.~Dorundo, R.~Desai, T.~Zhu, S.~Parajuli, M.~Guo \emph{et~al.}, ``The many faces of robustness: A critical analysis of out-of-distribution generalization,'' in \emph{CVPR}, 2021.

\bibitem{inske}
H.~Wang, S.~Ge, Z.~Lipton, and E.~P. Xing, ``Learning robust global representations by penalizing local predictive power,'' \emph{NeurIPS}, 2019.

\bibitem{objectnet}
A.~Barbu, D.~Mayo, J.~Alverio, W.~Luo, C.~Wang, D.~Gutfreund, J.~Tenenbaum, and B.~Katz, ``Objectnet: A large-scale bias-controlled dataset for pushing the limits of object recognition models,'' in \emph{NeurIPS}, 2019.

\bibitem{cifar}
A.~Krizhevsky, G.~Hinton \emph{et~al.}, ``Learning multiple layers of features from tiny images,'' 2009.

\bibitem{lecun1998gradient}
Y.~LeCun, L.~Bottou, Y.~Bengio, and P.~Haffner, ``Gradient-based learning applied to document recognition,'' \emph{Proceedings of the IEEE}, vol.~86, no.~11, pp. 2278--2324, 1998.

\bibitem{fei2004learning}
L.~Fei-Fei, R.~Fergus, and P.~Perona, ``Learning generative visual models from few training examples: An incremental bayesian approach tested on 101 object categories,'' in \emph{CVPRW}, 2004.

\bibitem{xiao2010sun}
J.~Xiao, J.~Hays, K.~A. Ehinger, A.~Oliva, and A.~Torralba, ``Sun database: Large-scale scene recognition from abbey to zoo,'' in \emph{CVPR}, 2010.

\bibitem{maji2013fine}
S.~Maji, E.~Rahtu, J.~Kannala, M.~Blaschko, and A.~Vedaldi, ``Fine-grained visual classification of aircraft,'' \emph{arXiv preprint arXiv:1306.5151}, 2013.

\bibitem{krause20133d}
J.~Krause, M.~Stark, J.~Deng, and L.~Fei-Fei, ``3d object representations for fine-grained categorization,'' in \emph{ICCVW}, 2013.

\bibitem{cimpoi14describing}
M.~Cimpoi, S.~Maji, I.~Kokkinos, S.~Mohamed, , and A.~Vedaldi, ``Describing textures in the wild,'' in \emph{CVPR}, 2014.

\bibitem{helber2019eurosat}
P.~Helber, B.~Bischke, A.~Dengel, and D.~Borth, ``Eurosat: A novel dataset and deep learning benchmark for land use and land cover classification,'' \emph{IEEE J. Sel. Top. Appl. Earth Obs. Remote Sens.}, 2019.

\bibitem{goodfellow2013challenges}
I.~J. Goodfellow, D.~Erhan, P.~L. Carrier, A.~Courville, M.~Mirza, B.~Hamner, W.~Cukierski, Y.~Tang, D.~Thaler, D.-H. Lee \emph{et~al.}, ``Challenges in representation learning: A report on three machine learning contests,'' in \emph{ICONIP}, 2013.

\bibitem{nilsback2008automated}
M.-E. Nilsback and A.~Zisserman, ``Automated flower classification over a large number of classes,'' in \emph{ICVGIP}, 2008.

\bibitem{bossard2014food}
L.~Bossard, M.~Guillaumin, and L.~Van~Gool, ``Food-101--mining discriminative components with random forests,'' in \emph{ECCV}, 2014.

\bibitem{stallkamp2012man}
J.~Stallkamp, M.~Schlipsing, J.~Salmen, and C.~Igel, ``Man vs. computer: Benchmarking machine learning algorithms for traffic sign recognition,'' \emph{Neural networks}, 2012.

\bibitem{veeling2018rotation}
B.~S. Veeling, J.~Linmans, J.~Winkens, T.~Cohen, and M.~Welling, ``Rotation equivariant cnns for digital pathology,'' in \emph{MICCAI}, 2018.

\bibitem{parkhi12a}
O.~M. Parkhi, A.~Vedaldi, A.~Zisserman, and C.~V. Jawahar, ``Cats and dogs,'' in \emph{CVPR}, 2012.

\bibitem{cheng2017remote}
G.~Cheng, J.~Han, and X.~Lu, ``Remote sensing image scene classification: Benchmark and state of the art,'' \emph{Proceedings of the IEEE}, 2017.

\bibitem{coates2011analysis}
A.~Coates, A.~Ng, and H.~Lee, ``An analysis of single-layer networks in unsupervised feature learning,'' in \emph{AISTAT}, 2011.

\bibitem{pascal-voc-2007}
M.~Everingham, L.~Van~Gool, C.~K.~I. Williams, J.~Winn, and A.~Zisserman, ``"the {PASCAL} {V}isual {O}bject {C}lasses {C}hallenge 2007 {(VOC2007)} {R}esults,'' "http://www.pascal-network.org/challenges/VOC/voc2007/workshop/index.html", 2007.

\bibitem{radford2021learning}
A.~Radford, J.~W. Kim, C.~Hallacy, A.~Ramesh, G.~Goh, S.~Agarwal, G.~Sastry, A.~Askell, P.~Mishkin, J.~Clark \emph{et~al.}, ``Learning transferable visual models from natural language supervision,'' in \emph{International conference on machine learning}.\hskip 1em plus 0.5em minus 0.4em\relax PMLR, 2021, pp. 8748--8763.

\bibitem{li2023scaling}
Y.~Li, H.~Fan, R.~Hu, C.~Feichtenhofer, and K.~He, ``Scaling language-image pre-training via masking,'' in \emph{Proceedings of the IEEE/CVF Conference on Computer Vision and Pattern Recognition}, 2023, pp. 23\,390--23\,400.

\bibitem{EVA-CLIP}
Q.~Sun, Y.~Fang, L.~Wu, X.~Wang, and Y.~Cao, ``Eva-clip: Improved training techniques for clip at scale,'' \emph{arXiv preprint arXiv:2303.15389}, 2023.

\bibitem{hudson2019gqa}
D.~A. Hudson and C.~D. Manning, ``Gqa: A new dataset for real-world visual reasoning and compositional question answering,'' in \emph{CVPR}, 2019.

\bibitem{goyal2017vqav2}
Y.~Goyal, T.~Khot, D.~Summers-Stay, D.~Batra, and D.~Parikh, ``Making the v in vqa matter: Elevating the role of image understanding in visual question answering,'' in \emph{CVPR}, 2017, pp. 6904--6913.

\bibitem{okvqa}
K.~Marino, M.~Rastegari, A.~Farhadi, and R.~Mottaghi, ``Ok-vqa: A visual question answering benchmark requiring external knowledge,'' in \emph{CVPR}, 2019.

\bibitem{singh2019textvqa}
A.~Singh, V.~Natarajan, M.~Shah, Y.~Jiang, X.~Chen, D.~Batra, D.~Parikh, and M.~Rohrbach, ``Towards vqa models that can read,'' in \emph{CVPR}, 2019, pp. 8317--8326.

\bibitem{lu2022learn}
P.~Lu, S.~Mishra, T.~Xia, L.~Qiu, K.-W. Chang, S.-C. Zhu, O.~Tafjord, P.~Clark, and A.~Kalyan, ``Learn to explain: Multimodal reasoning via thought chains for science question answering,'' \emph{NeurIPS}, 2022.

\bibitem{gurari2018vizwiz}
D.~Gurari, Q.~Li, A.~J. Stangl, A.~Guo, C.~Lin, K.~Grauman, J.~Luo, and J.~P. Bigham, ``Vizwiz grand challenge: Answering visual questions from blind people,'' in \emph{CVPR}, 2018, pp. 3608--3617.

\bibitem{agrawal2019nocaps}
H.~Agrawal, K.~Desai, Y.~Wang, X.~Chen, R.~Jain, M.~Johnson, D.~Batra, D.~Parikh, S.~Lee, and P.~Anderson, ``nocaps: novel object captioning at scale,'' in \emph{ICCV}, 2019.

\bibitem{plummer2015flickr30k}
B.~A. Plummer, L.~Wang, C.~M. Cervantes, J.~C. Caicedo, J.~Hockenmaier, and S.~Lazebnik, ``Flickr30k entities: Collecting region-to-phrase correspondences for richer image-to-sentence models,'' in \emph{ICCV}, 2015, pp. 2641--2649.

\bibitem{fu2023mme}
C.~Fu, P.~Chen, Y.~Shen, Y.~Qin, M.~Zhang, X.~Lin, Z.~Qiu, W.~Lin, J.~Yang, X.~Zheng \emph{et~al.}, ``Mme: A comprehensive evaluation benchmark for multimodal large language models,'' \emph{arXiv preprint arXiv:2306.13394}, 2023.

\bibitem{liu2023mmbench}
Y.~Liu, H.~Duan, Y.~Zhang, B.~Li, S.~Zhang, W.~Zhao, Y.~Yuan, J.~Wang, C.~He, Z.~Liu \emph{et~al.}, ``Mmbench: Is your multi-modal model an all-around player?'' \emph{arXiv preprint arXiv:2307.06281}, 2023.

\bibitem{yu2023mmvet}
W.~Yu, Z.~Yang, L.~Li, J.~Wang, K.~Lin, Z.~Liu, X.~Wang, and L.~Wang, ``Mm-vet: Evaluating large multimodal models for integrated capabilities,'' \emph{arXiv preprint arXiv:2308.02490}, 2023.

\bibitem{li2023seed}
B.~Li, R.~Wang, G.~Wang, Y.~Ge, Y.~Ge, and Y.~Shan, ``Seed-bench: Benchmarking multimodal llms with generative comprehension,'' \emph{arXiv preprint arXiv:2307.16125}, 2023.

\bibitem{alayrac2022flamingo}
J.-B. Alayrac, J.~Donahue, P.~Luc, A.~Miech, I.~Barr, Y.~Hasson, K.~Lenc, A.~Mensch, K.~Millican, M.~Reynolds \emph{et~al.}, ``Flamingo: a visual language model for few-shot learning,'' \emph{NeurIPS}, vol.~35, pp. 23\,716--23\,736, 2022.

\bibitem{li2023blip}
J.~Li, D.~Li, S.~Savarese, and S.~Hoi, ``Blip-2: Bootstrapping language-image pre-training with frozen image encoders and large language models,'' \emph{arXiv preprint arXiv:2301.12597}, 2023.

\bibitem{instructblip}
W.~Dai, J.~Li, D.~Li, A.~M.~H. Tiong, J.~Zhao, W.~Wang, B.~Li, P.~Fung, and S.~Hoi, ``Instructblip: Towards general-purpose vision-language models with instruction tuning,'' 2023.

\bibitem{laurenccon2023obelisc}
H.~Lauren{\c{c}}on, L.~Saulnier, L.~Tronchon, S.~Bekman, A.~Singh, A.~Lozhkov, T.~Wang, S.~Karamcheti, A.~M. Rush, D.~Kiela \emph{et~al.}, ``Obelisc: An open web-scale filtered dataset of interleaved image-text documents,'' \emph{arXiv preprint arXiv:2306.16527}, 2023.

\bibitem{bai2023qwen}
J.~Bai, S.~Bai, S.~Yang, S.~Wang, S.~Tan, P.~Wang, J.~Lin, C.~Zhou, and J.~Zhou, ``Qwen-vl: A frontier large vision-language model with versatile abilities,'' \emph{arXiv preprint arXiv:2308.12966}, 2023.

\bibitem{liu2023improvedllava}
H.~Liu, C.~Li, Y.~Li, and Y.~J. Lee, ``Improved baselines with visual instruction tuning,'' 2023.

\bibitem{han2023imagebind}
J.~Han, R.~Zhang, W.~Shao, P.~Gao, P.~Xu, H.~Xiao, K.~Zhang, C.~Liu, S.~Wen, Z.~Guo \emph{et~al.}, ``Imagebind-llm: Multi-modality instruction tuning,'' \emph{arXiv preprint arXiv:2309.03905}, 2023.

\bibitem{moon2023anymal}
S.~Moon, A.~Madotto, Z.~Lin, T.~Nagarajan, M.~Smith, S.~Jain, C.-F. Yeh, P.~Murugesan, P.~Heidari, Y.~Liu \emph{et~al.}, ``Anymal: An efficient and scalable any-modality augmented language model,'' \emph{arXiv preprint arXiv:2309.16058}, 2023.

\bibitem{warden2018speech}
P.~Warden, ``Speech commands: A dataset for limited-vocabulary speech recognition,'' \emph{arXiv preprint arXiv:1804.03209}, 2018.

\bibitem{kay2017kinetics}
W.~Kay, J.~Carreira, K.~Simonyan, B.~Zhang, C.~Hillier, S.~Vijayanarasimhan, F.~Viola, T.~Green, T.~Back, P.~Natsev \emph{et~al.}, ``The kinetics human action video dataset,'' \emph{arXiv preprint arXiv:1705.06950}, 2017.

\bibitem{wu2015modelnet}
Z.~Wu, S.~Song, A.~Khosla, F.~Yu, L.~Zhang, X.~Tang, and J.~Xiao, ``3d shapenets: A deep representation for volumetric shapes,'' in \emph{CVPR}, 2015.

\bibitem{pont20172017}
J.~Pont-Tuset, F.~Perazzi, S.~Caelles, P.~Arbel{\'a}ez, A.~Sorkine-Hornung, and L.~Van~Gool, ``The 2017 davis challenge on video object segmentation,'' \emph{arXiv preprint arXiv:1704.00675}, 2017.

\bibitem{khoreva2019video}
A.~Khoreva, A.~Rohrbach, and B.~Schiele, ``Video object segmentation with language referring expressions,'' in \emph{Computer Vision--ACCV 2018: 14th Asian Conference on Computer Vision, Perth, Australia, December 2--6, 2018, Revised Selected Papers, Part IV 14}.\hskip 1em plus 0.5em minus 0.4em\relax Springer, 2019, pp. 123--141.

\bibitem{seo2020urvos}
S.~Seo, J.-Y. Lee, and B.~Han, ``Urvos: Unified referring video object segmentation network with a large-scale benchmark,'' in \emph{Computer Vision--ECCV 2020: 16th European Conference, Glasgow, UK, August 23--28, 2020, Proceedings, Part XV 16}.\hskip 1em plus 0.5em minus 0.4em\relax Springer, 2020, pp. 208--223.

\bibitem{loshchilov2017decoupled}
I.~Loshchilov and F.~Hutter, ``Decoupled weight decay regularization,'' \emph{arXiv preprint arXiv:1711.05101}, 2017.

\bibitem{zhou2017scene}
B.~Zhou, H.~Zhao, X.~Puig, S.~Fidler, A.~Barriuso, and A.~Torralba, ``Scene parsing through ade20k dataset,'' in \emph{Proceedings of the IEEE conference on computer vision and pattern recognition}, 2017, pp. 633--641.

\end{thebibliography}
}

\clearpage
\setcounter{page}{1}

\section*{Appendix A: General Transformation from Dialted Convolution to Non-dilated Large-Kernel Convolution}

Since \emph{ignoring pixels of the input is equivalent to inserting extra zero entries into the conv kernel}, \emph{a dilated conv layer with a small kernel can be equivalently converted into a non-dilated layer with a sparse larger kernel}. Let $k$ be the kernel size and $r$ be the dilation rate of the dilated layer, by inserting zero entries, the kernel size of the corresponding non-dilated layer will be $(k-1)r+1$, which is referred to as the \emph{equivalent kernel size} for brevity. 

\begin{figure}[ht]
		\includegraphics[width=\linewidth]{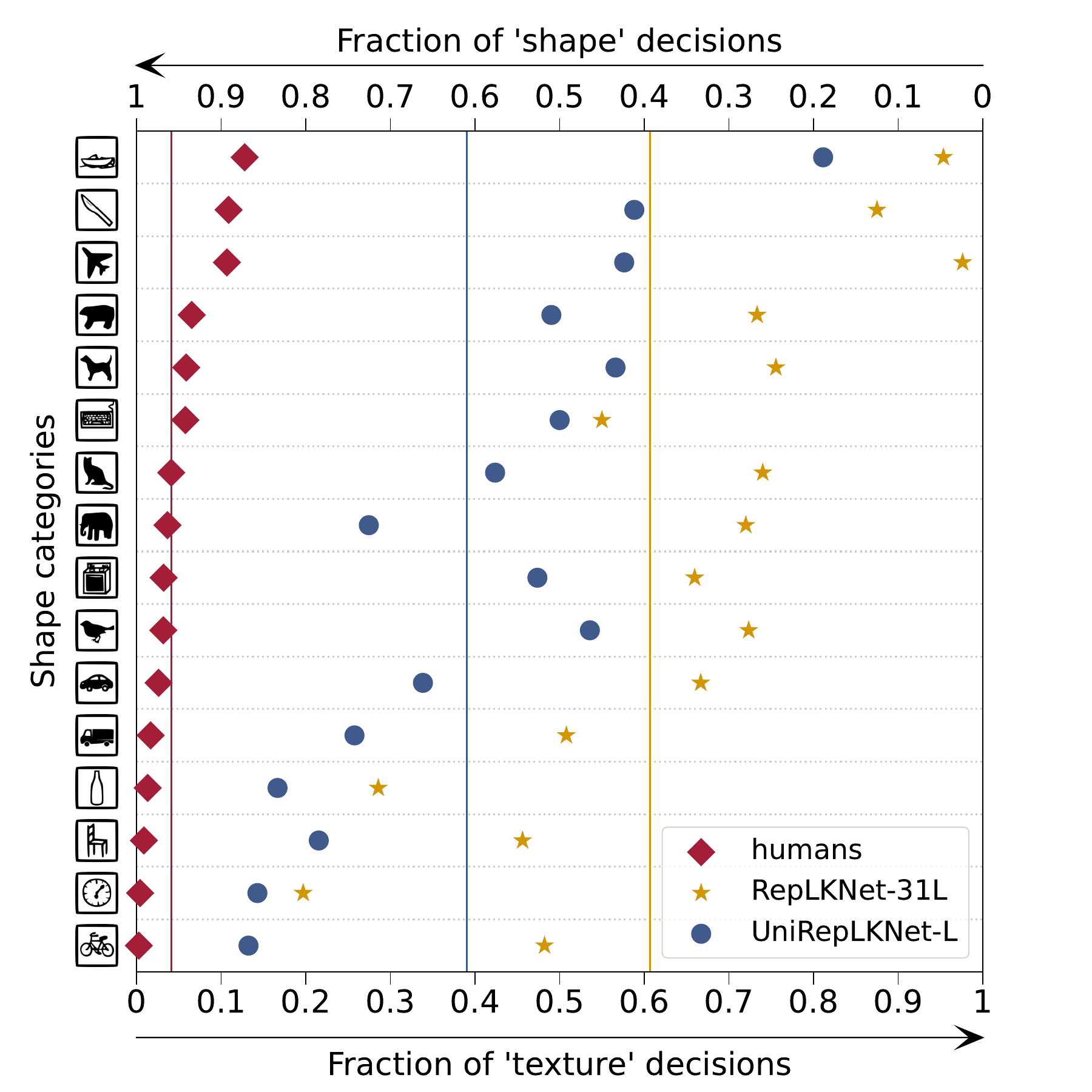}
  \vspace{-0.2in}
		\caption{Shape bias of ImageNet-22K-pretrained UniRepLKNet-L and RepLKNet-31L.}
  \label{fig-shape-bias-unireplknet}
\end{figure}

\begin{figure}[ht]
		\includegraphics[width=\linewidth]{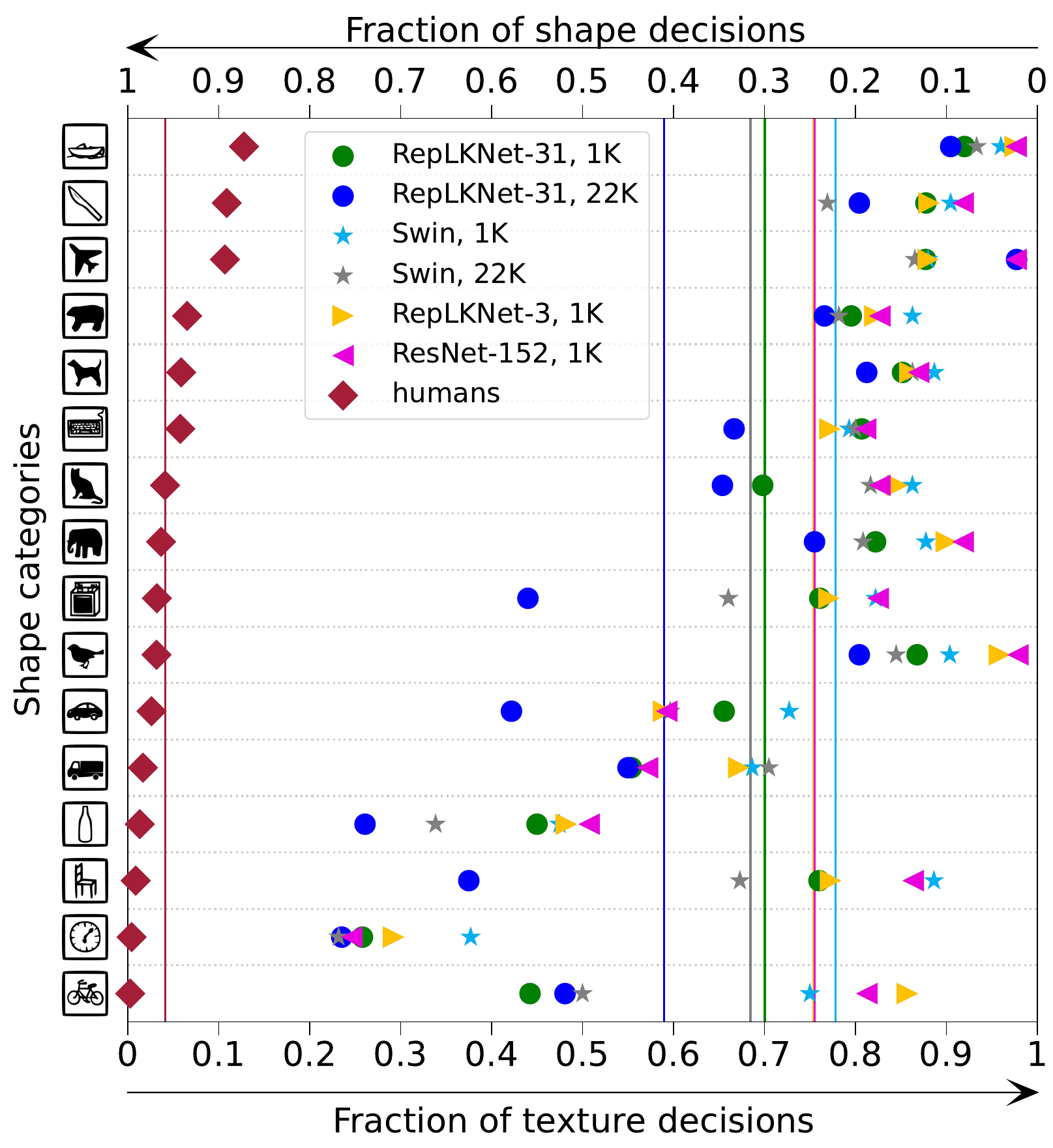}
  \vspace{-0.2in}
		\caption{Shape bias of ImageNet-1K and ImageNet-22K-pretrained RepLKNet-31B and Swin-B. This figure is directly taken from the supplementary material of RepLKNet without any modifications}
  \label{fig-shape-bias-replknet}
\end{figure}

As discussed in the paper, to eliminate the inference costs of the extra dilated conv layers in the Dilated Reparam Block, we propose to equivalently transform the whole block into a single non-dilated conv layer for inference. As discussed before, let $k$ and $r$ be the kernel size and dilation rate, respectively, the transformation from a dilated conv layer's kernel $\mathrm{W}\in\mathcal{R}^{k\times k}$ to a non-dilated layer's kernel $\mathrm{W}^\prime\in\mathcal{R}^{((k-1)r+1)\times ((k-1)r+1)}$ can be elegantly realized by a transpose convolution with a stride of $r$ and an identity kernel $\mathrm{I}\in\mathcal{R}^{1\times1}$, which is scalar 1 but viewed as a kernel tensor. That is
\begin{equation}\label{eq-merge_supp}
    \mathrm{W}^\prime = \mathtt{conv\_transpose2d}(\mathrm{W}, \mathrm{I}, \text{stride}=r) \,.
\end{equation}

In general cases with multi-channel conv layers, let the input channels, output channels, and number of groups be $c_{\text{in}}$, $c_{\text{out}}$, and $g$, respectively, we denote the kernel by a 4D tensor whose shape is $c_{\text{out}} \times \frac{c_{\text{in}}}{g} \times k\times k$. 

\textbf{1)} For a multi-channel depthwise (DW) layer, the transformation is easily generalized from 2D to 4D - the identity kernel $\mathrm{I}$ is viewed as a 4D tensor $\mathrm{I}\in\mathcal{R}^{1\times1\times1\times1}$ and we still follow function~\ref{eq-merge_supp} to derive the equivalent kernel by transpose convolution.

\textbf{2)} For non-DW cases (\ie, $g < c_{\text{in}}$), the transformation can be seen as splitting the kernel into slices (which can each be seen as a DW kernel), converting the slices respectively, and concatenating the resultant non-dilated slices up. We present the code in pytorch (Fig.~\ref{fig:code}) and a test case demonstrating the equivalency (Fig.~\ref{fig:testcase}).

\begin{figure*}
    \begin{lstlisting}[language=Python]
import torch
import torch.nn as nn
import torch.nn.functional as F

def convert_dilated_to_nondilated(kernel, dilate_rate):
    identity_kernel = torch.ones((1, 1, 1, 1))
    if kernel.size(1) == 1:
        #   This is a DW kernel
        dilated = F.conv_transpose2d(kernel, identity_kernel, stride=dilate_rate)
        return dilated
    else:
        #   This is a dense or group-wise (but not DW) kernel
        slices = []
        for i in range(kernel.size(1)):
            dilated = F.conv_transpose2d(kernel[:,i:i+1,:,:], identity_kernel, stride=dilate_rate)
            slices.append(dilated)
        return torch.cat(slices, dim=1)
    \end{lstlisting}
    \caption{Pytorch code to convert a dilated conv layer's small kernel to a non-dilated layer's larger sparse kernel.}
    \label{fig:code}
\end{figure*}

\begin{figure*}
    \begin{lstlisting}[language=Python]
def test_equivalency(in_channels, out_channels, groups, large_kernel_size, small_conv_r, small_conv_k):
    equivalent_kernel_size = small_conv_r * (small_conv_k - 1) + 1
    large_conv = nn.Conv2d(in_channels, out_channels, kernel_size=large_kernel_size,
                           padding=large_kernel_size // 2, groups=groups, bias=False)
    dilated_conv = nn.Conv2d(in_channels, out_channels, kernel_size=small_conv_k, 
    padding=equivalent_kernel_size // 2,
                             dilation=small_conv_r, groups=groups, bias=False)
    H, W = 19, 19
    x = torch.rand(2, in_channels, H, W)
    origin_y = large_conv(x) + dilated_conv(x)
    equivalent_kernel = convert_dilated_to_nondilated(dilated_conv.weight.data, small_conv_r)
    rows_to_pad = large_kernel_size // 2 - equivalent_kernel_size // 2
    merged_kernel = large_conv.weight.data + F.pad(equivalent_kernel, [rows_to_pad] * 4)
    equivalent_y = F.conv2d(x, merged_kernel, bias=None, padding=large_kernel_size // 2, groups=groups)
    print('relative error:', (equivalent_y - origin_y).abs().sum() / origin_y.abs().sum())

test_equivalency(in_channels=4, out_channels=4, groups=1, 
    large_kernel_size=13, small_conv_r=3, small_conv_k=3)
    \end{lstlisting}
    \caption{A test case demonstrating the equivalency of the transformation.}
    \label{fig:testcase}
\end{figure*}

\section*{Appendix B: Training Configurations}

We present the detailed training configurations for image classification, object detection, and semantic segmentation. We have publicly released a reproducible training script and trained weights for every model on GitHub.

\noindent\textbf{ImageNet image classification.} The training configurations for the ImageNet-1K-only results shown in Section 4 are presented in Table \ref{tab:supp_cls_1k}. These configurations are similar to common practices. For the experiments in Section 3, we use the same configurations, except that the training epochs are set to 100 and the drop path rate is set to 0.1. For the models pretrained with ImageNet-22K and then finetuned on ImageNet-22K, the configurations are shown in Table \ref{tab:supp_cls_1k}. Note that we follow the configurations adopted by ConvNeXt for a fair comparison with ConvNeXt-S/B, and the configurations used by InternImage for a fair comparison with InternImage-L/XL (the results with ImageNet-22K-pretrained InternImage-S/B were not reported).

\begin{table*}[t]
    \centering
    \renewcommand\arraystretch{1.0}
    \footnotesize
    \caption{\textbf{Detailed training configurations of ImageNet-1K-only models.} Apart from the configurations shown in the table, we use random left-right flipping, random resized crop, color jitter of 0.4, Auto-augment, and no repeated augmentation for every model.}
    \resizebox{0.98\linewidth}{!}{
\begin{tabular}{@{\ }l|c|c|c|c|c|c}
\hline
settings & UniRepLKNet-A & UniRepLKNet-F & UniRepLKNet-P & UniRepLKNet-N & UniRepLKNet-T & UniRepLKNet-S \\
\hline
input scale & 
224 & 
224 &
224 & 
224 & 
224 &
224 \\
batch size & 
4096 & 
4096 &
4096 & 
4096 & 
4096 &
4096 \\
optimizer &
AdamW & 
AdamW &
AdamW &
AdamW &
AdamW &
AdamW \\
LR      & 
4$\times10^{-3}$ & 
4$\times10^{-3}$ &
4$\times10^{-3}$ & 
4$\times10^{-3}$ & 
4$\times10^{-3}$ &
4$\times10^{-3}$ \\
LR schedule& 
cosine  &
cosine & 
cosine & 
cosine & 
cosine &
cosine \\
weight decay     &
0.05  & 
0.05  & 
0.05 & 
0.05 &
0.05 &
0.05 \\
warmup epochs & 
5 &
5 &
5 & 
5 &
5 &
5 \\
epochs & 
300 &
300 &
300 & 
300 &
300 &
300  \\
\hline
mixup alpha  & 
0.3 & 
0.3 & 
0.3 &
0.5 & 
0.8 &
0.8 \\
cutmix alpha &
0.3 & 
0.3 & 
0.3 &
0.5 & 
1.0 &
1.0 \\
erasing prob. &
0.25    &
0.25   &
0.25 &
0.25 &
0.25 & 
0.25 \\

\hline
label smoothing $\varepsilon$ & 
0.1 & 
0.1 &
0.1  &
0.1 & 
0.1  &
0.1  \\
drop path rate & 
0.0 & 
0.0 & 
0.1 & 
0.1 &
0.2 &
0.4 \\
\hline
\end{tabular}
}
    \label{tab:supp_cls_1k}
\end{table*}

\begin{table*}[t]
    \centering
    \renewcommand\arraystretch{1.0}
    \footnotesize
        \caption{\textbf{Detailed training configurations of models pretrained with ImageNet-22K (IN-22K pt) and then finetuned on ImageNet-1K (IN-1K ft).} Apart from the configurations shown in the table, we use random left-right flipping, random resized crop, color jitter of 0.4, Auto-augment, and no repeated augmentation for every model.}
    
\resizebox{0.98\linewidth}{!}{
\begin{tabular}{@{\ }l|cc|cc|cc|cc}
\hline
\multirow{2}{*}{settings} & \multicolumn{2}{c|}{UniRepLKNet-S} & \multicolumn{2}{c|}{UniRepLKNet-B} & \multicolumn{2}{c|}{UniRepLKNet-L} & \multicolumn{2}{c}{UniRepLKNet-XL} \\
\cline{2-9}
& 
IN-22K pt & 
IN-1K ft &
IN-22K pt & 
IN-1K ft &
IN-22K pt & 
IN-1K ft &
IN-22K pt & 
IN-1K ft \\
\hline
input scale & 
224 & 
384 &
224 & 
384 & 
192 & 
384 &
192 & 
384 \\
batch size & 
4096 &
512 &
4096 &
512 &
4096 &
512 &
4096 &
512 \\
optimizer &
AdamW & 
AdamW &
AdamW &
AdamW &
AdamW &
AdamW & AdamW & AdamW\\
LR      & 
4$\times10^{-3}$ &
5$\times10^{-5}$ &
4$\times10^{-3}$ &
5$\times10^{-5}$ & 
4$\times10^{-3}$ &
5$\times10^{-5}$ & 
4$\times10^{-3}$ &
5$\times10^{-5}$ \\
LR schedule& 
cosine  &
cosine & 
cosine & 
cosine & 
cosine &
cosine & cosine & cosine\\
weight decay     &
0.05  &
1$\times10^{-8}$ & 
0.05  &
1$\times10^{-8}$ & 
0.05  &
1$\times10^{-8}$ & 
0.05  &
1$\times10^{-8}$ \\ 
warmup epochs & 
5 &
0 &
5 &
0 &
5 &
0 &
5 &
0 \\
epochs & 
90 &
30 &
90 &
30 &
90 &
20 &
90 &
20 \\
\hline

mixup alpha  & 
0.8 & 
0.0 &
0.8 & 
0.0 &
0.8 & 
0.0 &
0.8 & 
0.0 \\
cutmix alpha &
1.0 & 
0.0 &
1.0 & 
0.0 &
1.0 & 
0.0 &
1.0 & 
0.0 \\
erasing prob. &
0.25    &
0.25   &
0.25 &
0.25 &
0.25    &
0.25   &
0.25 &
0.25 \\
\hline
label smoothing & 
0.1 & 
0.1 &
0.1  &
0.1 & 
0.1 & 
0.3 &
0.1  &
0.3 \\
drop path rate & 
0.1 & 
0.2 & 
0.1 & 
0.2 &
0.1 &
0.3 & 
0.2 & 
0.3 \\
\hline
\end{tabular}
}
    \label{tab:supp_cls_22k}
\end{table*}

\noindent\textbf{COCO object detection.} For fair comparisons, we follow common practices~\cite{liu2021swin,liu2022convnet} to initialize the backbone with pretrained weights and train the models using a 3$\times$ (36 epochs) schedule by default. The shorter side is resized to 480$-$800 pixels, while the longer side does not exceed 1,333 pixels. All the models are trained with a batch size of 16 and AdamW~\cite{loshchilov2017decoupled} optimizer with an initial learning rate of $1\times10^{-4}$. We have publicly released the training configuration files used in the MMDetection framework and trained weights.

\noindent\textbf{ADE20K semantic segmentation.} We evaluate UniRepLKNet models on the ADE20K dataset~\cite{zhou2017scene}, and initialize them with the pre-trained classification weights. The learning rate is initialized with $1\times10^{-4}$ and decayed with the polynomial decay schedule with a power of 1.0. Following previous methods~\cite{liu2021swin,liu2022convnet}, the crop size is set to 512 for the ImageNet-1K-pretrained models, and 640 for ImageNet-22K-pretrained models. All segmentation models are trained with a batch size of 16 for 160k iterations. We have publicly released the training configuration files used in the MMSegmentation framework and trained weights.

\section*{Appendix C: Shape Bias}

A higher shape bias means the model makes predictions based more on the shape of objects rather than the textures, \ie, the model behaves more similarly to humans. Therefore, a model with a higher shape bias may transfer better to downstream tasks. UniRepLKNet demonstrates significantly higher shape bias than existing ConvNets and ViTs. Concretely, we test the shape bias of ImageNet-22K-pretrained UniRepLKNet-L and RepLKNet-L with the \textit{modelvshuman} toolbox~\footnote{\url{https://github.com/bethgelab/model-vs-human}}. Fig.~\ref{fig-shape-bias-unireplknet} shows a significantly higher shape bias of UniRepLKNet - UniRepLKNet makes 20\% more decisions based on the overall shapes of objects. This improvement is particularly remarkable since RepLKNet is already known to have a high shape bias (Fig.~\ref{fig-shape-bias-replknet} is directly taken from the supplementary material of the RepLKNet paper without any modifications).

\subsection{Appendix D: Training Memory Footprint}

The extra parallel dilated branches in Dilated Reparam Block consume more training resources, which is acceptable considering the performance improvements. We present the peak GPU memory footprint and training speed in Table~\ref{table-costs}. With a bigger model and bigger data, we may trade the performance for higher training speed and lower memory consumption by replacing the Dilated Reparam Block with a single large-kernel conv layer followed by Batch Normalization layer. We test the peak memory footprint and actual training throughput while training UniRepLKNet-S with 224$\times$224 inputs and a batch size of 4096 on a node with eight A100 GPUs. Note that such results are significantly influenced by the hardware environment and specific implementation; thus, they should be considered as references only.

	\begin{table}
		\caption{Training costs.}
		\label{table-costs}
		% \vspace{-0.3in}
		\begin{center}
        \resizebox{1.0\linewidth}{!}{
			\tiny
			\begin{tabular}{lcccccccc}
				\hline
			 & Peak memory     & Training throughput \\
				\hline
                Dilated Reparam Block   & 24.6GB   &  6642 images/s   \\
                Single large-kernel conv layer       & 20.8GB    & 9675 images/s \\

				\hline
			\end{tabular}
        }
		\end{center}
		\vspace{-0.3in}
	\end{table}

\end{document}